\documentclass[journal]{IEEEtran}
\usepackage{amsmath,amsfonts}
\usepackage{algorithmic}
\usepackage{algorithm}
\usepackage{array}
\usepackage[caption=false,font=normalsize,labelfont=sf,textfont=sf]{subfig}
\usepackage{textcomp}
\usepackage{stfloats}
\usepackage{url}
\usepackage{verbatim}
\usepackage{graphicx}
\usepackage{cite}
\usepackage[dvipsnames]{xcolor}    
\usepackage{pifont}
\usepackage{booktabs}
\usepackage{multirow}
\definecolor{ieeeblue}{rgb}{0.21,0.49,0.74}
\usepackage[hidelinks,colorlinks,citecolor=ieeeblue, urlcolor=ieeeblue]{hyperref} 
\hypersetup{linktoc=page} 
\usepackage{colortbl}  
\usepackage{graphicx}
\usepackage{fontawesome}

\makeatletter
\renewcommand{\l@section}{\@dottedtocline{1}{0em}{2.5em}}
\renewcommand{\l@subsection}{\@dottedtocline{2}{2em}{3.5em}}
\renewcommand{\l@subsubsection}{\@dottedtocline{3}{3.5em}{4em}}
\makeatother

\begin{document}

\title{A Survey of Task-Oriented Knowledge Graph Reasoning: Status, Applications, and Prospects}

\author{Guanglin Niu, Bo Li, Yangguang Lin\\
\faGithub\ \href{https://github.com/ngl567/KGR-Survey}{{\texttt{https://github.com/ngl567/KGR-Survey}}}
\thanks{Guanglin Niu and Bo Li are with the School of Artificial Intelligence, Beihang University. Yangguang Lin is with the School of Information Science and Technology, Beijing Forestry University. (E-mail: beihangngl@buaa.edu.cn; boli@buaa.edu.cn; sunshinelin0314@bjfu.edu.cn)

Corresponding author: Bo Li.}
}

\markboth{Journal of \LaTeX\ Class Files,~Vol.~14, No.~8, August~2021}%
{Shell \MakeLowercase{\textit{et al.}}: A Sample Article Using IEEEtran.cls for IEEE Journals}


\maketitle

\begin{abstract}
Knowledge graphs (KGs) have emerged as a powerful paradigm for structuring and leveraging diverse real-world knowledge, which serve as a fundamental technology for enabling cognitive intelligence systems with advanced understanding and reasoning capabilities. Knowledge graph reasoning (KGR) aims to infer new knowledge based on existing facts in KGs, playing a crucial role in applications such as public security intelligence, intelligent healthcare, and financial risk assessment. From a task-centric perspective, existing KGR approaches can be broadly classified into static single-step KGR, static multi-step KGR, dynamic KGR, multi-modal KGR, few-shot KGR, and inductive KGR. While existing surveys have covered these six types of KGR tasks, a comprehensive review that systematically summarizes all KGR tasks particularly including downstream applications and more challenging reasoning paradigms remains lacking. In contrast to previous works, this survey provides a more comprehensive perspective on the research of KGR by categorizing approaches based on primary reasoning tasks, downstream application tasks, and potential challenging reasoning tasks. Besides, we explore advanced techniques, such as large language models (LLMs), and their impact on KGR. This work aims to highlight key research trends and outline promising future directions in the field of KGR.

\end{abstract}


{\renewcommand{\baselinestretch}{1.1}\normalsize
\tableofcontents}

\section{Introduction}
Deep learning has achieved remarkable success in perceptual intelligence tasks, capitalizing on substantial computational power and vast quantities of labeled data. However, its limitations in knowledge representation hinder its ability to perform complex tasks such as knowledge reasoning. In contrast, knowledge graphs (KGs) have emerged as a powerful technique for modeling and utilizing knowledge by representing entities and their relations through a directed graph structure\cite{1,2}, such as some notable examples Freebase\cite{3}, YAGO\cite{4}, WordNet\cite{5}, NELL\cite{6}, and DBpedia\cite{7}. In recent years, KGs have played crucial roles in various tasks, such as relation extraction\cite{8}, semantic search\cite{9}, dialogue systems\cite{10}, question answering systems\cite{11}, recommendation systems\cite{12}, and have also facilitated large language models such as GraphRAG\cite{graphrag}.

In real-world scenarios, most KGs are inevitably incomplete. For instance, 75\% of the people in Freebase lack nationality information\cite{14}, and 60\% of the people in DBpedia lack birthplace information\cite{15}. Thus, knowledge graph reasoning (KGR) (also known as knowledge graph completion) techniques~\cite{nguyen-2020-survey} aim to discover semantic associations from the existing knowledge in a KG and infer unknown facts to complete the KG. Currently, KGR techniques have been widely applied in various practical scenarios, including COVID-19 drug discovery\cite{19}, bond market risk supervision\cite{120}, product recommendation\cite{21}, voice assistants\cite{22}, and intelligent education~\cite{457}.

\begin{figure}
    \centering
    \includegraphics[width=1\linewidth]{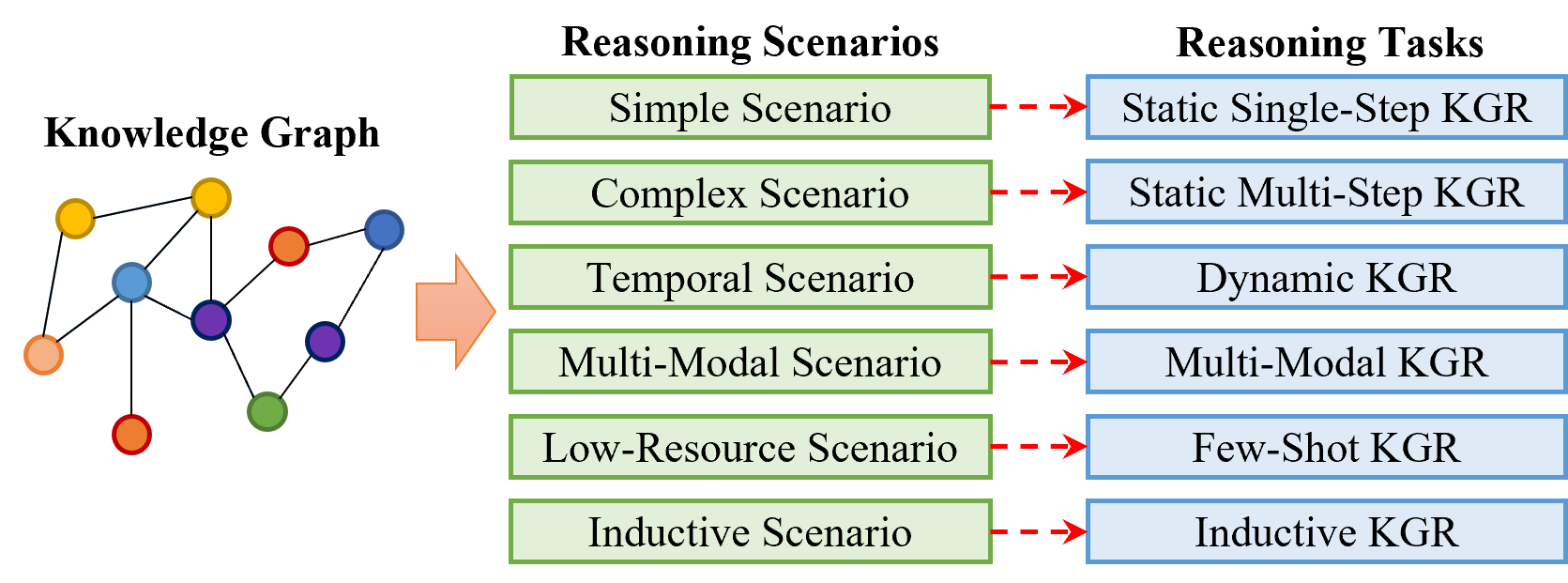}
    \caption{Knowledge graph reasoning tasks for various reasoning scenarios.}
    \label{fig:intro}
\end{figure}

Despite the progress made in KGR, existing models face several challenges. Rule-based methods, which rely on domain experts to construct logic rules, often struggle with high computational complexity and poor scalability, rendering them inefficient for large-scale KGs\cite{KGEsurvey}. In recent years, knowledge graph embedding (KGE) has emerged as a popular technique, which embeds the symbolic entities and relations in KGs into embeddings spaces. KGE models allow for the calculation of the likelihood of unknown fact triples through numerical computations, benefiting to good scalability and computational efficiency and have become a major research direction in KGR. However, these models lose the symbolic nature of knowledge, leading to poor explainability.

\begin{table*}[ht]
\scriptsize
\centering
\caption{Comparison of our survey with some typical review papers in the recent three years concerning KGR\label{tab:review_comparison}}
\setlength{\tabcolsep}{3pt} 
\renewcommand\arraystretch{1.2}
\begin{tabular}{cc|cccccccccccc}
\toprule
\multicolumn{2}{c|}{Survey Papers} &~\cite{52} &~\cite{54} &~\cite{61} &~\cite{53} &~\cite{55} &~\cite{56} &~\cite{57} &~\cite{58} &~\cite{59} &~\cite{64} &~\cite{43} & Ours \\
\midrule
 & Single-Step KGR    & \textcolor{ForestGreen}{\checkmark} & \textcolor{ForestGreen}{\checkmark} & \textcolor{ForestGreen}{\checkmark} & \textcolor{ForestGreen}{\checkmark} & \textcolor{red!70!black}{\ding{55}} & \textcolor{red!70!black}{\ding{55}} & \textcolor{red!70!black}{\ding{55}} & \textcolor{red!70!black}{\ding{55}} & \textcolor{red!70!black}{\ding{55}} & \textcolor{ForestGreen}{\checkmark} & \textcolor{ForestGreen}{\checkmark} & \textcolor{ForestGreen}{\checkmark} \\
 & Multi-Step KGR     & \textcolor{red!70!black}{\ding{55}}  & \textcolor{red!70!black}{\ding{55}}  & \textcolor{red!70!black}{\ding{55}} & \textcolor{ForestGreen}{\checkmark} & \textcolor{red!70!black}{\ding{55}} & \textcolor{red!70!black}{\ding{55}} & \textcolor{red!70!black}{\ding{55}} & \textcolor{red!70!black}{\ding{55}} & \textcolor{red!70!black}{\ding{55}} & \textcolor{ForestGreen}{\checkmark} & \textcolor{ForestGreen}{\checkmark} & \textcolor{ForestGreen}{\checkmark} \\
KGR & Dynamic KGR      & \textcolor{red!70!black}{\ding{55}}  & \textcolor{red!70!black}{\ding{55}}  & \textcolor{red!70!black}{\ding{55}} & \textcolor{ForestGreen}{\checkmark} & \textcolor{ForestGreen}{\checkmark} & \textcolor{ForestGreen}{\checkmark} & \textcolor{red!70!black}{\ding{55}} & \textcolor{red!70!black}{\ding{55}} & \textcolor{red!70!black}{\ding{55}} & \textcolor{red!70!black}{\ding{55}} & \textcolor{ForestGreen}{\checkmark} & \textcolor{ForestGreen}{\checkmark} \\
Tasks & Multi-Modal KGR  & \textcolor{red!70!black}{\ding{55}}  & \textcolor{red!70!black}{\ding{55}}  & \textcolor{red!70!black}{\ding{55}} & \textcolor{red!70!black}{\ding{55}} & \textcolor{red!70!black}{\ding{55}} & \textcolor{red!70!black}{\ding{55}} & \textcolor{red!70!black}{\ding{55}} & \textcolor{red!70!black}{\ding{55}} & \textcolor{red!70!black}{\ding{55}} & \textcolor{ForestGreen}{\checkmark} & \textcolor{ForestGreen}{\checkmark} & \textcolor{ForestGreen}{\checkmark} \\
 & Few-Shot KGR      & \textcolor{red!70!black}{\ding{55}}  & \textcolor{red!70!black}{\ding{55}}  & \textcolor{red!70!black}{\ding{55}} & \textcolor{red!70!black}{\ding{55}} & \textcolor{red!70!black}{\ding{55}} & \textcolor{red!70!black}{\ding{55}} & \textcolor{ForestGreen}{\checkmark} & \textcolor{ForestGreen}{\checkmark} & \textcolor{ForestGreen}{\checkmark} & \textcolor{ForestGreen}{\checkmark} & \textcolor{ForestGreen}{\checkmark} & \textcolor{ForestGreen}{\checkmark} \\
 & Inductive KGR     & \textcolor{red!70!black}{\ding{55}}  & \textcolor{red!70!black}{\ding{55}}  & \textcolor{red!70!black}{\ding{55}} & \textcolor{red!70!black}{\ding{55}} & \textcolor{red!70!black}{\ding{55}} & \textcolor{red!70!black}{\ding{55}} & \textcolor{ForestGreen}{\checkmark} & \textcolor{red!70!black}{\ding{55}} & \textcolor{ForestGreen}{\checkmark} & \textcolor{red!70!black}{\ding{55}} & \textcolor{ForestGreen}{\checkmark} & \textcolor{ForestGreen}{\checkmark} \\
\midrule
\multirow{2}{*}{Applications} & Benchmarks      & \textcolor{red!70!black}{\ding{55}}  & \textcolor{red!70!black}{\ding{55}}  & \textcolor{red!70!black}{\ding{55}} & \textcolor{red!70!black}{\ding{55}} & \textcolor{red!70!black}{\ding{55}} & \textcolor{red!70!black}{\ding{55}} & \textcolor{ForestGreen}{\checkmark} & \textcolor{ForestGreen}{\checkmark} & \textcolor{ForestGreen}{\checkmark} & \textcolor{ForestGreen}{\checkmark} & \textcolor{ForestGreen}{\checkmark} & \textcolor{ForestGreen}{\checkmark} \\
                              & Downstream Tasks & \textcolor{red!70!black}{\ding{55}}  & \textcolor{red!70!black}{\ding{55}}  & \textcolor{red!70!black}{\ding{55}} & \textcolor{red!70!black}{\ding{55}} & \textcolor{ForestGreen}{\checkmark} & \textcolor{ForestGreen}{\checkmark} & \textcolor{red!70!black}{\ding{55}} & \textcolor{red!70!black}{\ding{55}} & \textcolor{red!70!black}{\ding{55}} & \textcolor{red!70!black}{\ding{55}} & \textcolor{red!70!black}{\ding{55}} & \textcolor{ForestGreen}{\checkmark} \\
\midrule
 & Sparseness        & \textcolor{red!70!black}{\ding{55}}  & \textcolor{red!70!black}{\ding{55}}  & \textcolor{red!70!black}{\ding{55}} & \textcolor{red!70!black}{\ding{55}} & \textcolor{red!70!black}{\ding{55}} & \textcolor{red!70!black}{\ding{55}} & \textcolor{red!70!black}{\ding{55}} & \textcolor{red!70!black}{\ding{55}} & \textcolor{red!70!black}{\ding{55}} & \textcolor{red!70!black}{\ding{55}} & \textcolor{red!70!black}{\ding{55}} & \textcolor{ForestGreen}{\checkmark} \\
Challenge & Uncertainty & \textcolor{ForestGreen}{\checkmark} & \textcolor{red!70!black}{\ding{55}}  & \textcolor{red!70!black}{\ding{55}} & \textcolor{red!70!black}{\ding{55}} & \textcolor{red!70!black}{\ding{55}} & \textcolor{red!70!black}{\ding{55}} & \textcolor{red!70!black}{\ding{55}} & \textcolor{red!70!black}{\ding{55}} & \textcolor{red!70!black}{\ding{55}} & \textcolor{red!70!black}{\ding{55}} & \textcolor{red!70!black}{\ding{55}} & \textcolor{ForestGreen}{\checkmark} \\
and & Error          & \textcolor{red!70!black}{\ding{55}}  & \textcolor{red!70!black}{\ding{55}}  & \textcolor{red!70!black}{\ding{55}} & \textcolor{red!70!black}{\ding{55}} & \textcolor{red!70!black}{\ding{55}} & \textcolor{red!70!black}{\ding{55}} & \textcolor{red!70!black}{\ding{55}} & \textcolor{red!70!black}{\ding{55}} & \textcolor{red!70!black}{\ding{55}} & \textcolor{red!70!black}{\ding{55}} & \textcolor{red!70!black}{\ding{55}} & \textcolor{ForestGreen}{\checkmark} \\
Opportunities & Explainability & \textcolor{red!70!black}{\ding{55}}  & \textcolor{red!70!black}{\ding{55}}  & \textcolor{ForestGreen}{\checkmark} & \textcolor{ForestGreen}{\checkmark} & \textcolor{ForestGreen}{\checkmark} & \textcolor{ForestGreen}{\checkmark} & \textcolor{red!70!black}{\ding{55}} & \textcolor{red!70!black}{\ding{55}} & \textcolor{red!70!black}{\ding{55}} & \textcolor{ForestGreen}{\checkmark} & \textcolor{ForestGreen}{\checkmark} & \textcolor{ForestGreen}{\checkmark} \\
           & LLM           & \textcolor{red!70!black}{\ding{55}}  & \textcolor{red!70!black}{\ding{55}}  & \textcolor{red!70!black}{\ding{55}} & \textcolor{red!70!black}{\ding{55}} & \textcolor{ForestGreen}{\checkmark} & \textcolor{ForestGreen}{\checkmark} & \textcolor{red!70!black}{\ding{55}} & \textcolor{red!70!black}{\ding{55}} & \textcolor{red!70!black}{\ding{55}} & \textcolor{red!70!black}{\ding{55}} & \textcolor{red!70!black}{\ding{55}} & \textcolor{ForestGreen}{\checkmark} \\
\midrule
\multicolumn{2}{c|}{\multirow{2}{*}{Main Characteristics}} & Embedding & Negative & Logics \& & Causal & \multicolumn{2}{c}{Temporal} & Unseen & Common- & Few-shot \& & Multi-modal \& & Graph & Task-oriented \\
\multicolumn{2}{c|}{} & Spaces    & Sampling & Embeddings  & Reasoning & \multicolumn{2}{c}{KGR} & Elements & sense   & Inductive  & Hyper-relation  & Types & KGR Survey \\
\bottomrule
\end{tabular}
\end{table*}

\subsection{Motivation and Contribution}

Existing review papers on KGR have primarily focused on specific scenarios of KGR, such as multi-modal KGR\cite{39}, temporal KGR\cite{53,55}, and KGR with unseen entities and relations\cite{57,58}. Liang et al.\cite{43} provided a review of KGR approaches for static, dynamic, and multi-modal KGs. However, the previous reviews lack a more systematic and comprehensive taxonomy of KGR approaches, together with the downstream applications drivend by KGR techiniques and some more challenging tasks from the perspective of knowledge characteristics. Thus, this survey aims to address this gap by proposing a task-oriented classification system for KGR approaches. We categorize KGR tasks into six types based on the characteristics of KGs and the reasoning scenarios as shown in Fig.~\ref{fig:intro}, namely static single-step KGR for simple reasoning scenario, static multi-step KGR for complex reasoning scenario, dynamic KGR for temporal reasoning scenario, multi-modal KGR for multi-modal reasoning scenario, few-shot KGR for low-resource scenario, and inductive KGR for inductive reasoning scenario. We provide a detailed review of existing approaches for each category, highlighting their strengths, limitations, downstream applications and potential directions for future research.

\subsection{Related Literature Reviews}

In the recent three years, there are several representative review papers in the field of KGR\cite{43,52,53,54,55,56,57,58,59,61,64}. However, most of these reviews focus on single KGR task or partial KGR tasks and ignore the more challenging aspects for KGR such as sparseness and uncertainty of KGs. In specific, our survey highlights the differences and main characteristics as shown in TABLE~\ref{tab:review_comparison}.

\begin{itemize}
    \item The previous reviews~\cite{52, 54, 61} all introduce single-step KGR models specific to static KGs. Specifically,~\cite{52} discusses knowledge graph embedding (KGE) models from the perspective of embedding spaces based on algebraic, geometric, and analytic structures. It highlights the distinct characteristics and advantages of these embedding spaces for KGE.~\cite{54} provides a comprehensive summary of several negative sampling techniques required for training KGE models.~\cite{61} reviews approaches that integrate symbolic logic with KGE models and highlights the advantages of incorporating symbolic logic into KG embeddings in terms of explainability.

    \item Three surveys~\cite{53, 55, 56} all provide review of existing KGR approaches for dynamic KGs. It is worth noting that these surveys focus on the temporal KGR task, which is a sub-task of the dynamic KGR. Specifically,~\cite{53} discusses temporal KGR models from the perspectives of causal reasoning and commonsense reasoning.~\cite{55} reviews totally ten categories of temporal KGR models, analyzing the current state-of-the-art techniques and their practical applications.~\cite{56} provides a detailed introduction to the research status and development of temporal KGR, categorizing the approaches into two main principles namely KG completion and event prediction.

    \item The existing reviews~\cite{57, 58, 59} summarize models for the few-shot KGR task, and~\cite{57} and~\cite{58} further discuss inductive KGR approaches. Both few-shot KGR and inductive KGR tasks differ from traditional transductive reasoning tasks, focusing particularly on long-tailed entities and relations or even those that have not been observed during training. Specifically,~\cite{57} systematically reviews existing few-shot and inductive KGR models from two specific perspectives: unseen entities and unseen relations.~\cite{58} reviews typical few-shot KGR approaches from two perspectives: structural information based on neighbors or paths, and commonsense knowledge derived from pre-trained language models.~\cite{59} provides an overview of the latest research on inductive, zero-shot, and few-shot KGR models.

    \item Shen et al.~\cite{64} review the status of KGR models from five perspectives: (1) solely on triples, (2) incorporating external information, (3) temporal KGs, (4) commonsense KGs, and (5) hyper-relational KGs. Liang et al.~\cite{43} categorize KGR models based on static, dynamic, and multi-modal KGs, providing a comprehensive summary of the approaches from the perspective of KG types.
    
    \item Although these survey papers have covered the six types of KGR tasks discussed in our paper, there is still a lack of reviews that systematically summarize research on all types of KGR tasks together with downstream tasks and more challenging tasks. Compared to the previous reviews, we achieve a more comprehensive survey for KGR models from the prospective of primary KGR tasks, downstream applications and challenging tasks with some advanced techniques such as large language models (LLMs). We aim to highlight the research hotspots on KGR and future directions worth further exploring.
\end{itemize}

\begin{figure*}
    \centering
    \includegraphics[width=1\linewidth]{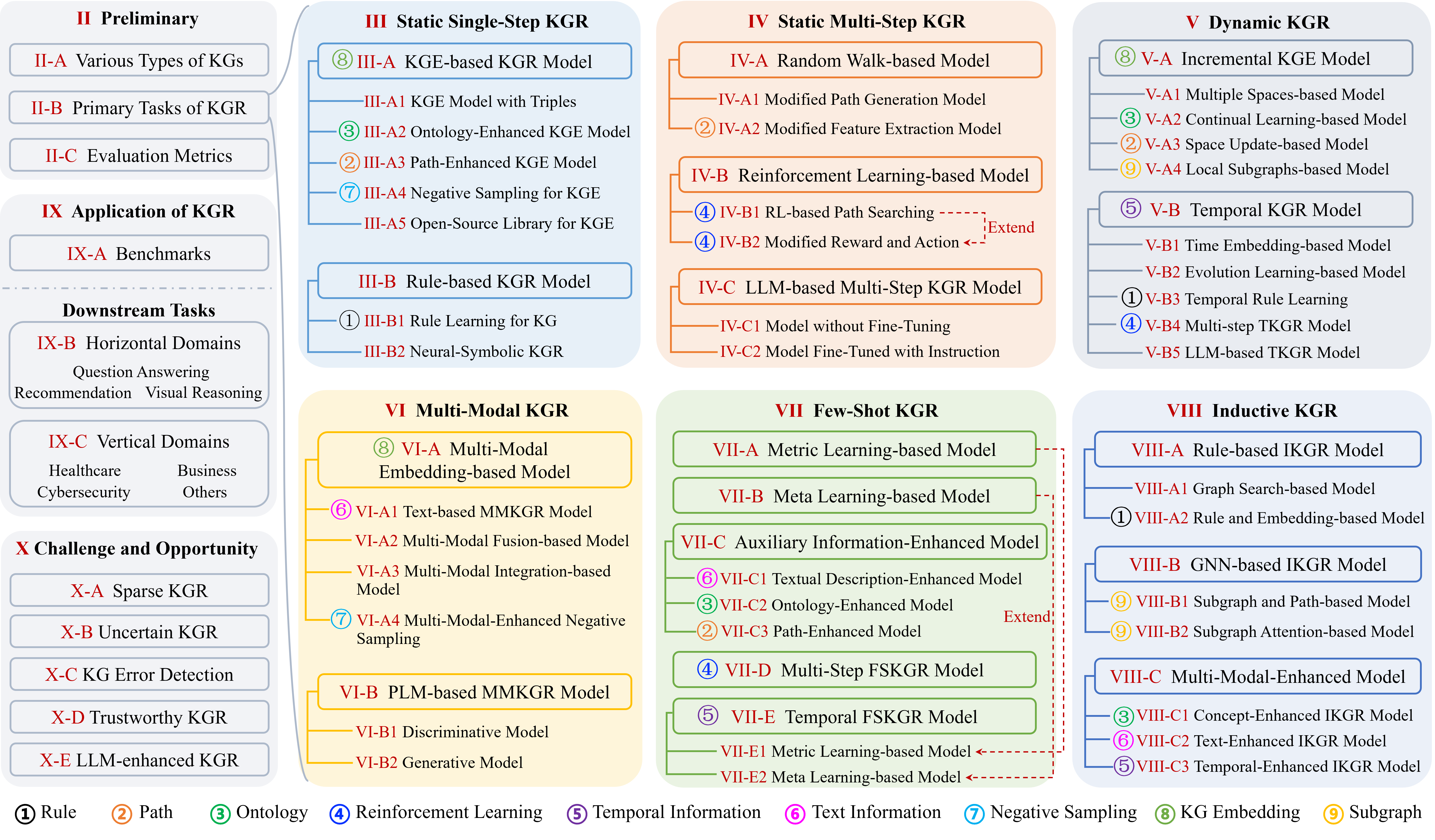}
    \caption{Comprehensive overview framework of our survey. The same number (\ding{172}-\ding{180}) indicates that different approaches share similar ideas, and the keywords corresponding to each number are provided at the bottom of the figure.}
    \label{fig:framework}
\end{figure*}

\subsection{Article’s Organization}

In this era of large language models (LLMs), our work provides a comprehensive and novel review of KGR approaches from a task-oriented perspective, as illustrated in Fig.~\ref{fig:framework}. Specifically, Section~\ref{sec:pre} categorizes several types of KGs and introduces six classes of primary KGR tasks, along with the evaluation metrics for KGR. Sections~\ref{sec:single-step}-\ref{sec:inductive} systematically review the models corresponding to these six primary KGR tasks, namely Static Single-Step KGR, Static Multi-Step KGR, Dynamic KGR, Multi-Modal KGR, Few-Shot KGR, and Inductive KGR. Notably, we analyze the logical connections between approaches of various KGR tasks. As shown in Fig.~\ref{fig:framework}, dashed red lines indicate extension associations between linked models, while numbered circles (\ding{172} to \ding{180}) denote the approaches sharing similar ideas. For instance, \ding{172} denotes that the consistent idea of rule learning for static single-step KGR, temporal rule learning for dynamic KGR as well as rule and embedding-based model for inductive KGR concentrates on logic rule.

Section~\ref{sec:app} discusses the applications of KGR, including commonly used benchmark datasets for KGR tasks and downstream tasks of KGR technology in both horizontal and vertical domains. Furthermore, in Section~\ref{sec:challenge}, we introduce several research directions that are both challenging and replete with opportunities, deserving further in-depth investigation in the future. This comprehensive review aims to provide a clear roadmap for researchers in the field, facilitating further advancements in KGR by highlighting current research hotspots and future directions. Finally, Section~\ref{sec:conclusion} concludes this paper.

\section{Preliminary}
\label{sec:pre}

Considering that the characteristics of primary KGR tasks are closely associated with the types of KGs and their respective reasoning scenarios, to facilitate the following systematic review and understanding of KGR approaches, this section first introduces several typical types of KGs, including static KGs, dynamic KGs, and multi-modal KGs, with descriptions for each category as followings.

\subsection{Various Types of KGs}

(1) Static KGs~\cite{2}:
A traditional static KG is typically composed of factual triples, where each triple consists of a head entity, a tail entity, and their relation. Here, a static KG can be represented as a directed graph comprising nodes (entities) and edges (relations) labeled with distinct relational categories. In this directed graph structure, inter-node paths and neighborhood subgraphs constitute critical topological information, reflecting multi-step semantic associations between nodes and contextual semantics of nodes, respectively. Since most publicly available KGs fall under the static category, the majority of existing KGR research focuses on static KGs.

(2) Dynamic KGs~\cite{44}:
In real-world scenarios, data from domains such as finance, news, and academia often exhibit inherent dynamic characteristics. KGs constructed from such data are classified as dynamic KGs. Specifically, the ``dynamic'' nature manifests in two aspects. First, it refers to knowledge evolution, including addition, deletion, and modification of factual triples, as well as the emergence of novel entities and relations. Second, it pertains to temporal property, where knowledge evolves over time by incorporating a temporal dimension into triples. Such graphs are commonly termed temporal KGs and are represented as event quadruples $(s,p,o,t)$, where $s$, $p$, and $o$ denote the subject, predicate, and object, respectively, and $t$ represents the timestamp.

(3) Multi-Modal KGs~\cite{45}:
In practice, KG construction relies on multi-modal data sources (e.g., text, images, audio). However, symbolic entities and relations extracted from these data often lack the rich features inherent to multi-modal inputs. To address this limitation, multi-modal KGs (MMKGs) have recently gained research attention. This type of KG augments entities and relations with multi-modal data (e.g., images, textual descriptions) to enrich their embeddings~\cite{46}.

\begin{figure*}
    \centering
    \includegraphics[width=1\linewidth]{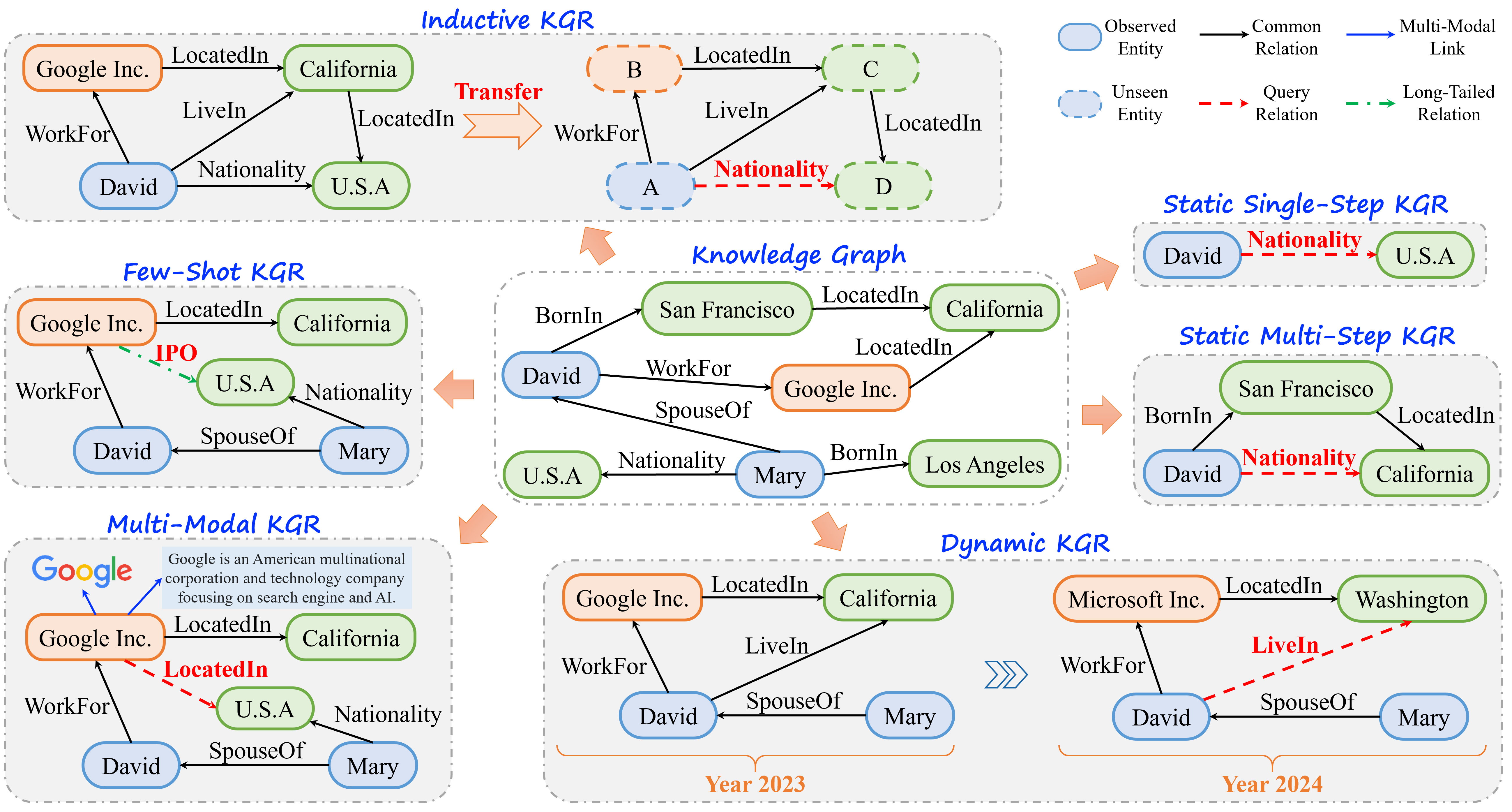}
    \caption{The illustration of the six primary KGR tasks.}
    \label{fig:KGRtask}
\end{figure*}

\subsection{Primary Tasks of KGR}

The core objective of KGR is to predict missing entities or relations in a factual triple or event quadruple based on the observed knowledge. For this goal, KGR tasks can be categorized into six types, depending on the KG types and especially reasoning scenarios: static single-step KGR, static multi-step KGR, dynamic KGR, multi-modal KGR, few-shot KGR, and inductive KGR, as summarized in Fig.~\ref{fig:KGRtask}.

The descriptions of these tasks are as follows:

(1) Static Single-Step KGR~\cite{61}:
This foundational task involves mining semantic associations between entities from known triples in a static KG and transferring these associations to predict missing elements in unknown triples. For example, given the head entity $David$ and the relation $Nationality$, the task infers the tail entity as shown in Fig.~\ref{fig:KGRtask}. Current approaches include knowledge graph embedding (KGE) and symbolic rule-based methods, discussed in Section~\ref{sec:single-step}.

(2) Static Multi-Step KGR~\cite{75}:
Unlike single-step KGR, the static multi-step KGR task requires capturing intermediate semantic associations between entities along multi-step paths. For instance in Fig.~\ref{fig:KGRtask}, predicting the tail entity with the given head entity $David$ and relation $Nationality$ involves analyzing multi-step relational paths in the graph namely $David \xrightarrow{BornIn} San\ Francisco \xrightarrow{LocatedIn} California$. The models for addressing this task are detailed in Section~\ref{sec:multi-step}.

(3) Dynamic KGR~\cite{56}:
In dynamic KGR task, it is essential to capture both the semantic information of changes such as additions, deletions, and modifications of knowledge as well as the temporal information related to the evolution of events. This dual focus ensures a comprehensive understanding of the KG's dynamic nature and enhances the accuracy of reasoning processes. Approaches are introduced in Section~\ref{sec:dynamic}.

(4) Multi-Modal KGR~\cite{39}:
This task leverages multi-modal data (text, images, videos) to enrich entity and relation representations. For instance, images and textual descriptions of $Google\ Inc.$ in Fig.~\ref{fig:KGRtask} supplement structural features for representing the entity $Google\ Inc.$. The key challenge is fusing multi-modal features with topological patterns, often using KGE frameworks, as reviewed in Section~\ref{sec:multi-modal}.

(5) Few-Shot KGR~\cite{fewshotsurvey}:
This task focuses on tackling the challenges of long-tailed entity/relation distributions and sparse associations, such as those involving newly added entities in dynamic KGs. In such cases, the available training samples are extremely limited. For example, the relation $IPO$ may be associated with only a few triples as in Fig.~\ref{fig:KGRtask}. The primary objective is to develop approaches capable of generalizing associations from this limited knowledge to enable accurate predictions in scenarios with sparse data. Various approaches to address this challenge are explored and discussed in Section~\ref{sec:few-shot}.

(6) Inductive KGR~\cite{57}:
Inductive KGR distinguishes itself by handling unseen entities (e.g., $A$, $B$, $C$ and $D$ in Fig.~\ref{fig:KGRtask}) during inference, unlike tasks that rely solely on known entities/relations. The objective of inductive KGR is to derive representations for unseen entities/relations by generalizing semantic patterns from observed triples and inferring new facts. The approaches to achieve this objective are outlined in Section~\ref{sec:inductive}.

\subsection{Performance Evaluation Metrics}

For each category of KGR tasks, the primary goal is to infer the unseen factual triple or event quadruplet when an element is missing. To assess the reasoning performance of each method, the current standard approach employs an information-retrieval-based evaluation mechanism. This mechanism determines the ranking of the correct reasoning results among all candidates. A higher ranking indicates better inference accuracy. Taking head entity prediction $(?,r,t)$ as an example, each entity in the KG is filled into the head entity position to construct candidate triples $(h,r,t)$, respectively. The candidates are then ranked according to the likelihood of being valid, as defined by the evaluation mechanism of each method. This process allows for the determination of the ranking of each correct prediction result. The evaluation methods for tail entity prediction and relation prediction follow a similar procedure~\cite{93}. This paper declares three commonly used metrics in KGR~\cite{94}:

(1) MR (Mean Rank): The average rank of all correct triples, calculated as:
\begin{equation}
    MR = \frac{1}{N} \sum_{i=1}^{N} rank_i
\end{equation}
where \( rank_i \) represents the rank of the correct triple corresponding to the \( i^{th} \) test sample, and \( N \) is the total number of test samples.

(2) MRR (Mean Reciprocal Rank): The average reciprocal rank of all correct triples, calculated as:
\begin{equation}
    MRR = \frac{1}{N} \sum_{i=1}^{N} \frac{1}{rank_i}
\end{equation}

(3) Hits@n: The proportion of correct triples ranked within the top \( n \), calculated as:
\begin{equation}
    Hits@n = \frac{1}{N} \sum_{i=1}^{N} I(rank_i \leq n)
\end{equation}
where \( I(rank_i \leq n) \) is an indicator function that equals 1 if \( rank_i \leq n \) is true. Typically, the value of \( n \) is always set as 1, 3 or 10.

Notably, in KGR model evaluation, a lower Mean Rank (MR), a higher Mean Reciprocal Rank (MRR), and a higher Hits@n indicate superior model performance. MR is sensitive to outliers, potentially compromising its robustness in assessing overall reasoning results. In contrast, MRR effectively mitigates this issue, making MRR and Hits@n more reliable and universal evaluation metrics for diverse KGR tasks. However, the presence of other correct candidate triples ranked before the correct prediction result can affect the reliability of reasoning results. To ensure evaluation validity, existing strategies filter reasoning results by removing candidate triples already existing in the KG. This practice enhances the reliability of the evaluation outcomes.

\section{Static Single-Step KGR}
\label{sec:single-step}

For static single-step KGR tasks, the core requirement is the ability to learn the direct semantic associations between entities from observed factual triples. Subsequently, given two elements of a factual triple, the correct result is inferred by evaluating the plausibility of candidate triples. As shown in Fig.~\ref{fig:KGRtask}, the relation between the entities $David$ and $U.S.A.$ is missing and requires to be inferred as $Nationality$. Existing static single-step KGR methods are primarily divided into two categories: KGE-based methods and symbolic rule-based methods. KGE-based models learn the topological features of the KG as embeddings of entities and relations, and then identify the relation $Nationality$ that best satisfies the constraints for the embeddings of the entities $David$ and $U.S.A.$ as the reasoning result. Logic rule-based approaches first automatically mine symbolic rules based on the frequent patterns of triples in the KG, and then instantiate the rules using queries and triples that match the rules. For instance, the observed factual triples $(David, BornIn, San\ Francisco)$ and $(San\ Francisco, LocatedIn, U.S.A.)$ in Fig.~\ref{fig:KGRtask} could match the logic rule $Nationality(x,y) \Leftarrow BornIn(x,z)$ $\wedge LocatedIn$ $(z, y)$. Consequently, the rule is instantiated as $Nationality(David,U.S.A.) \Leftarrow BornIn(David,$ $ California) \wedge LocatedIn(California, U.S.A.)$. KGE-based and logic rule-based methods represent the two major genres of artificial intelligence, connectionism and symbolism, respectively.

\subsection{KGE-based Model}

For static single-step KGR tasks, the essential requirement is the ability to learn direct semantic associations between entities from known factual triples. Given any two elements of a triple, the correct inference is made by evaluating the likelihood of all the candidate triples. This section reviews and introduces various models and open-source libraries of the KGE technique that (1) rely solely on factual triples, (2) incorporate ontology information, (3) integrate path information as well as (4) some negative sampling strategies for training KGE models.

\subsubsection{KGE Model with Triples}
\label{sec:KGEwtriple}

\begin{figure*}
    \centering
    \includegraphics[width=1\linewidth]{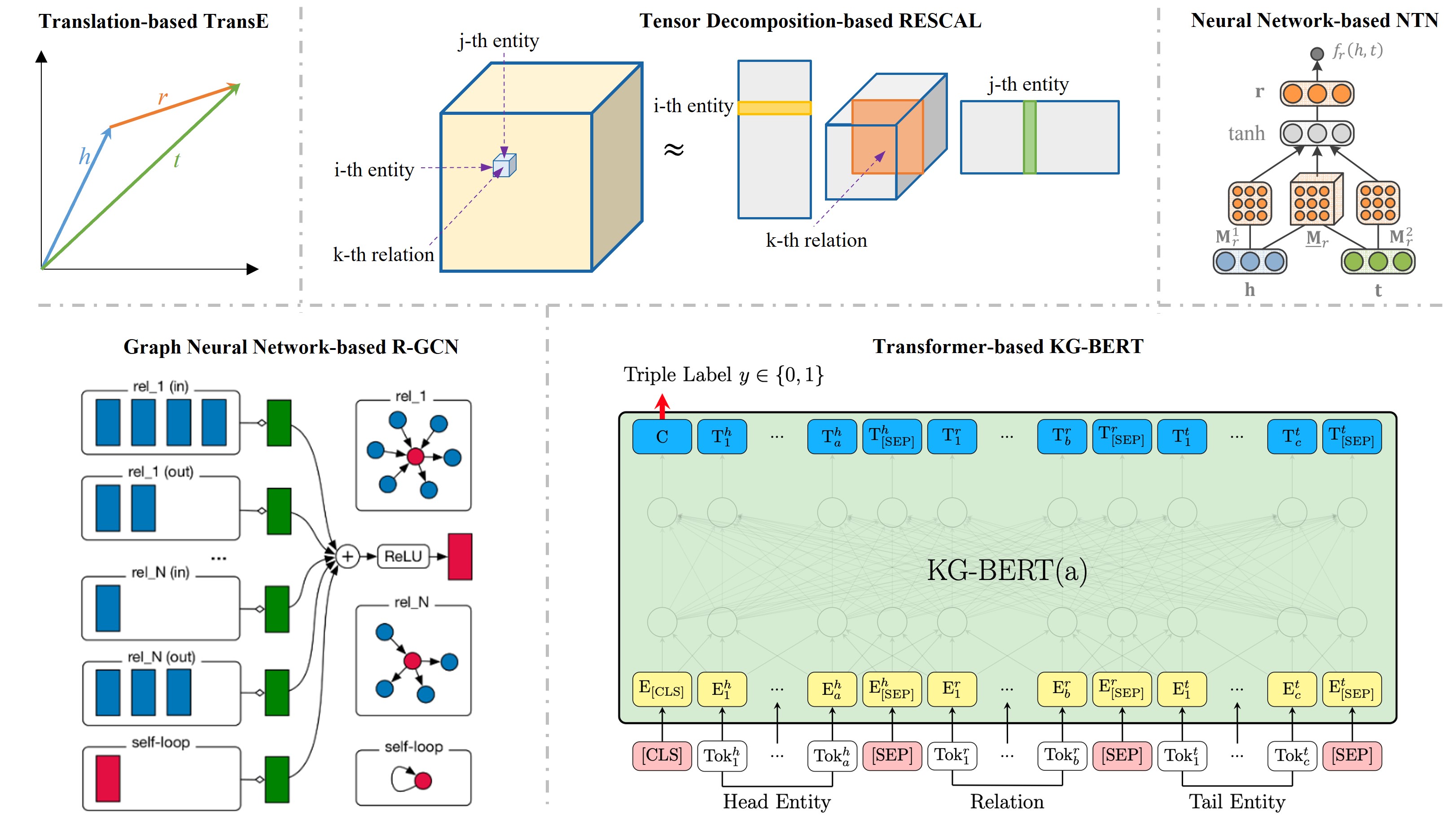}
    \caption{The illustration of five representative KGE models. The framework diagrams of NTN~\cite{67}, R-GCN~\cite{100} and KG-BERT~\cite{101} are directly derived from their original papers.}
    \label{fig:KGE}
\end{figure*}

KGE models can be primarily classified based on how each triple is modeled according to the interaction between entities and relations. According to the way each triple models the interaction between entities and relations, KGR models are mainly divided into several categories: (a) translation-based models, tensor decomposition-based models, neural networks-based models, graph neural networks-based models, and Transformer-based models. Fig.~\ref{fig:KGE} presents a typical representative from each category. Particularly, the core of these KGE models is the selection of an embedding space, the design of encoding strategies for entities and relations, and the formulation of a scoring function that quantifies the plausibility of a given triple.

(a) \textbf{Translation-based Model}:
TransE~\cite{69} is one of the most seminal models in this category. Inspired by the word embedding method word2vec~\cite{102}, TransE represents relations as translation operations in the embedding space. As illustrated in Fig.~\ref{fig:KGE}, the model conceptualizes a relation as the vector difference between the head and tail entities. In terms of a correct triple $(h, r, t)$, the optimization objective can be formulated as:
\begin{equation}
    \bf{h} + \bf{r} \approx \bf{t}
\end{equation}
where $\textbf{h}$, $\textbf{r}$ and $\textbf{t}$ denote the embeddings of the head entity $h$, relation $r$, and tail entity $t$ in a real-valued Euclidean space. The corresponding scoring function is defined as:
\begin{equation}
    E(h, r, t)=\Vert \bf{h} + \bf{r} - \bf{t} \Vert
\end{equation}
where $\Vert \cdot \Vert$ represents either the L1 or L2 norm.

\begin{figure*}
    \centering
    \includegraphics[width=1\linewidth]{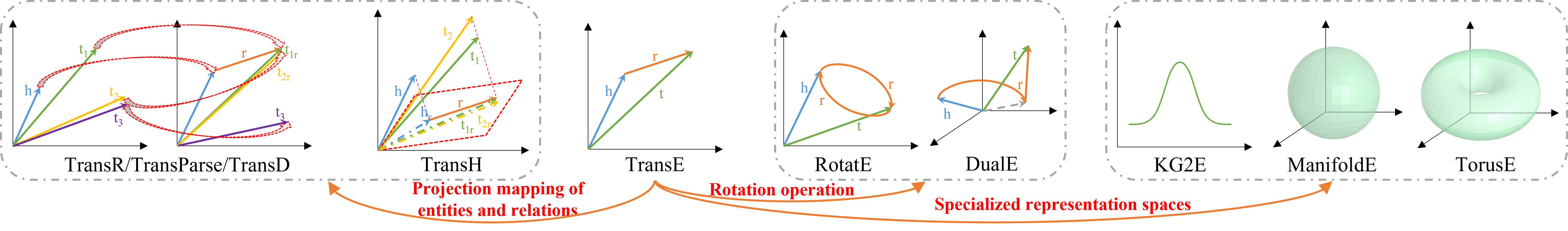}
    \caption{The illustration of TransE and its variants.}
    \label{fig:TransE}
\end{figure*}

Although TransE’s simplicity and efficiency have garnered considerable attention in the KGE community, its embedding into a real-valued Euclidean space inherently limits its representation power. For instance, in the case of one-to-many relations, the optimization objective fails to differentiate among multiple tail entities associated with a single head entity, thereby impeding the effective modeling of one-to-many, many-to-one, and many-to-many relations. Furthermore, the translation mechanism does not capture symmetric relations. To address these limitations, several extensions of TransE as shown in Fig.~\ref{fig:TransE} have been proposed, which can be broadly categorized into three approaches: (a) projection mapping of entities and relations~\cite{72}, (b) employing specialized representation spaces~\cite{52}, and (c) extending the translation operation to a rotation operation~\cite{103}.

TransH~\cite{72} and TransR~\cite{104} extend TransE by assigning each relation its own hyperplane or projection space, onto which entities are projected. This mechanism allows entities to adaptively learn relation-specific embeddings, thereby mitigating issues in modeling complex relations such as one-to-many and many-to-many. However, since the hyperplanes or projection spaces in TransH and TransR are solely relation-dependent, they do not account for the distinct semantic contributions of head and tail entities during projection. To address this issue, TransD~\cite{105} employs different projection matrices for the head and tail entities, while TranSparse~\cite{106} measures the sparsity of relation connected entity pairs to design an adaptive sparse transfer matrix for projection. PairE~\cite{107} learns pairwise embeddings for each relation, which are then used to project the head and tail entities separately before computing their distance. For TransA~\cite{108}, each relation is modeled as a symmetric non-negative matrix and an adaptive Mahalanobis distance is used in the scoring function.

KG2E~\cite{109} introduces uncertainty into the embedding space by modeling entites and relations as multi-dimensional Gaussian distributions, where the mean vector captures the definitive position of an entity or relation, while the covariance matrix reflects its uncertainty. ManifoldE~\cite{110} represents entities and relations on a manifold, modeling a triple as the tail entity residing on the surface of a hypersphere centered at the head entity, with the relation vector defining the radius. Additionally, TorusE~\cite{111} embeds the KG onto a compact Lie group, learning embeddings under a TransE-like optimization objective. To model hierarchical relations among entities, one approach embeds the KG into a hyperbolic space. Poincaré~\cite{112} was among the first to achieve this by mapping entities into the Poincaré ball, where entities closer to the center represent higher-level concepts, while those near the boundary represent lower-level ones. MuRP~\cite{113} further refines this idea by introducing Mobius matrix–vector multiplication and addition to learn relation-aware entity projections in hyperbolic space. Alternatively, hierarchy can be modeled using polar coordinates. Specifically, HAKE~\cite{114} employs the modulus to indicate hierarchical levels and the phase angle to differentiate entities within the same level. Moreover, H2E~\cite{115} and HBE~\cite{116} combine hyperbolic geometry with polar coordinates to capture hierarchical structures more comprehensively.

RotatE~\cite{117} employs the Hadamard product to model relations as rotation operations in the complex plane. This approach can effectively handle symmetric relations, as it imposes the constraint $\bf{r} \circ \bf{r} = 1$ for symmetric relations, which cannnot be addressed by translation-based models like TransE. While RotatE restricts operations to rotations on a hyperplane, QuatE~\cite{118} represents entities and relations in a quaternion space and leverages Hamiltonian multiplication to associate the embeddings of the head entity, relation, and tail entity, thereby enabling spatial rotations in a higher-dimensional space. Furthermore, DualE~\cite{119} models relations as exhibiting both translation and rotation characteristics in a dual quaternion space, offering enhanced representational capabilities compared to QuatE.

(b) \textbf{Tensor Decomposition-based Model}:
RESCAL~\cite{120} employs tensor decomposition by representing the entire KG as a large-scale third-order tensor, where each entry indicates whether the corresponding triple exists. By approximating these tensor entries via matrix multiplication between low-dimensional entity vectors and relation matrices, RESCAL effectively addresses reasoning tasks involving both symmetric and antisymmetric relations. DistMult~\cite{121} simplifies this approach by representing each relation as a diagonal matrix, thereby reducing computational complexity. ComplEx~\cite{122} is the first model to embed entities and relations into the complex vector space, using the Hamiltonian product to perform tensor decomposition among the complex vectors of the head entity, relation, and tail entity. HolE~\cite{123} combines the expressive power of tensor decomposition with the simplicity and efficiency of translation-based methods, modeling the relationship between entity pairs via the circular correlation of their vectors.

(c) \textbf{Neural Network-based Model}:
SME~\cite{124} is among the earliest approaches to employ neural networks for KGE. In SME, entities and relations in a triple are first mapped to vector embeddings at the input layer. The head and tail entity vectors are then combined with the relation vector in the hidden layer, and the triple score is computed as the dot product of the resulting hidden states corresponding to the head and tail entities. NTN~\cite{67} feeds the head and tail entity vectors into a relation-specific neural network, deriving the triple score through bilinear operations followed by a nonlinear activation function. NAM~\cite{125} concatenates the head entity and relation embeddings as input to a neural network and subsequently fuses the output of the final hidden layer with the tail entity representation to compute the triple score. Inspired by techniques from computer vision, ConvE~\cite{126} and ConvKB~\cite{127} reshape the embedded embeddings of entities and relations into two-dimensional matrices and utilize convolutional kernels to capture the interactions between entities and relations.

(d) \textbf{Graph Neural Network-based Model}:
Considering that KGs are inherently structured as graphs, graph neural networks (GNNs) are naturally well-suited for modeling their topological properties, they have been widely used to encode the neighborhood context of entities for representation learning. R-GCN~\cite{100} pioneered this approach by leveraging graph convolutional networks (GCNs) to aggregate information from neighboring entities connected via the same relation to update entity embeddings. In relatively dense graphs, this approach effectively captures topological nuances in a single pass, which could enhance entity embeddings. SACN~\cite{129} further extends this paradigm by employing a weighted GCN to extract structural features and learn matrix embeddings for entities and relations. In its decoder, SACN applies convolutional operations (inspired by ConvE) on the learned matrices and uses a TransE-style scoring function to evaluate triple plausibility. KBGAT~\cite{130} introduces multi-head attention within the GCN framework, enabling the model to focus on more salient information from an entity’s neighborhood when representing entities and relations. Similarly, KE-GCN~\cite{131} integrates conventional graph convolution operations with KGE techniques to iteratively refine entity and relation embeddings.

(e) \textbf{Transformer-based Model}:
In recent years, Transformer has achieved remarkable success in both natural language processing and computer vision tasks~\cite{132}. Some KGE approaches now utilize Transformers to incorporate contextual information into graph embeddings. For instance, R-MeN~\cite{133} employs the self-attention mechanism to design a relation memory network that facilitates interaction between triple encodings and memory embeddings, thereby capturing latent dependencies among triples. Both CoKE~\cite{134} and HittER~\cite{135} use Transformers to encode the contextual information within triples, transform the KGR task into a cloze test task for entities. GenKGC~\cite{136} transforms KGR into a sequence-to-sequence generation task by leveraging pre-trained language models. Similarly, iHT~\cite{137} pre-trains a Transformer-based language model on large-scale KGs and fine-tunes it on domain-specific graphs, thus achieving robust transferability. SimKGC~\cite{138} demonstrates superior reasoning performance by combining pre-trained language models with contrastive learning techniques. In addition, KG-BERT~\cite{101} and StAR~\cite{139} utilize pre-trained language models to jointly encode entities, relations, and their associated textual descriptions while preserving the intrinsic structure of the KG. Relphormer~\cite{140} proposes a novel Transformer variant tailored to the topological characteristics of KGs by dynamically sampling contextual subgraph sequences as input, and then conducting reasoning with a pre-trained language model. KoPA~\cite{141} introduces a new paradigm for KGR that leverages LLMs by designing knowledge prefix adapters, which convert structural embeddings from traditional KGE into text embeddings, thereby bridging the gap between structured and unstructured modalities. Finally, KICGPT~\cite{142} integrates a large language model (LLM) with a triple retriever, addressing data sparsity without incurring additional training overhead. It employs a knowledge-prompted in-context learning strategy that encodes structured knowledge into exemplars to effectively guide the reasoning process of the LLM.

\subsubsection{Ontology-Enhanced KGE Model}

\begin{figure}
    \centering
    \includegraphics[width=1\linewidth]{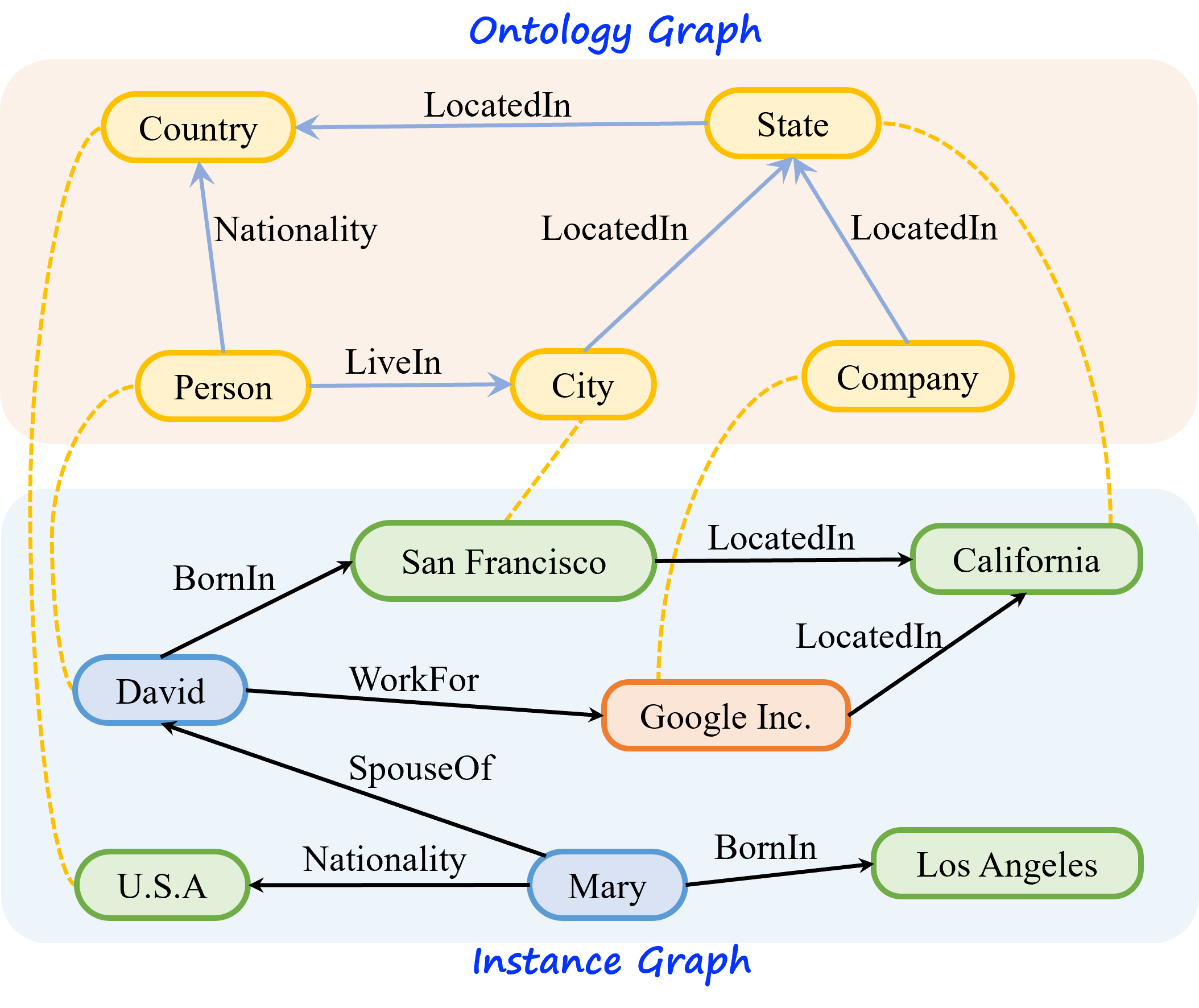}
    \caption{The illustration of ontology graph and instance graph.}
    \label{fig:ontology}
\end{figure}

Traditional KGE‐based KGR models typically consider only the fact triples present in the graph. However, in practice, constructing a KG often involves establishing an ontology that captures hierarchical concept information and inter-concept relationships. Fig.~\ref{fig:ontology} illustrates a KG comprising both an ontology layer and an instance layer, where links exist between ontology concepts and instance entities. For instance, the entity $California$ is linked to the concept $State$. Specifically, the ontology graph contains semantic associations among concepts such as $(Person, Nationality, Country)$.

To fully exploit this high-level semantic and ontological information, two principal types of KGE models have been developed to incorporate ontology: ontology-based models and entity type-based models.

(a) \textbf{Ontology Graph-based Model}:
One straightforward approach is to treat the relation $instance of$ between concepts and instances as an additional relation within the instance graph, thereby generating new triples to augment the training set~\cite{143}. However, such approaches tend to treat concepts and entities equivalently, which may prevent them from fully leveraging the high-level semantic richness embedded in the concepts. JOIE~\cite{97} addresses this limitation by jointly learning embeddings for both the ontology and instance layers, thus harnessing the hierarchical nature of concepts and their interrelations. CISS~\cite{144} further considers concept inheritance and structural similarity between these graphs, utilizing class sets composed of concepts at the same level to learn fine-grained concept representations and constructs a virtual entity layer view from clustered entities, subsequently comparing the subgraph representations of the ontology and instance layers to enhance their structural congruence. In another approach, Wang et al.~\cite{145} incorporate ontology information such as concept hierarchies and attribute constraints directly as constraints during the learning of entity and relation embeddings. Concept2Box~\cite{146} employs box embeddings to capture overlapping characteristics among ontology concepts and represents hierarchical relationships via the geometric properties (e.g., the area) of rectangular boxes. Finally, CAKE~\cite{147} automatically constructs a commonsense graph from both ontology and instance layers, and it devises a two-view reasoning algorithm that refines the reasoning process from a coarse to a fine level.

(b) \textbf{Entity Type-based Model}:
An alternative strategy treats ontology concepts as entity types to guide the learning of entity embeddings, while some methods directly learn separate embeddings for entity types. For instance, SSE~\cite{148} constrains entities of the same type to cluster together in the embedding space. Krompaß et al.~\cite{15} exploit the fact that certain entity types typically appear only in the head or tail position, thereby generating higher-quality negative samples for embedding learning. However, SSE presumes that each entity possesses a single type, an assumption that does not hold in most real-world KGs where entities are associated with multiple types. To capture the hierarchical structure and multi-typed nature of entities, TKRL~\cite{149} learns embeddings for all entity types at various levels, effectively converting a fact triple into associations among multiple type embeddings and yielding superior reasoning performance compared to methods that rely solely on fact triples. Similarly, TransET~\cite{150} utilizes a projection mechanism to map entity embeddings into corresponding type embeddings, thereby enhancing entity embeddings with type information. Nevertheless, neither TKRL nor TransET can dynamically adjust the type embeddings of an entity in different relational contexts. To overcome this limitation, AutoETER~\cite{151} automatically learns commonsense triple embeddings for each fact, adaptively deriving entity type embeddings that align with the current relational semantics and calculating triple scores from both factual and commonsense perspectives, thus improving reasoning accuracy.

It is worth noting that there is minimal overlap between the relations in the ontology graph and those in the instance graph, ontology information primarily serves to enrich entity embeddings rather than to directly support reasoning. Furthermore, not every KG is endowed with comprehensive entity type information. For instance, in automatically or semi-automatically constructed KGs such as NELL, entity types may merely indicate the domain of the entity, and some KGs such as WordNet, even lack any entity type annotations. In such cases, ontology information is hard to be leveraged for KGE models.

\subsubsection{Path-Enhanced KGE Model}

The path-enhanced KGE models can be classified into the following three types according to the characteristics of paths.

(a) \textbf{Relational Path-based Model}:
The existing approaches Path-RNN~\cite{152}, PTransE~\cite{98}, and PRN~\cite{153} integrate the embeddings of all relations in a path using recurrent neural networks (RNNs), addition or multiplication operations to obtain path embedding and introduce it into the KGE model based on TransE for overall model optimization. To enhance the temporal order characteristics of relations when representing paths, OPTransE~\cite{154} projects the head and tail entities of each relation into different spaces and introduces a sequence matrix to maintain the order of relations in the path. Since different relations in a real-world relation path may have varying degrees of semantic importance for expressing the relationship between entities, both TransE\&RW~\cite{155} and HARPA~\cite{156} employ hierarchical attention networks to select important relations in the path and learn path embeddings. To improve the accuracy and explainability of path embeddings, RPJE~\cite{157} introduces symbolic rules for interpretable path embedding, enhancing the ability to represent paths and thus improving the performance of KGR.

(b) \textbf{Complete Path-based Model}: 
The aforementioned methods consider only the relations in the path and lack entity information, which makes it difficult to accurately represent the semantics of the entire path. To integrate entities and relations in the path and learn complete path embeddings, PARL~\cite{158} composes the entities and relations in a path into a sequence and inputs them into an RNN for path encoding. However, the heterogeneity between entities and relations reduces the effectiveness of path embedding. To address the heterogeneity between entities and relations, RPE~\cite{159} maps entities into both relation and path spaces, reducing the heterogeneity between entities and relations and representing more complete paths. Both Das et al.~\cite{160} and Jiang et al.~\cite{161} represent each entity in a path by taking the average of all its type embeddings, using entity types to reduce the heterogeneity between entities and relations, and combine the path relations to learn path embeddings.

(c) \textbf{Multiple Path-based Model}: 
Considering that there are multiple paths between entities, multiple path-based models encode multi-path embeddings from the perspective of selecting effective paths and perform reasoning. CPConvKE~\cite{162} designs a gate-based path embedding method to filter noisy paths, ensuring that path embeddings are highly relevant to the semantic relationships between entities. PaSKoGE~\cite{163} encodes the correlations between relations and paths between entity pairs, adaptively determining the size of the margin parameter in the loss function for each path. Jagvaral et al.~\cite{164} use convolutional operations and BiLSTM to encode paths and employ attention mechanisms to capture the semantic correlations between candidate relations and each path, thereby performing a weighted fusion of the embeddings of multiple paths. PathCon~\cite{165} uses a relation message passing algorithm to iteratively pass relation messages in paths between entities to aggregate semantic information in the path and uses attention mechanisms to selectively aggregate the embeddings of different paths, thereby achieving relation prediction through path embeddings. PTrustE~\cite{166} constructs a path reliability network based on correlations to learn global and local features in paths, and then integrates all features of the path using bidirectional gated recursive units to obtain a path score matrix and path reliability. TAPR~\cite{167} designs a path-level attention mechanism to assign different weights to different paths and then fuses the weighted path embeddings to predict the semantic relationships between entities. Niu et al.~\cite{168} develop a rule and data co-driven path embedding scheme to represent each path between entities, then integrate the embeddings of multiple paths and utilize a bidirectional reasoning mechanism to achieve multi-step reasoning in KGs.

\subsubsection{Negative Sampling for KGE}

KGR operates under the open-world assumption~\cite{169}, meaning that triples not present in the KG may be either incorrect or correct but missing. Consequently, the training of KGE models cannot strictly assume that non-existent triples are definitely incorrect. To enable the model to evaluate the plausibility of triples, a pairwise loss function with positive and negative triples as input is typically employed. However, since no negative triples exist in the KG, negative sampling strategies are necessary to construct them. Generally, negative triples are constructed by replacing one entity or relation in a positive triple with another, following the local closed-world assumption~\cite{170}, to generate a triple not present in the KG and treating it as a negative triple. The quality of negative triples and the efficiency of negative sampling significantly impact the training effect of KGE models. Thus, in recent years, negative sampling techniques for KGE have become one of the important research directions in the field of KGR~\cite{171}.

\begin{figure}
    \centering
    \includegraphics[width=1\linewidth]{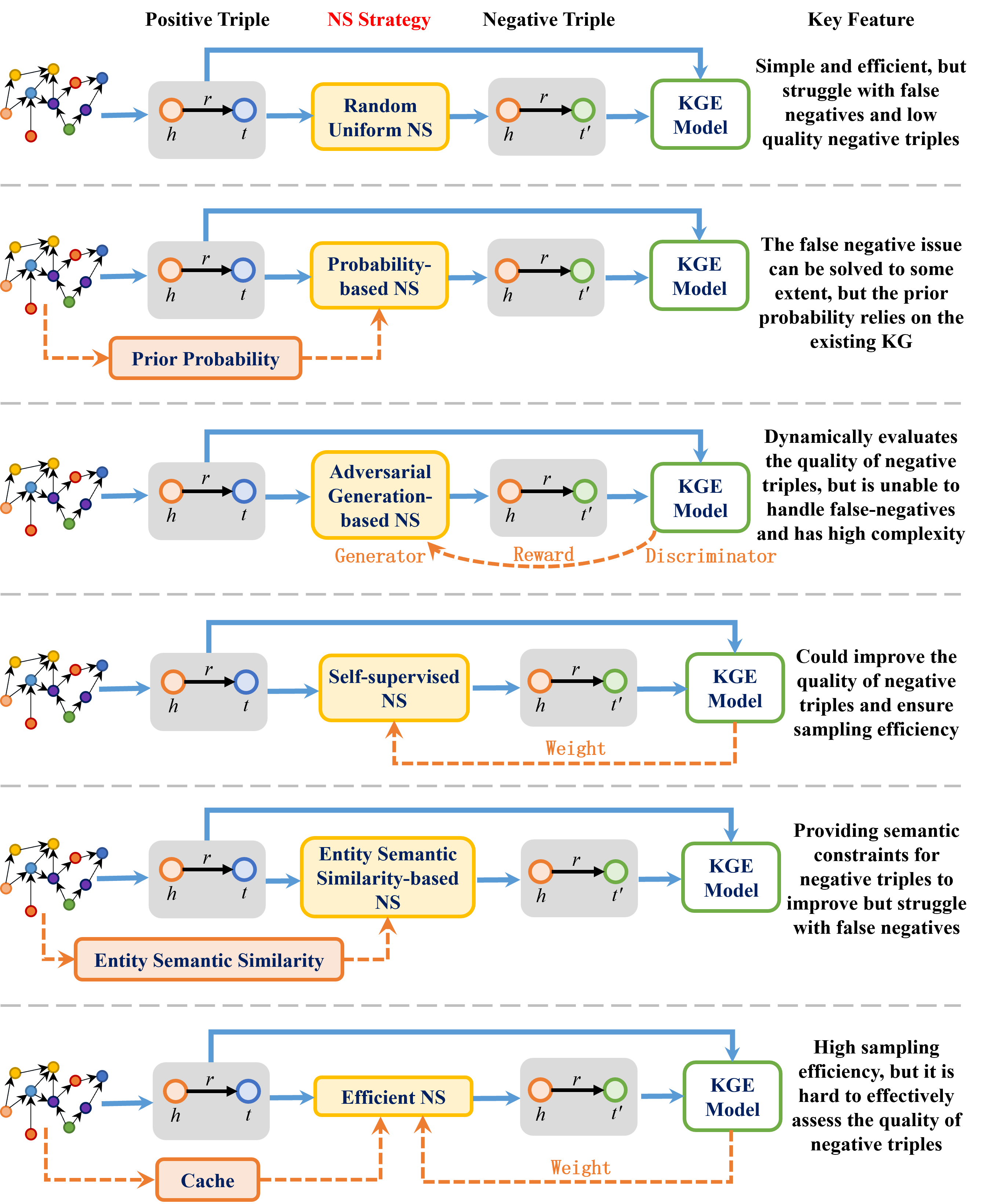}
    \caption{The illustration of six types of negative sampling (NS) strategies.}
    \label{fig:NS}
\end{figure}

As shown in Fig.~\ref{fig:NS}, existing negative sampling strategies for KGE can be categorized into the following six types.

(a) \textbf{Random Uniform Negative Sampling}: 
For KGE models, the simplest negative sampling mechanism involves randomly replacing one entity or relation in a positive triple to generate negative triples, with the replacement probability following a uniform distribution~\cite{72}. To enhance sampling efficiency, the Batch NS method~\cite{172} performs negative sampling within a batch. However, due to the lack of quality assessment for negative triples, the generated negative triples may either have excessive semantic differences from positive triples, potentially causing gradient disappearance during training and limiting learning effectiveness, or be correct triples missing in the KG (false negatives), introducing noise into the learning process.

(b) \textbf{Probability-based Negative Sampling}: 
Bernoulli negative sampling~\cite{173} employs a sparse attention mechanism to discover relation concepts and replaces the head or tail entity of a positive triple with different probabilities. The replacement probability for the head entity is calculated based on the average number of tail entities associated with each head entity. Zhang et al.~\cite{174} extend this manner by incorporating relation replacement alongside entity replacement and using the Bernoulli mechanism for probability calculation. SparseNSG~\cite{175} generates replacement probabilities by analyzing relation-entity association frequencies and also estimates the likelihood of generating false negatives to measure the quality of triple. Compared to random uniform sampling, probability-based methods use prior probabilities from the KG to reduce false negatives but rely heavily on existing knowledge, limiting robustness.

(c) \textbf{Adversarial Generation-based Negative Sampling}: 
KBGAN~\cite{99} integrates KGE into the adversarial training framework (GAN)~\cite{176} to select high-quality negative triples beneficial for learning. Subsequent adversarial-based approaches, such as IGAN~\cite{177}, GraphGAN~\cite{178}, KSGAN~\cite{179}, and RUGA~\cite{180}, are designed based on KBGAN with improvements like different generators or additional steps. These methods dynamically adjust the sampling probability distribution, effectively assessing the quality of triple. However, false-negative triples are more likely to be produced based on the assumption that higher-ranked candidate triples are of better quality may still produce false negatives, and the increased model complexity due to the GAN framework is a drawback.

(d) \textbf{Self-supervised Negative Sampling}: 
The Self-Adv method~\cite{117}, akin to KBGAN, assigns higher weights to quality negative triples but uses the scoring function of the KGE model directly to assess constructed triples, employing self-adversarial sampling. This approach avoids complex generators, enhancing sampling efficiency. LAS~\cite{181} prioritizes triples more likely to be incorrect, reducing false negatives. ASA~\cite{182} focuses on medium-difficulty triples to balance learning, while CAKE~\cite{147} leverages commonsense knowledge for sampling. Self-supervised methods assess the quality of triple internally, addressing false negatives and low quality, with notable efficiency and comprehensive effects.

(e) \textbf{Entity Semantic Similarity-based Negative Sampling}: 
To mitigate low-quality triples, replacing entities in positive triples with semantically similar ones is a straightforward idea for achieving negative triples with higher quality. The previous entity semantic similarity-based NS models design several strategies as followings. AN~\cite{183} and EANS~\cite{184} use K-means clustering to group similar entities, replacing originals with cluster members. Truncated NS~\cite{185} leverages structural and attribute embeddings, DNS~\cite{186} uses same-type entities, ESNS~\cite{187} incorporates entity context, while RCWC~\cite{188} and conditional constraint sampling~\cite{189} utilize relation domain/range restrictions to measure the similarity among entities and generate negative triples. LEMON~\cite{190} employs pre-trained language models for entity description encoding, while DHNS\cite{dhns} uses diffusion models in multi-modal entity information. These methods provide semantic constraints using auxiliary information but cannot fully eliminate false negatives.

(f) \textbf{Efficient Negative Sampling}: 
NSCaching~\cite{191} uses a caching mechanism for negative triple candidates. MDNCaching~\cite{192} generates semantically invalid but structurally similar negative triples using a KGE model to evaluate triple likelihood, addressing false negatives. TuckerDNCaching~\cite{193} extends MDNCaching with Tucker decomposition for more accurate likelihood assessment. CCS~\cite{194} introduces clustering caching based on entity similarity, and NS-KGE~\cite{195} converts the loss function to squared loss to reduce negative triple dependence. While efficient methods enhance sampling speed, they often struggle to effectively assess the quality of triple, limiting learning performance.

\subsubsection{Open-Source Library for KGE}

KGE has emerged as a pivotal research focus within the KGR domain in recent years, prompting the development of numerous open-source libraries dedicated to KGE models. This paper select the most commonly used and highly popular open-source KGE libraries available on GitHub, each boasting over 100 stars, with the majority achieving even greater popularity of more than 1000 stars, as detailed in TABLE~\ref{tab:kgegithub}.

\begin{table*}[ht]
    \centering
    \caption{Popular Open-Source Libraries for KGE}
    \label{tab:kgegithub}
    \renewcommand{\arraystretch}{1.2}
    \begin{tabular}{lp{2.0cm}p{7cm}p{5cm}}
        \toprule
        \textbf{Library} & \textbf{Implementation} & \textbf{Key Features} & \textbf{GitHub Repository} \\
        \midrule
        OpenKE\cite{openke} & Pytorch, TensorFlow, C++ & Efficiently implements fundamental operations such as data loading, negative sampling, and performance evaluation using C++ for high performance. & \url{https://github.com/thunlp/OpenKE} \\
        \midrule
        AmpliGraph\cite{ampligraph} & TensorFlow & Provides a Keras-style API with improved efficiency over OpenKE. & \url{https://github.com/Accenture/AmpliGraph} \\
        \midrule
        torchKGE\cite{arm2020torchkge} & Pytorch & Achieves twice the efficiency of OpenKE and five times that of AmpliGraph. & \url{https://github.com/torchkge-team/torchkge} \\
        \midrule
        LibKGE\cite{libkge} & Pytorch & Enables direct configuration of hyperparameters and model settings via configuration files. & \url{https://github.com/uma-pi1/kge} \\
        \midrule
        KB2E\cite{104} & C++ & One of the earliest KGE libraries and the predecessor of OpenKE. & \url{https://github.com/thunlp/KB2E} \\
        \midrule
        scikit-kge\cite{123} & Python & Implements multiple classical KGE models and supports a novel negative sampling strategy. & \url{https://github.com/mnick/scikit-kge} \\
        \midrule
        NeuralKG\cite{NeuralKG} & Pytorch & Integrates KGE techniques with graph neural networks (GNNs) and rule-based reasoning models. & \url{https://github.com/zjukg/NeuralKG} \\
        \midrule
        PyKEEN\cite{PyKEEN} & Pytorch & Offers 37 datasets, 40 KGE models, 15 loss functions, 6 regularization mechanisms, and 3 negative sampling strategies. & \url{https://github.com/pykeen/pykeen} \\
        \midrule
        Pykg2vec\cite{PyKG2vec} & Pytorch, TensorFlow & Supports automated hyperparameter tuning, exports KG embeddings in TSV or RDF formats, and provides visualization for performance evaluation. & \url{https://github.com/Sujit-O/pykg2vec} \\
        \midrule
        $\mu$KG\cite{muKG} & Pytorch, TensorFlow & Supports multi-process execution and GPU-accelerated computation, making it well-suited for large-scale KGs. & \url{https://github.com/nju-websoft/muKG} \\
        \midrule
        DGL-KE\cite{DGL-KE} & Pytorch, MXNet & Optimized for execution on CPU and GPU clusters, offering high scalability for large-scale KGs. & \url{https://github.com/awslabs/dgl-ke} \\
        \midrule
        GraphVite\cite{GraphVite} & Pytorch & Provides efficient large-scale embedding learning, supports visualization of graph data, and enables multi-processing and GPU parallelization. & \url{https://github.com/DeepGraphLearning/graphvite} \\
        \midrule
        PBG\cite{pbg} & Pytorch & Designed for distributed training, capable of handling KGs with billions of entities and trillions of edges. & \url{https://github.com/facebookresearch/PyTorch-BigGraph} \\
        \bottomrule
    \end{tabular}
\end{table*}

In recent years, KGR models grounded in KGE have seen significant development. These methods transform the symbolic representations of KGs into numerical forms and utilize scoring functions to evaluate the likelihood of candidate triples. However, such reasoning approaches have limitations: they provide reasoning results without an interpretable process and struggle to make precise semantic judgments about candidate triples, thus constraining their overall performance.

\subsection{Logic Rule-based Model}
\label{sec:ruleKGR}

To preserve the inherent explainability of symbolic KGs and recognizing that human reasoning predominantly relies on empirical rules, a key research direction in KGR involves rule learning. Logic rule-based KGR capitalizes on the symbolic nature of knowledge, offering precision and explainability. As illustrated in Fig.~\ref{fig:KGRtask}, a logic rule could be mined from the KG: $Nationality(x,y) \Leftarrow BornIn(x,z)$ $\wedge LocatedIn$ $(z, y)$. Using this rule and two matching observed factual triples $(David, BornIn, San\ Francisco)$ and $(San\ Francisco, LocatedIn, U.S.A.)$, a new triple $(David, Nationality, U.S.A.)$ can be inferred. Generally, a rule typically consists of a rule head and a rule body, structured as: $Rule\ Head \Leftarrow Rule\ Body$, indicating that the knowledge in the rule head can be directly deduced from the rule body~\cite{196}. The rule head contains one atom, while the rule body comprises one or more atoms. An atom is a fact that includes variables, such as $BornIn(x, z)$ which signifies that $x$ was born in $z$. In the rule body, different atoms are combined through logical conjunction, and these atoms can be either positive or negative. Rules with only positive atoms in the rule body are called Horn rules. Given that KGs contain only positive facts, Horn rules are well-suited for KGR tasks.

To address the inefficiency of domain experts manually summarizing and inducing rules from large-scale KGs, certain rule learning methods can rapidly and effectively mine rules from KGs. Early rule learning methods treated the triples in KGs as facts described by binary predicates, enabling the use of inductive logic programming (ILP)~\cite{ILP} techniques to learn first-order logic rules from KGs, such as FOIL~\cite{198}, MDIE~\cite{199}, and Inspire~\cite{200}. However, ILP-based rule learning methods are designed for smaller datasets and are challenging to apply to large-scale KGs~\cite{201}.

\subsubsection{Rule Learning for KG}
\label{sec:rule}

On account of rule search or construction schemes, rule learning approaches for KGs fall into two categories: those grounded in inductive logic programming and those based on neural networks, representing the symbolism and connectionism, respectively.

(a) \textbf{Inductive Logic Programming}:
For large-scale KGs, efficient rule learning algorithms have been developed, including AMIE+~\cite{202}, ScaLeKB~\cite{203}, RDF2rules~\cite{204}, Swarm~\cite{205}, RuDiK~\cite{206}, and RuLES~\cite{207}. Notably, AMIE+ and ScaLeKB excel in automatically mining Horn rules from KGs with millions of entities. AMIE+ enhances the efficiency of rule learning through optimized query techniques and a refined rule quality evaluation algorithm, while ScaLeKB improves further via automatic data cleaning and partitioning strategies. Most current methods rely on exhaustive rule candidate searches and heuristic pruning, limiting efficiency and rule quality. To address this, Evoda~\cite{208} employs a genetic logic programming algorithm, extending Horn rules to generalized Datalog rules~\cite{209} and incorporating a scoring mechanism based on relational representation learning. Despite these advances, the computational efficiency of purely symbolic rule learning methods remains constrained, prompting the development of data-driven approaches to boost rule mining efficiency.

(b) \textbf{Neural Networks-based Model}:
NeuralLP~\cite{210} and DRUM~\cite{196} utilize one-hot entity representations and matrix-encoded relations, learning rule patterns end-to-end and assessing rule confidence. RLvLR~\cite{211} integrates entity and relation embeddings from KGE with a novel scoring function to prune and prioritize rules. RNNLogic~\cite{212} employs an RNN-based rule generator coupled with a reasoning predictor. RARL~\cite{213} efficiently traverses term facts (TBox), learning Horn rules by leveraging semantic relevance between relations in body and head atoms, and separates candidate rule generation from quality evaluation for scalability. The Transformer-based Ruleformer~\cite{214} treats rule mining as a sequence-to-sequence task, using an enhanced attention mechanism to encode head entity subgraph context and decode relation probability distributions, simultaneously scoring reasoning results. The performance of rule-based reasoning depends not only on rule quality and quantity but also on aggregation strategies when multiple rules match. Unlike previous methods focused on rule mining, Ott et al.~\cite{215} exploit the dependencies among rules, enhancing aggregation performance of multiples rules.

\subsubsection{Neural-Symbolic KGR}

To effectively combine the explainability of symbolic rules with the generalization capabilities of KGE, two primary approaches have emerged: open-loop rule-enhanced KGE methods and closed-loop iterative methods that integrate rule learning with KGE.

(a) \textbf{Rule-Enhanced KGE}:
Models such as KALE~\cite{68} and RUGE~\cite{74} incorporate axiomatized rules to capture various relational patterns. For instance, for the symmetric relation $hasFriend$ the corresponding rule is $hasFriend(y,x) \Leftarrow hasFriend(x,y)$. By instantiating variables in these rules with specific entities, new triples are generated at the rule head. These models utilize t-norm fuzzy logic~\cite{216} to compute the truth values of these newly generated triples, effectively expanding the KG and enhancing the performance of KGE, particularly in sparse KGs. However, the uncertain nature of these newly generated triples may not always ensure correct semantic associations between entities. Additionally, these methods primarily use rules for data augmentation without integrating them directly into the KGE and reasoning processes. RulE~\cite{217} addresses this by jointly embedding entities, relations and symbolic rules within a unified space, allowing for the calculation of confidence scores for each rule to ensure consistency with observed triples. However, these approaches often convert symbolic rules into numerical formulas, potentially compromising the original symbolic characteristics and explainability of the rules. To maintain the significant explainability of rules, RPJE~\cite{157} performs interpretable path semantic composition and models semantic associations between relations, thereby injecting the logical semantics of rules into KG embeddings. This approach allows for the effective use of rules during the reasoning process, enhancing both the performance of KGE and the explainability of the inferred results.

(b) \textbf{Iterative Integration of Rule Learning and KGE}:
To achieve a more sufficient integration of rule learning and KGE, models such as IterE~\cite{218}, UniKER~\cite{219}, and EngineKG~\cite{220} develop iterative mechanisms that alternate between these two processes. Specifically, IterE extends the methodologies of KALE and RUGE by combining symbolic rules with KG embeddings through t-norm fuzzy logic in each iteration. This process generates new triples, updates the KG, and facilitates the discovery of additional rules, thereby refining KGE in subsequent iterations. UniKER takes a similar approach by inferring new triples based on symbolic rules in each iteration to expand the KG, followed by KGE on the updated graph. It further enhances the graph through reasoning using KG embeddings and scoring functions. EngineKG is designed based on RPJE by integrating pre-extracted path information and relation embeddings to optimize the efficiency of rule learning, constructing a closed-loop system where KGE and rule learning complement each other. This iterative process continually improves the effectiveness of KGE and facilitates the discovery of more comprehensive rules.

\begin{figure*}
    \centering
    \includegraphics[width=1\linewidth]{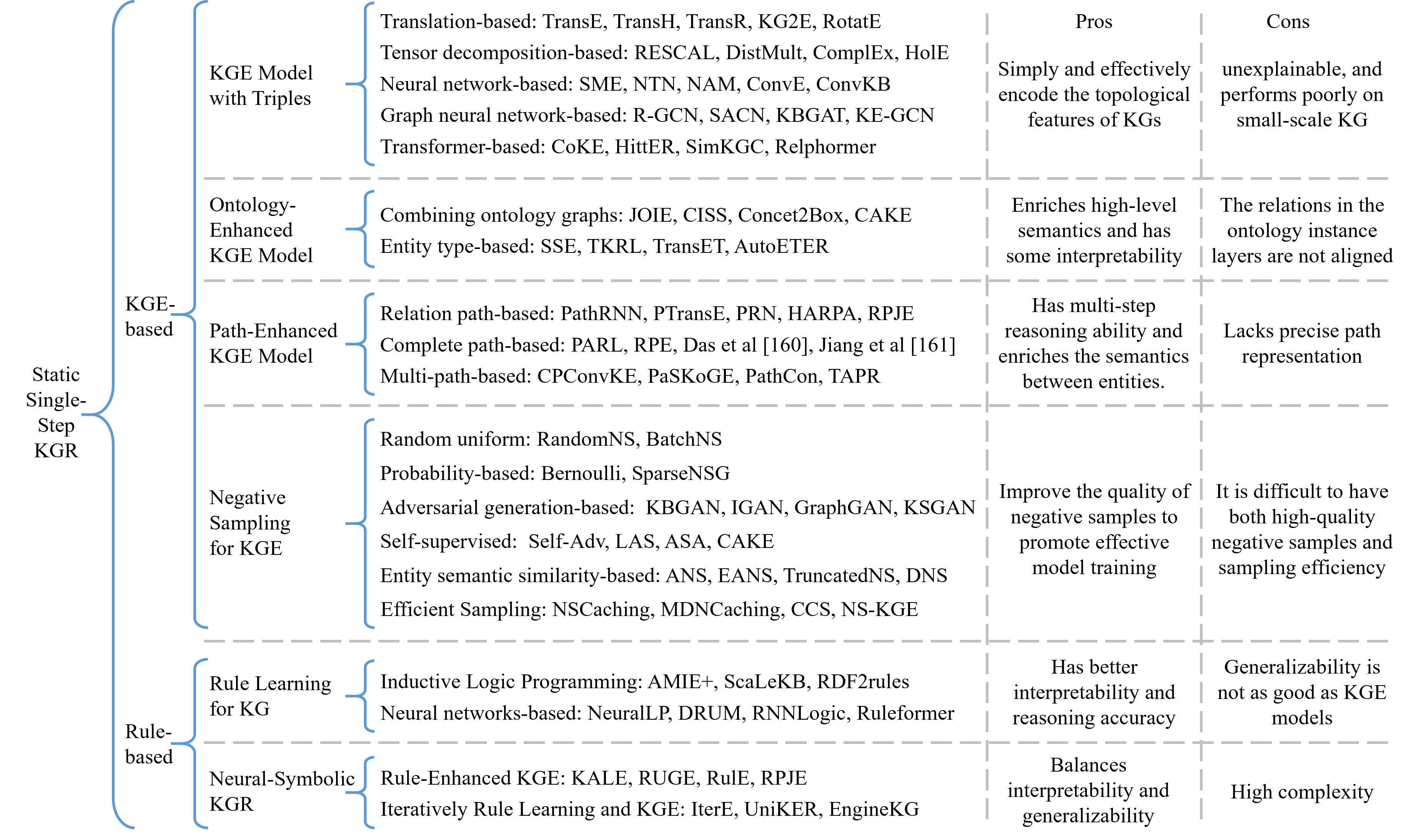}
    \caption{Taxonomy of static single-step KGR approaches with the comparison of their advantages and disadvantages.}
    \label{fig:tax_singlestep}
\end{figure*}

Fig.~\ref{fig:tax_singlestep} provides a systematic overview of representative models for static single-step KG reasoning tasks that integrate KGE with logic rules, highlighting the advantages and limitations of each approach.

\section{Static Multi-Step KGR}
\label{sec:multi-step}

As is well known, individuals often extract evidence chains from existing knowledge, regarding the conclusions inferred by these reliable chains as results. Similarly, in static multi-step KGR tasks, multi-hop paths within KGs are employed to identify effective routes from a head entity to address specific queries, as illustrated in Fig.~\ref{fig:KGRtask}. For instance, within a KG, a path such as $David \xrightarrow{BornIn} San\ Francisco \xrightarrow{LocatedIn} California \xrightarrow{LocatedIn} U.S.A.$ connects the entities $David$ and $U.S.A.$. By representing this path and leveraging it for reasoning, one can infer the implicit relation of 'Nationality' between $David$ and the $U.S.A.$.

\subsection{Random Walk-based Model}

Current approaches specific to static multi-step KGR predominantly encompass three categories: random walk-based models, reinforcement learning-based models, and LLM-based techniques. Random walk strategies, extensively applied in graph data analysis, initiate from a starting node and randomly sample subsequent nodes to construct a path. The Path Ranking Algorithm (PRA) proposed by Lao et al.~\cite{221} is pioneering in utilizing random walks to discover paths between two entities, treating the sequence of relations within these paths as features, which are then input into a classifier to predict the relation between the entities. Subsequent random walk-based multi-step KGR models are developed by extending PRA algorithm, primarily focusing on optimizing path generation and feature extraction.

\subsubsection{Enhanced Path Generation Model}

In the context of multi-step KGR tasks, enhancements in path generation methods have concentrated on refining the PRA algorithm's path search process to explore more effective path information that represents semantic associations between entities. Lao et al. initially improve their PRA by introducing a relation inference algorithm based on constrained and weighted random walks~\cite{222}, which adjusts the weights of different paths generated by random walks to infer target relations. Subsequently, they utilize syntactic-semantic inference rules learned from large-scale web text corpora and KGs to propose a path-constrained random walk model~\cite{223}, employing these learned rules for inference. Gardner et al. leverage implicit syntactic features mined from extensive corpora, incorporating them to add implicit syntactic labels between entities, thereby enriching features and assisting the PRA algorithm in generating more diverse relation sequences as paths, ultimately enhancing multi-step reasoning task performance. CPRA~\cite{225} employs clustering methods to automatically discover highly related relations and adopts a multi-task learning strategy to exploit interactions between paths, improving relation prediction. C-PR~\cite{226} learns the global semantics of entities in KGs through word embeddings and introduces a selective path exploration strategy using bidirectional random walks to enumerate context-related paths, which not only improve reasoning performance but also enhance explainability. To address the inefficiencies in path search caused by random walks, A\*Net~\cite{227} integrates the heuristic A* search algorithm into the path search process, selecting the most pertinent entities and relations as actions to enhance training and reasoning efficiency.

\subsubsection{Advanced Feature Extraction Model}

SFE~\cite{228} introduces subgraph feature extraction into random-walk-based multi-step KGR, capturing local structural information to better understand associations between entities and relations, extracting more expressive features to predict relations between entities, and improving the efficiency of multi-step reasoning tasks. PathCon~\cite{165} combines relational context and relational paths between entity pairs mined by random walks to predict relations between entities. While random-walk-based methods offer explainability, the inherent path search mechanism leads to a vast search space, resulting in low efficiency in path mining and feature extraction. These methods lack the capability to automatically identify paths that are semantically closest to the target triples and rely on path features for predicting relations between entities, making it challenging to perform effective reasoning.

\subsection{Reinforcement Learning-based Model}

In multi-step KG reasoning tasks, reinforcement learning (RL)-based approaches aim to discover paths within the KG that correspond to an entity and a relation in a given query. Initiating from the head entity, RL-based methods utilize the current state to inform subsequent moves. This process is typically modeled as a Markov Decision Process (MDP), wherein deep RL techniques train an agent to navigate the graph. The agent selects pertinent entities and relations as actions to maximize a reward function, which is crafted to steer the search towards the correct tail entity~\cite{229}.

\subsubsection{RL-based Path Search}

DeepPath~\cite{75} pioneers the application of RL techniques to static multi-step KGs by defining relations as the action space and formulating a reward function based on the successful identification of the correct tail entity. However, DeepPath's state representation encompasses the tail entity, restricting its applicability to triple classification tasks where both head and tail entities are provided. To address this limitation, MINERVA~\cite{229} proposes an RL strategy that initiates the path search using only the head entity and the query relation. This approach employs a Long Short-Term Memory (LSTM) network to encode historical states and actions, coupled with a fully connected network to generate action probabilities. Besides, DIVA~\cite{230} introduces a variational inference framework to manage uncertainty within KGs, dynamically adjusting entity and relation embeddings in response to uncertainty.

\subsubsection{Enhanced Reward Functions and Action Selection}

Enhancing reward functions and action selection mechanisms is vital for the efficacy of RL-based methods. Initial approaches employed sparse binary rewards, which often led to slow or suboptimal learning processes. To mitigate this, MultiHopKG~\cite{231} introduces soft rewards by leveraging pre-trained KGE models such as ComplEx and ConvE, and implements action dropout technique to prevent convergence to local optimum. M-Walk~\cite{232} utilizes Recurrent Neural Networks (RNNs) to encode historical paths and incorporated Monte Carlo Tree Search to generate effective reasoning paths. RARL~\cite{233} integrates high-quality logical rules as prior knowledge to inform action selection, while AttnPath~\cite{234} combines LSTM networks with graph attention mechanisms to enrich state embeddings. Furthermore, DIVINE~\cite{235} employs a generative adversarial imitation learning framework to train an inference agent, utilizing a discriminator to autonomously adjust rewards.

While RL-based methods evade the extensive search space associated with the PRA algorithm, they encounter challenges such as cold start issues and necessitate effective interventions to facilitate efficient training. Moreover, these models may struggle in scenarios where no finite-length path exists between the query and target entities, thereby limiting their robustness.

\subsection{LLM-based Model}

To mitigate the cold start and robustness challenges inherent in RL-based multi-step KGR approaches, recent approaches have harnessed the capabilities of large language models (LLMs). These strategies position LLMs as agents to navigate KGs, thereby enhancing reasoning performance and addressing issues such as cold starts and missing answers~\cite{236}.

\subsubsection{Models without Fine-Tuning}

Contemporary LLM-based multi-step reasoning techniques predominantly employ a no-fine-tuning paradigm. For instance, StructGPT~\cite{237} introduces an iterative call-linearize-generation framework. This approach initially utilizes LLM interfaces to extract paths pertinent to reasoning tasks from KGs as evidence. These paths are then linearized into textual prompts, which LLMs process to generate outputs, iteratively selecting valuable paths until reaching the final results. Similarly, KSL~\cite{238} devises effective prompts that transform retrieval tasks into multi-hop decision sequences, enabling LLMs to perform zero-shot knowledge searches and produce comprehensive retrieval paths, thereby enhancing explainability. KD-CoT~\cite{239} formalizes the LLM-based multi-step reasoning process into a structured multi-turn question-answering (QA) format. In each iteration, LLMs interact with retrieval-based QA systems to generate reliable reasoning paths through precise answers. ToG~\cite{240} employs beam search algorithms on KGs, leveraging LLMs to uncover valuable paths and deliver the most probable reasoning outcomes. KnowledgeNavigator~\cite{241} guides the reasoning process by optimizing implicit constraints within queries, selectively gathering supporting information from KGs based on LLM insights and question requirements, and organizing this information into prompts to enhance LLM reasoning. Nguyen et al.~\cite{242} propose a discriminative and generative chain-of-thought (CoT) evaluation framework, assessing both the reasoning results and the intermediate steps generated by CoT prompts.

\subsubsection{Models Fine-Tuned with Instructions}

However, direct LLM-based path searches may encounter semantic misalignments between LLMs and specific KGs. To address this, another category of LLM-based multi-step reasoning methods involves fine-tuning LLMs with tailored instruction sets. KG-Agent~\cite{243} employs an iterative mechanism that autonomously selects tools for KGR and fine-tunes the base LLM using an instruction dataset. AgentTuning~\cite{244} constructs a lightweight instruction fine-tuning dataset, integrating open-source instructions from general domains to enhance LLM multi-step reasoning performance. Besides, GLaM~\cite{glam} introduces a fine-tuning framework that aligns LLMs with domain-specific KGs, transforming them into alternative text representations with labeled question-answer pairs. This grounding in specific graph-based knowledge expands the models' capacity for structure-based reasoning.

\begin{figure*}
    \centering
    \includegraphics[width=1\linewidth]{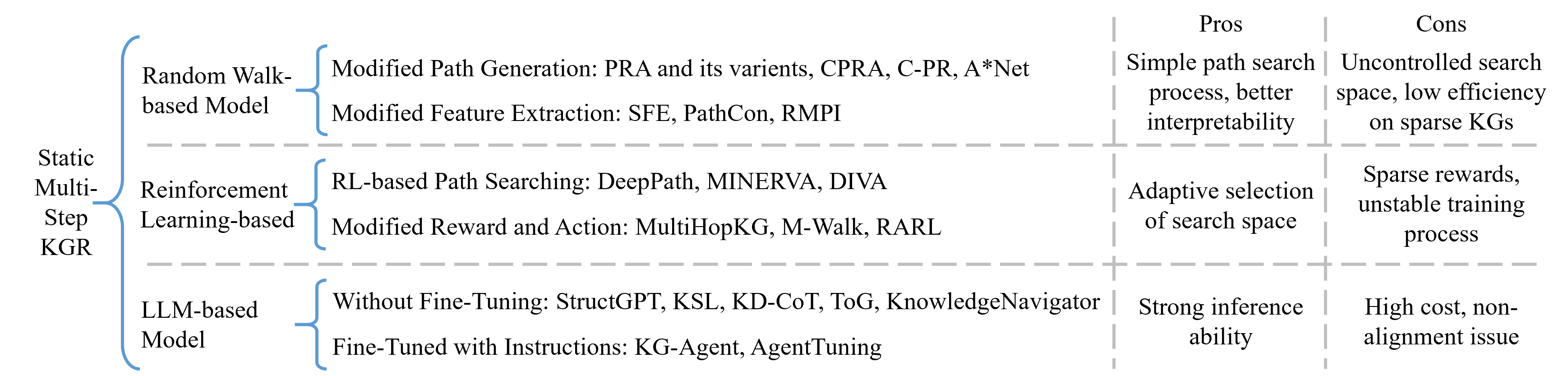}
    \caption{Taxonomy of static multi-step KGR approaches with the comparison of their advantages and disadvantages.}
    \label{fig:tax_multistep}
\end{figure*}

Fig.~\ref{fig:tax_multistep} illustrates representative models for each category of approaches for static multi-step KGR tasks, along with a summary of their respective advantages and disadvantages.

\section{Dynamic KGR}
\label{sec:dynamic}

\begin{figure*}
    \centering
    \includegraphics[width=1\linewidth]{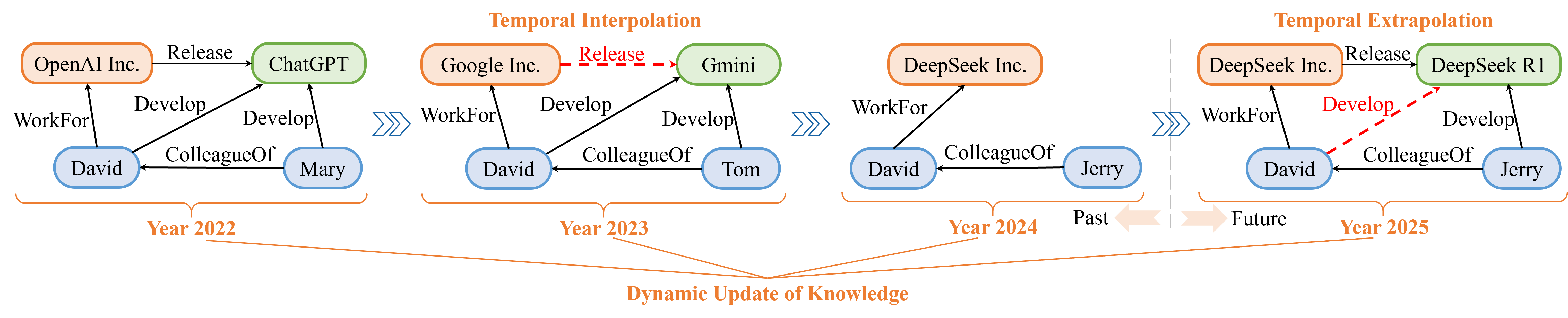}
    \caption{The illustration of the dynamic KGR task. This task implies three characteristics, namely dynamic update of knowledge, KGR for interpolation in the past timestamps and KGR for extrapolation in the future.}
    \label{fig:dynamicKGR}
\end{figure*}

In real-world application domains such as financial analytics, news monitoring, and public security systems, streaming data exhibits continuous temporal evolution. This dynamic nature necessitates the construction of dynamic KGs that support incremental updates through triple additions, modifications, and deletions~\cite{IncreKGE}. As illustrated in Fig.~\ref{fig:dynamicKGR}, these requirements have driven significant advances in incremental KGE models.

A critical sub-class of dynamic KGs incorporates explicit temporal annotations, namely temporal KGs (TKGs)~\cite{57}. In such structures, entity and relations are temporally scoped and all quadruples in TKGs are only applicable during specific timestamps or time intervals. Following the generalised way of describing events, a temporal quadruple that model the time-dependent event is formed as $(subject, predicate, object, timestamp/time\ interval)$ in TKGs.

Temporal KGR (TKGR) task focuses on predicting events through historical pattern analysis under temporal constraints~\cite{56}. The existing approaches categorize this task into two sub-tasks: (1) Temporal interpolation for timestamped prediction within observed sequences, typically addressed through time embedding-based techniques, and (2) Temporal extrapolation for future event forecasting, requiring evolution learning-based models that capture temporal associations among events over time.

\subsection{Incremental KGE Model}

\subsubsection{Multiple Spaces-based Model}

PuTransE~\cite{245} is among the earliest incremental KGE methods. It extends the traditional TransE by creating multiple embedding spaces for different segments of the KG. When the KG is updated, new embedding spaces are learned and subsequently aggregated with existing ones to predict the plausibility of a triple. Based on this idea, Liu et al.~\cite{246} partition the overall embedding space into subspaces according to relation types, thereby allowing updated triples to identify the most semantically relevant subspace for embedding. ABIE~\cite{247} employs $k$-shell decomposition from complex networks to identify key knowledge anchors within the KG. These anchors determine the embedding spaces, ensuring that only the updated entities and relations within the corresponding subspace are retrained, which improves training efficiency. However, the multi-space approach increases the overall space complexity and poses challenges in balancing new and old knowledge.

\subsubsection{Continual Learning-based Model}

CKGE~\cite{248} is a continual KGE model that incorporates incremental distillation mechanism. This approach employs a hierarchical strategy for learning new triples while retaining previous knowledge via distillation. LKGE~\cite{249} leverages a lifelong embedding learning strategy that utilizes a masked auto-encoder to update knowledge representations, and transfer learning is employed to inject learned knowledge into the embeddings of new entities and relations. AIR~\cite{250} is an adaptive incremental update framework, which measures the importance of triples to select those most affected by updates, and employs an embedding propagation mechanism to avoid full retraining. Existing methods have primarily focused on incorporating new knowledge following updates while neglecting the deletion of outdated information. TIE~\cite{251} addresses this by using deleted triples as negative triples and fine-tuning the model with new ones, effectively adapting to semantic changes in the KG. Although continual learning strategies can address issues related to parameter size and the balance between new and old knowledge, these methods tend to be more complex than multiple space-based approaches, causing increased training and inference costs.

\subsubsection{Space Update-based Model}

RotatH~\cite{252} updates embedding spaces via hyperplane rotations, ensuring that the KG embeddings preserve both timeliness and accuracy. An extension of this approach, MMRotatH~\cite{253}, employs multi-modal embedding techniques to handle previously unseen modalities in new entities, incrementally embedding them into the pre-trained space. These methods update embedding spaces using projection hyperplane mechanisms that preserve existing knowledge while adapting to new information. However, when updating entity and relation embeddings, they consider only the triples in which the entities appear, thereby lacking a full understanding of the current contextual semantics.

\subsubsection{Local Subgraph-based Model}

DKGE~\cite{254} introduces dual representations for each entity and relation, consisting of both knowledge embeddings and context element embeddings. It models entities, relations, and their contextual information using two attentional graph convolutional networks, a gating strategy, and translation operations. This approach confines the impact of KG updates to local subgraphs, enabling rapid online learning of updated embeddings. Navi~\cite{255} learns entity embeddings from local neighborhood information using an RGCN model, updating KG embeddings without relying on global graph structure. Similarly, UOKE~\cite{256} performs localized updates on modified triples by encoding subgraph information with an RGCN to update the embeddings of affected entities and relations. It uses gradient descent and regularization techniques to balance the integration of new and old knowledge. Xiao et al.~\cite{257} propose a temporal KG incremental construction model that captures the dynamic characteristics of entities and relations over time. By leveraging inter-entity relations and higher-order path information, this model enhances the understanding of entity-related contextual semantics and updates entity embeddings accordingly. Although these methods learn embeddings for new knowledge from updated neighborhood subgraphs, their effectiveness is limited in very sparse KGs due to insufficient contexts.

\subsection{Temporal KGR Model}

Temporal KGs (TKGs) extend static KGs by incorporating explicit time dimensions, making the modeling of events a critical challenge in temporal KGR. Existing approaches address this challenge by conforming to two principal strategies: (1) time embedding-based models for event prediction at known timestemps, and (2) evolution learning-based models for forecasting future events. The core objective is to effectively model time information, capturing both event timeliness and the temporal order of events.

\subsubsection{Time Embedding-based Model}

Time embedding-based approaches learn representations of time typically as vectors, tensors, or hyperplanes and combine these with entity and predicate embeddings to compute scores for event quadruples, thereby estimating the plausibility of these events. For instance, TA-TransE~\cite{80} treats time as a numerical attribute, combining time and predicate into time-aware predicate sequences. These sequences are encoded using LSTMs to capture temporal features, which are further introduced into a scoring function analogous to that of TransE~\cite{69} to evaluate quadruples. However, this model does not learn distinct time embeddings, and the reliance on LSTM-based feature extraction may compromise the precision of time representations. To address this issue, TTransE~\cite{258} extends TransE by modeling the interactions among event quadruples as translations that incorporate both predicates and time. However, it is unable to differentiate events involving the same subject occurring at different times, thereby limiting its temporal expressiveness. In contrast, HyTE~\cite{81} projects entities and predicates onto time-specific hyperplanes (inspired by TransH~\cite{72}) for distinguishing event representations across different timestemps. TERO~\cite{259} further refines this idea by combining HyTE and RotatE~\cite{117}, which replaces HyTE's hyperplane projection with RotatE's rotation operation to more effectively align entities and predicates at specific timestemps.

TDistMult~\cite{260} and TComplEx~\cite{261} extend DistMult~\cite{121} and ComplEx~\cite{122} models via integrating a time scale component to compute event quadruple scores through four-way tensor decompositions. However, these models typically assume that an entity's embedding remains constant over time, an assumption that may not hold in practice. DE-SimplE~\cite{262} addresses this limitation by learning distinct entity embeddings for different timestamps using temporal embeddings derived from word embeddings. ATiSE~\cite{263} represents entities and predicates as time-related Gaussian distributions, thereby modeling uncertainty over time. TARGAT~\cite{264} leverages a temporal Transformer to encode event quadruples, learning time-aware neighborhood representations via graph attention mechanisms. Most of the time embedding-based models excel at modeling event timeliness, they often struggle to capture the temporal order between events. LCGE~\cite{265} integrates time-aware entity and predicate embeddings with temporal rule learning, meanwhile, leverages temporal rules to regularize predicate embeddings for modeling both timeliness and temporal order properties of events.

\subsubsection{Evolution Learning-based Model}

An alternative approach for modeling temporal information in TKGs involves constructing a sequence of subgraphs, each representing a snapshot of the KG at a particular timestemp~\cite{55}. These subgraphs are encoded using graph neural networks (GNNs), while recurrent neural networks (RNNs) or analogous temporal models capture the sequential dependencies among events. This type of framework facilitates the learning of implicit temporal sequences and could predict future events through evolution learning-based techniques.

Know-Evolve~\cite{266} is the first model to introduce the deep evolutionary knowledge network, which captures causal dependencies among historical events to predict future occurrences. However, Know-Evolve does not effectively model concurrent events at the same time. To overcome this limitation, RE-NET~\cite{267} aggregates historical events related to a target entity at each time step into subgraphs. Each subgraph is encoded separately via a GCN-based aggregator, and the sequence of subgraph representations is processed by a gated recurrent unit (GRU) to capture temporal dependencies, ultimately allowing for event prediction at the current time step. Despite these advances, a key challenge remains: events occurring at adjacent timestemps may differ significantly, making it difficult to construct coherent subgraph sequences. To address this, EvolveRGCN~\cite{evolvegcn} introduces time-dependent GCN parameters, in which the GCN weights at each time step are dynamically updated based on the previous time step, effectively replacing traditional subgraph sequences when sequential dependencies are weak.

Besides, to specifically model recurring events over time, CyGNet~\cite{268} draws inspiration from replication strategies in natural language generation tasks by introducing a time-aware replication mechanism. This mechanism identifies repetitive events and improves future event prediction by referencing historical occurrences. However, not all historical information is pertinent for TKGR. Many evolution-based reasoning methods indiscriminately incorporate historical data, thereby introducing noise. To mitigate this issue, CluSTeR~\cite{269} proposes a two-stage TKGR framework that combines evidence search with TKGR. The evidence selection process is modeled as a Markov decision process, enabling the model to identify and prioritize single-step or two-step paths that are most beneficial for reasoning at the current timestamp.

Overall, evolution-based TKGR methods primarily leverage causal associations between events but tend to lack explicit time-specific representations. Consequently, when reasoning relies on a limited set of past events, these approaches may struggle to effectively exploit temporal dependencies, leading to sub-optimal reasoning performance.

\subsubsection{Temporal Rule Learning}

StreamLearner~\cite{270} is an early algorithm that extends the static rule learning algorithm RLvLR~\cite{211} to TKGs by appending temporal attributes for the mined static rules. TLogic~\cite{271} employs temporal random walks to mine rules with more diverse temporal characteristics. TILP~\cite{272} further refines this approach by designing a Markov-constrained random walk strategy to search for candidate temporal rules that incorporate time intervals, and by applying non-Markovian constraints for rule filtering to yield more reliable rules. Based on TILP, TEILP~\cite{273} introduces a differentiable random walk method for efficient rule learning and path construction within TKGs. TEILP associates each rule with a conditional probability density function that represents the probability distribution of an event’s occurrence given certain conditions, enabling more precise modeling and prediction of complex temporal distributions. However, these approaches are primarily designed for entity reasoning tasks and are limited in their ability to predict the precise time of an event. To address this limitation, NeuSTIP~\cite{274} is developed within a neural-symbolic framework, utilizing an intuitive rule-based language to enhance temporal consistency between adjacent predicates in the rule body, and introducing a confidence evaluation mechanism for temporal rules by integrating symbolic rules with KG embeddings. By effectively combining symbolic reasoning with KGE, NeuSTIP improves both the accuracy and explainability of TKGR.

\subsubsection{Multi-Step TKGR Model}

Multi-step TKGR is a comprehensive task that integrates both temporal KGR and multi-step KGR. Han et al.~\cite{79} introduce the first multi-step TKGR model xERTE. Given a TKGR query in the format of $(subject, predicate, ?, time)$, xERTE employs a message-passing strategy inspired by graph learning. Starting from the query subject, it iteratively samples relevant edges within the current entity's subgraph and propagates attention along these edges. To ensure that subgraph expansion aligns with the query’s intent, xERTE incorporates a novel temporal-aware relational graph attention mechanism that enforces temporal constraints on message propagation, thereby maintaining causal dependencies among events. After several rounds of expansion and pruning, the missing object in the query is inferred via interpretable subgraph paths.

Based on reinforcement learning-based approaches from static multi-step KGR, TPath~\cite{275} formulates path reasoning in TKGs within an RL framework. Unlike static KGR, TPath integrates temporal information into the design of states, action spaces, policy networks, and reward functions, enabling effective path search and multi-step KGR in TKGs. CluSTeR~\cite{269} applies reinforcement learning to identify all paths related to a query subject across the entire TKG. Each event quadruple encountered is converted into a factual representation at a specific timestamp. The model encodes these time-stamped facts using an R-GCN and employs a GRU to capture temporal dependencies across timestamps. By modeling all facts within a given time window and their temporal associations, CluSTeR leverages an MLP to compute a probability distribution over candidate entities, thereby enhancing reasoning accuracy. The T-GAP model~\cite{276} further improves KGR by encoding contextual knowledge related to the query event using a GNN. During decoding, attention is propagated from entities to their chronologically reachable neighbors, enabling a selective, step-by-step exploration of the KG and ensuring interpretable temporal multi-step KGR.

Most existing multi-step TKGR models primarily focus on event ordering while neglecting the modeling of specific time intervals between events. To address this gap, RTTI~\cite{277} introduces a novel time interval representation that leverages the median of two timestamps and variations in timestamp embeddings. By incorporating this representation into a reinforcement learning framework, RTTI facilitates more effective multi-step reasoning in TKGs. TITer~\cite{278} further addresses the challenge of multi-step reasoning for unseen timestamps by incorporating relative time encoding into an RL-based path search framework. This enhancement improves the model’s capacity to capture unknown temporal information during action selection. Besides, TITer integrates a temporal reward function based on the Dirichlet distribution, effectively embedding temporal information into the reward mechanism, which in turn guides the model toward more efficient training and enhances its overall reasoning ability over TKGs.

\begin{figure*}
    \centering
    \includegraphics[width=1\linewidth]{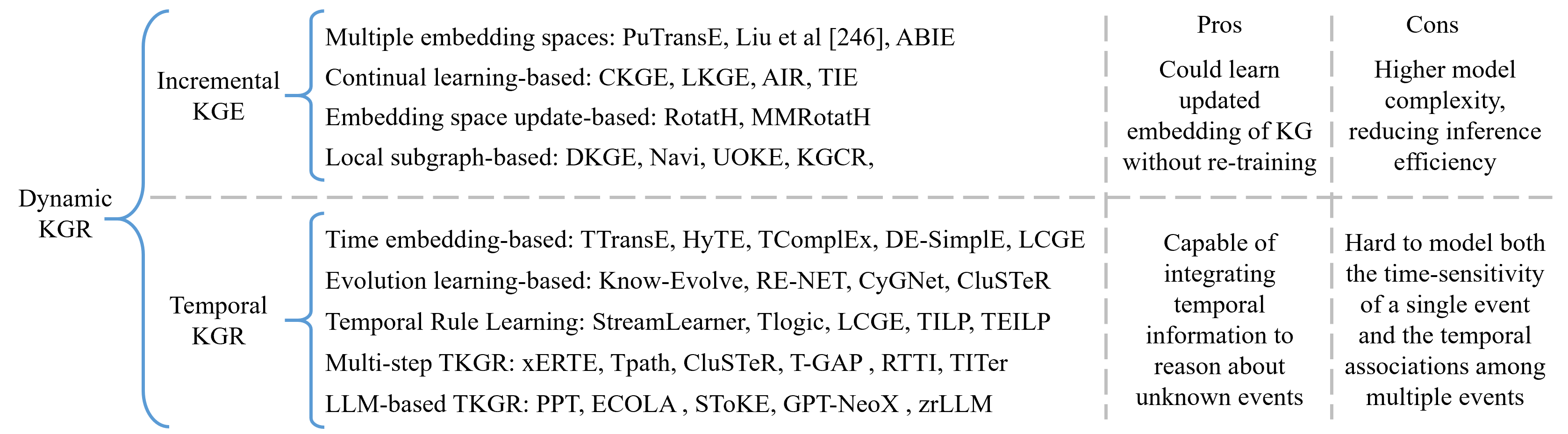}
    \caption{Taxonomy of dynamic KGR approaches with the comparison of their advantages and disadvantages.}
    \label{fig:tax_dynamic}
\end{figure*}

\subsubsection{LLM-based TKGR Model}

With rapid advancements in large language model (LLM) technology, a variety of LLM-based approaches for TKGR have recently developed. These models typically encode event quadruples and their associated contextual information as inputs to LLMs, thereby enabling event reasoning. Broadly, LLM-based TKGR models can be classified into two categories: (a) the models relying solely on LLMs, and (b) the models combining LLMs with KGE technique. The former category reformulates KGR as a mask prediction or in-context learning task, directly deriving reasoning results from LLM. In contrast, the latter category uses the output from LLM as input to a KGE model, which then produces the final reasoning results.

(a) \textbf{Models relying solely on LLMs}: PPT~\cite{279} redefines TKGR as a masked word prediction task tailored for language models. It converts a series of event quadruples associated with the query subject into a textual input format for a LLM. To encode temporal information, PPT introduces interval-based cues, designating the earliest timestamp as ``start'' with subsequent time intervals described using phrases such as ``the next day'', ``three weeks later'' or ``one month later''. LLM subsequently predicts the masked entities, thereby enhancing its ability to comprehend and infer temporal knowledge. Similarly, ECOLA~\cite{280} employs a mask prediction paradigm, but it integrates both event quadruples and their descriptive textual information into the input sequence. By employing a multi-level masking strategy such as masking words, entities, and predicates, ECOLA is able to effectively encode event semantics and improve reasoning performance.

SToKE~\cite{281} can be regarded as an extension of K-BERT~\cite{282} specific to static KGs, jointly encoding graph topology and temporal context. A key innovation of SToKE is the construction of an Event Evolution Tree (EET) for each query, which organizes structured event sequences according to their temporal characteristics. The model embeds both topological and temporal context information at each timestamp and fine-tunes a LLM via a masked prediction task, thereby enhancing reasoning accuracy. GPT-NeoX~\cite{283} fine-tunes a LLM through in-context learning to endow it with TKGR capabilities. For a given query, the model retrieves relevant recent events from the TKG as context and reformulates them into lexical cues for event prediction. Then, the LLM generates probability distributions over candidate entities, which serve as the basis for reasoning. Notably, experimental results indicate that in-context learning substantially improves the temporal reasoning ability of LLMs without necessitating additional training. Moreover, the model’s performance remains robust even when entity and relation names are replaced with random tokens, suggesting that it does not depend solely on prior semantic knowledge. While models such as SToKE retain topological information in a tree structure, they may not fully preserve the intrinsic graph structure of the KG. To address this limitation, CSProm-KG~\cite{284} uses the entity and relation embeddings to generate conditional soft prompts which are then fed into a frozen pre-trained language model to fuse the textual and structural knowledge together. The fused conditional soft prompts are then input into a KGE model (e.g., ConvE~\cite{126}), which computes similarity scores between candidate objects and the conditional soft prompts, thereby effectively integrating both graph-structural information and textual knowledge for improving reasoning performance.

(b) \textbf{Models combining LLMs with KGE}: Another hybrid strategy leverages LLMs for data augmentation. zrLLM~\cite{285} utilizes GPT-3.5 to generate enriched relational textual descriptions, which are subsequently encoded using a T5-based model to produce relational textual embeddings. A fully connected network then aligns these embeddings with the KGE model’s embedding space. Finally, the relational textual embeddings and event quadruples are jointly processed by a temporal KGE model, which assigns scores to event quadruples. This approach facilitates low-resource TKGR by leveraging knowledge generated by LLM to improve the performance of KGE models.

Fig.~\ref{fig:tax_dynamic} provides a comprehensive overview of representative models across various approaches for dynamic KGR tasks, which categorizes the models into two primary sub-tasks namely incremental KGE and temporal KGR, and briefly summarizes the respective advantages and disadvantages of these models. On one hand, incremental KGE methods excel at continuously updating KG embeddings as new data arrives, but they may struggle with scalability and the gradual accumulation of noise over time. On the other hand, temporal KGR approaches effectively capture the evolving temporal dynamics of events, yet often encounter challenges in accurately modeling fine-grained time intervals and causal dependencies. Particularly, Fig.~\ref{fig:tax_dynamic} could serve as a valuable roadmap for identifying current research gaps and guiding future innovations in dynamic KGR.

\section{Multi-Modal KGR}
\label{sec:multi-modal}

Static KGR tasks traditionally consider only the factual triples contained in the KG, thereby ignoring a wealth of auxiliary information associated with entities. To further enhance reasoning performance, recent approaches have incorporated multi-modal data such as textual descriptions and images of entities and relations into the reasoning process, a paradigm referred to as multi-modal KGR (MMKGR). Current MMKGR models can be broadly categorized into two groups according to the manners exploiting multi-modal information. The first category, termed multi-modal embedding–based models, designs KGE models that integrate multi-modal information directly into the entity embeddings. The second category consists of fine-tuning–based methods that leverage pre-trained language models (PLMs), adapting them to MMKGs for KGR.

\subsection{Multi-Modal Embedding-based Model}

In terms of the scheme of learning and applying multi-modal embeddings, there are four streams for MMKGR tasks: (1) text-based MMKGR model, (2) multi-modal fusion-based model, (3) multi-modal integration-based model, and (4) multi-modal-enhanced negative sampling.

\subsubsection{Text-based MMKGR Model}

Early MMKGR methods focused exclusively on textual descriptions associated with entities and relations. For instance, Wang et al.~\cite{286} are among the first to incorporate textual description information into KGE. Their key idea is to align entities with their corresponding textual descriptions, using entity names or anchor text from sources such as Wikipedia to jointly learn embeddings in a unified space, thereby enhancing entity representation capabilities. Furthermore, DKRL~\cite{287} employs a bag-of-words approach combined with a convolutional neural network (CNN) to encode textual description of each entity. It reconstructs a triple by combining the encoded descriptions of the head and tail entities with the relation representation, and applies a mechanism akin to TransE~\cite{69} to model interactions between entities and relations, achieving improved reasoning accuracy over the traditional model TransE. TEKE~\cite{288} further refines this approach by employing an entity linking tool to establish more accurate connections between textual descriptions and entities within a corpus, and extracts contextual word sets to further enhance entity embeddings.

In contrast, KG-BERT~\cite{101} leverages a Transformer-based BERT model to jointly encode entities, relations, and their associated textual descriptions. The model then considers the KGR task as a triple classification task. Similarly, SimKGC~\cite{138} combines a PLM with a contrastive learning strategy to encode the textual information of entities, thereby achieving superior KGR performance. StAR~\cite{139} also employs a PLM to encode textual descriptions, but it additionally preserves the structural information of the KG to maintain graph topology during reasoning.

\subsubsection{Multi-Modal Fusion-based Model}

To further enhance MMKGR performance, a class of embedding-based methods has been proposed to enhance entity embeddings by fusing multi-modal features including both text and images. For instance, IKRL~\cite{289} encodes images corresponding to entities into the entity embedding space, thereby modeling the association among an entity pair and the linked relation in each triple to learn entity and relation embeddings in a manner similar to TransE. Besides, both TransAE~\cite{290} and RSME~\cite{291} models enhance entity embeddings by integrating visual information with structural features into a unified embedding space, thereby improving the accuracy of KGR. To mitigate the influence of irrelevant or noisy images among the multiple images associated with an entity, RSME incorporates a specialized gate mechanism to effectively select the most valuable image information.

Moreover, Cao et al.~\cite{292} develop an optimal transport approach to fuse multi-modal data by minimizing the Wasserstein distance between multi-modal distributions, effectively modeling the fusion process as a transportation task where different modality embeddings are aligned into a unified space. HRGAT~\cite{293} further advances multi-modal fusion by modeling both intra- and inter-modal information through low-rank fusion. Specifically, HRGAT transforms the original KG into a hyper-graph and learns structural features using a graph attention network that incorporates relation-specific attention and entity–relation fusion operations. In addition, MKBE~\cite{294} treats multi-modal information as an additional triad, designing combinatorial coding components to jointly learn entities and their multi-modal embeddings.

Considering the challenges posed by unbalanced multi-modal information distribution, NativE~\cite{87} introduces a relation-guided bi-adaptive fusion module that adaptively fuses any modality, and employs a collaborative modal adversarial training framework to augment under-represented modal information. Besides, MMKGR~\cite{295} utilizes a gated attention network to generate multi-modal complementary features, facilitating extensive multi-modal interaction and noise reduction. These features are subsequently fed into a reinforcement learning framework for multi-step KGR to address the sparse reward issue inherent in such tasks. In general, while models that fuse multi-modal features are typically computationally efficient and scalable to large datasets, the presence of noisy or poorly encoded information in one modality might significantly impact the entire performance, limiting the model's robustness.

\begin{figure*}
    \centering
    \includegraphics[width=1\linewidth]{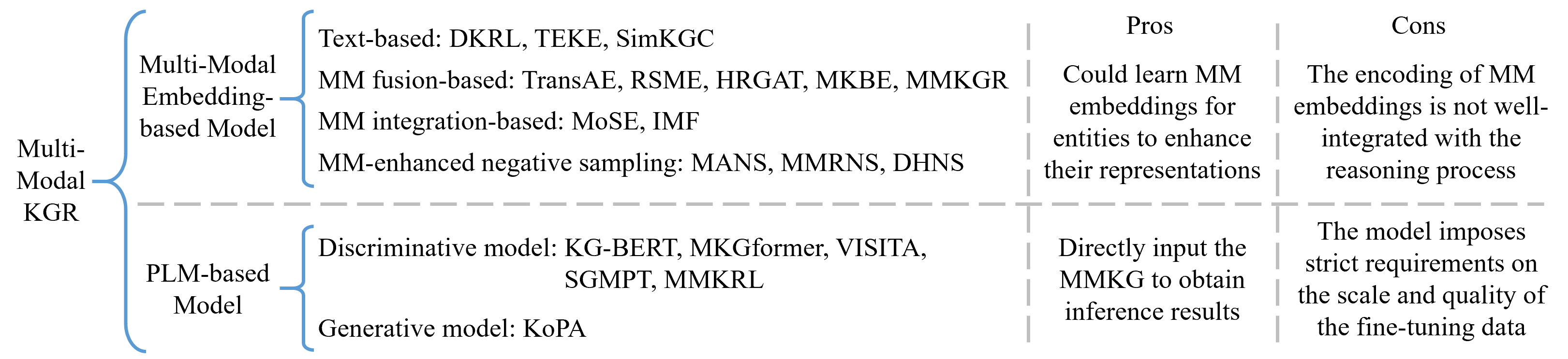}
    \caption{Taxonomy of multi-modal KGR approaches with the comparison of their advantages and disadvantages.}
    \label{fig:tax_mmkgr}
\end{figure*}

\subsubsection{Multi-Modal Integration-based Model}

Another class of embedding-based methods employs ensemble learning, in which distinct models are trained on different modalities and their outputs are subsequently fused to form the final multi-modal entity embeddings for KGR. For instance, MoSE~\cite{296} decouples tightly coupled multi-modal embeddings during training by leveraging factual triples, texts, and images to independently train three distinct KGR models. During reasoning, the model adaptively adjusts the weights of the different modalities through an ensemble learning scheme and performs joint reasoning. Similarly, IMF~\cite{297} uses modality-specific encoders to independently extract topological, visual, and textual features. In the integration stage, the scores obtained from these modalities are combined to yield the final reasoning result. This integration approach fully exploits the complementary information inherent in different modalities, thereby enhancing overall model robustness. In cases where one modality underperforms, the contributions from the others can compensate, thus improving the robustness. However, these approaches require the training of multiple models, which increases the overall computational complexity.

\subsubsection{Multi-Modal-Enhanced Negative Sampling}

Multi-modal data can not only directly enhance entity embeddings but also improve the KGE training procedure via generating higher-quality negative triples. By leveraging context-assisted information, it could generate negative embeddings that do not correspond to the current entity from the perspective of visual feature, thereby enforcing modality alignment between KG embeddings and visual embeddings and further boosting performance in KGE and KGR. To enhance the mining of challenging negative triples, MMRNS~\cite{85} introduces a knowledge-guided cross-modal attention mechanism coupled with a contrastive loss. This mechanism constructs a contrast semantic sampler that learns the multi-modal semantic dissimilarity between positive and negative triples, enabling a more precise estimation of sampling distributions and yielding higher-quality negatives. DHNS~\cite{dhns} is the first approach to leverage a diffusion model for capturing the diverse semantics across different modalities to generate hierarchical, high-quality negative triples, while directly controlling hardness levels through diffusion time steps.

\subsection{PLM-based Model}

Approaches based on fine-tuning pre-trained language models (PLMs) enhance MMKGR performance by transforming triples from a MMKG into token sequences, feeding these sequences into PLMs. Then, they could fine-tune the models on the tasks that PLMs specialize in, such as classification or generation. Particularly, this process capitalizes on the rich multi-modal comprehension capabilities of these models. For instance, single-stream PLMs like VLBERT~\cite{299}, VisualBERT~\cite{300}, Unicoder-VL~\cite{301}, and UNITER~\cite{302} encode image and text embeddings in a unified framework to learn deep contextual embeddings. In contrast, dual-stream models such as LXMERT~\cite{303} and ViLBERT~\cite{304} process visual and textual information separately before fusing them via cross-channel or joint attention mechanisms. Specifically, PLM-based approaches for KGR can be further categorized into discriminative model-based and generative model-based methods.

\subsubsection{Discriminative Model}

Discriminative models formulate the KGR as a classification task. For instance, MKGformer~\cite{84} and VISITA~\cite{305} both utilize a Transformer-based framework employing Vision Transformer (ViT) for image encoding and BERT for text encoding to construct a multi-level fused encoder that integrates image and text features for entity representation. They typically cast the KGR task as a mask prediction problem, where masked entities are predicted by combining entity descriptions, relations, and images. SGMPT~\cite{306} extends the capabilities of MKGformer by a dual-policy fusion module that incorporates KG embeddings obtained by the KGE model HAKE and multi-modal embeddings via ViT and BERT. MMKRL~\cite{307} further augments SGMPT by integrating the original multi-modal data with structural knowledge using a TransE model and a pre-training phase to reconstruct embeddings in a unified space, accompanied by an alignment module that minimizes reconstruction error.

\subsubsection{Generative Model}

Generative model-based approaches recast the MMKGR task as a sequence-to-sequence problem, whereby a language model is fine-tuned for both encoding and decoding. For example, KoPA~\cite{308} integrates pre-trained KG embeddings with a LLM through the use of a knowledge prefix adapter. This adapter maps KG embeddings into the textual space to generate virtual token sequences that serve as prefixes for the input, thereby injecting cross-modal structural information into the LLM. Although this generative strategy can better preserve the contextual relationships inherent in the sequences, its output tends to be less stable and it demands significantly more computational resources compared to discriminative approaches. It is also important to note that, while the incorporation of multi-modal data can substantially enrich the semantic information of KGs and enhance KGE, not all KGs contain effective multi-modal data. Consequently, the practical use of MMKGR models remains limited.

Fig.~\ref{fig:tax_mmkgr} provides an overview of representative models for each category in the MMKGR task. It categorizes the models into those based on multi-modal embeddings and those leveraging fine-tuned PLMs, summarizing their respective advantages and disadvantages.

\section{Few-Shot KGR}
\label{sec:few-shot}

The previously reviewed KGR tasks assume that all entities and relations are associated with a sufficient number of triples. This assumption facilitates the learning of semantic associations during training, thereby enabling effective reasoning. However, real-world KGs often exhibit a long-tail distribution, and dynamic KGs may introduce new entities or relations~\cite{fewshotgraph}. In both cases, certain entities or relations may have only a few or even without any associated triples. Such reasoning tasks are classified as few-shot KGR. Notably, existing researches in few-shot KGR primarily address scenarios where relations are associated with limited triples, while reasoning over unseen entities is discussed in the next section on inductive KGR.

\begin{figure}
    \centering
    \includegraphics[width=1\linewidth]{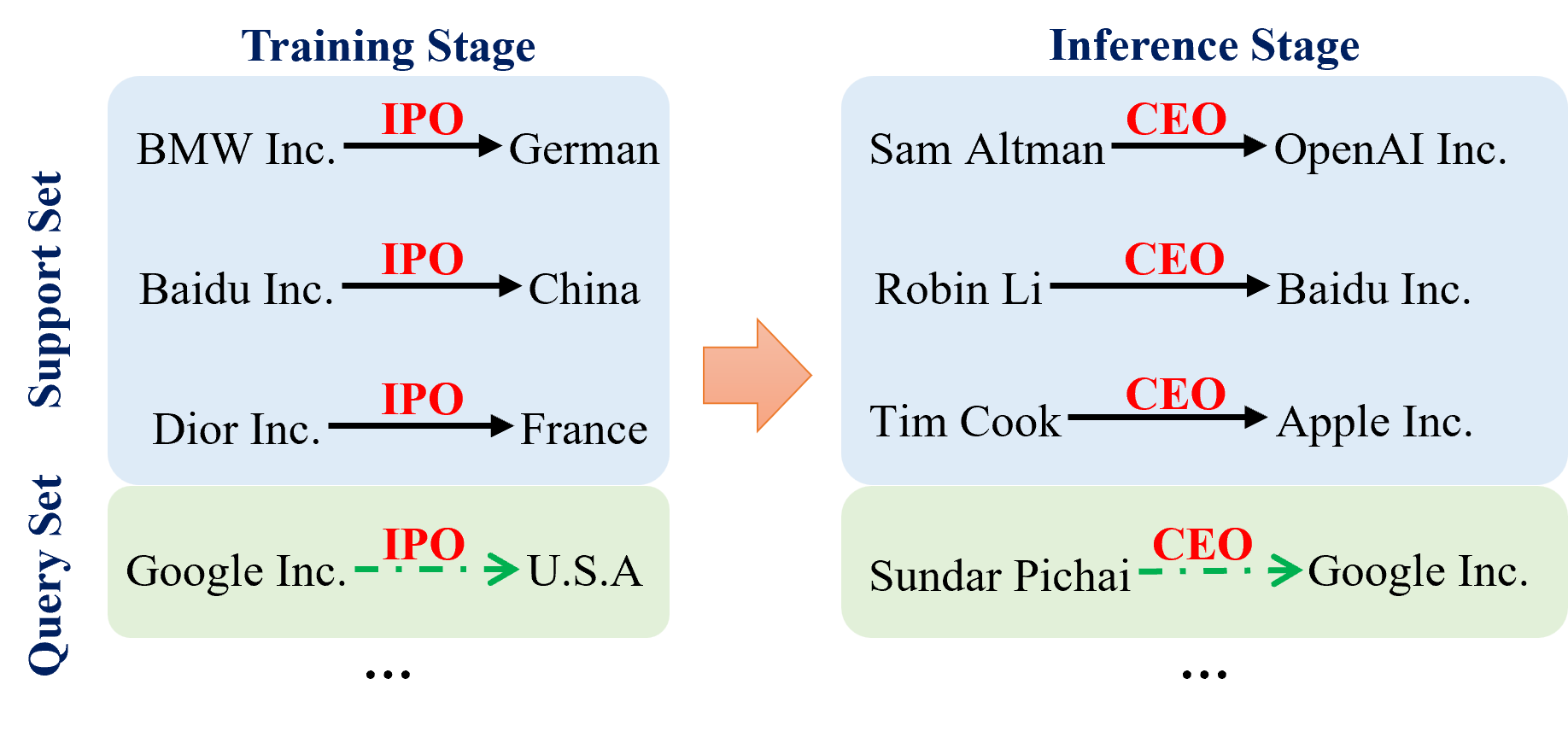}
    \caption{The illustration of few-shot KGR in the 3-shot setting.}
    \label{fig:fskgr}
\end{figure}

The few-shot KGR (FSKGR) task follows the conventional N-way K-shot paradigm commonly used in few-shot learning research in fields such as image and text classification~\cite{309}. Here, ``N-way'' denotes the number of relations involved in the reasoning process, and ``K-shot'' indicates that the support set for each relation contains K triples. Based on this setup, the objective is to infer the missing triples in the query set typically defined according to the same K-shot configuration. The training procedure mirrors the inference setup, with mutually exclusive training, validation, and test sets. Specifically, for each long-tailed relation, given a support set containing only a small number of triples $S_r=\{(h_i,r,t_i\ )|(h_i,r,t_i\ )\in T_r\}_(i=1)^K$, inference is performed on the remaining query set $Q_r={{\left(h_i,r,t_i\right)|\left(h_i,r,t_i\right)\in T_r}}_{i=1}^{\left|T_r\right|-K}$ of the relation $r$.  Fig.~\ref{fig:fskgr} illustrates an example of few-shot KGR under the 3-shot setting.

Currently, researches on few-shot KGR primarily focus on three sub-tasks: (1) few-shot single-step reasoning, (2) few-shot multi-step reasoning, and (3) few-shot temporal reasoning. Among these, approaches for few-shot single-step reasoning can be broadly categorized into metric learning-based models, meta-learning-based models, and auxiliary information-based models. The following sections provide a detailed overview of these sub-tasks and the corresponding methodologies.

\subsection{Metric Learning-based Model}

Xiong et al.~\cite{89} firstly define the few-shot KGR task and propose the metric learning-based model GMatching, primarily addressing the extreme 1-shot setting. Recognizing that a single triple in the support set is insufficient to effectively represent a long-tailed relation, GMatching~\cite{89} leverages the neighborhood subgraph of an entity to encode each candidate head-tail entity pair. A metric learning model is then trained to match the encoded representations of entity pairs in the support set with those in the query set, thereby predicting whether the reference and the query entity pairs are connected by the same relation. On the basis on GMatching, FSRL~\cite{310} employs a relation-aware neighbor encoder to capture the varying influences of neighboring entities, and utilizes a recurrent auto-encoder to aggregate multiple reference entity pairs from the support set. Further extending this work, FAAN~\cite{311} uses a Transformer to encode entity pairs and introduces an adaptive matcher that selectively aligns a query entity pair with multiple reference entity pairs, acknowledging that different triples may contribute unevenly to representing the same relation.

TransAM~\cite{312} serializes entity pairs from both support and query sets and employs a Transformer with local-global attention to capture both intra-triple and inter-triple interactions, thereby facilitating the matching process. Similarly, FRL-KGC~\cite{313} incorporates a gating mechanism during the encoding of an entity's neighborhood subgraph to mitigate the impact of noisy neighbors. By integrating LSTM and Transformer architectures for separate encoding of reference and query triples, FRL-KGC enhances the capture of contextual semantics in entities and relations, with final inference achieved through a matching process. While most metric learning-based approaches focus exclusively on matching entity pairs, they often neglect the direct representation specific to relations. To address this issue, HMNet~\cite{314} introduces a hybrid matching network that jointly computes matching scores at both the entity and relation levels via entity-aware and relation-aware matching networks. Furthermore, MetaP~\cite{315} employs a CNN-based pattern learner to extract relation-specific patterns from a limited number of reference triples, and a balanced pattern matcher computes the similarity between the pattern representations of both positive and negative reference triples together with that of the query triple, thus predicting the query triple's plausibility.

In summary, metric learning-based few-shot KGR models primarily encode the entity pairs associated with long-tail relations. They focus on representing the neighborhood subgraphs of these pairs and then matching the support and query set representations to determine whether they share the same relation. However, by relying solely on entity pair matching, these methods do not directly learn relation representations from the available triples, which limits their performance in few-shot reasoning scenarios for long-tail relations.

\subsection{Meta-Learning-based Model}

To effectively learn embeddings for long-tailed relations from a limited number of triples, some studies have leveraged the Model-Agnostic Meta-Learning (MAML) mechanism~\cite{316} to transfer existing knowledge into relation meta-representations. MetaR~\cite{317} encodes entity pairs from the support set to obtain relation meta representations, which are then rapidly optimized using standard KGE models (e.g., TransE) and gradient descent strategies. The optimized relation meta-representations are subsequently used to refine the query set, with candidate triples scored based on their likelihood of being valid. Although MetaR demonstrates promising performance upon its introduction, challenges remain in effectively encoding entity pairs to derive relation meta-representations and in accounting for the varying importance of different support set triples during optimization. To better exploit neighboring information while suppressing the influence of noisy neighbors, GANA~\cite{318} employs a gated graph attention neighbor aggregator to encode entities and obtain relation meta-representations. Besides, an attention-based bidirectional LSTM (BiLSTM) fuses various support set triples into an integrated relation meta-representation. Notably, this approach develops a MTransH module by integrating the MAML mechanism into the KGE model TransH~\cite{72}, which computes triple scores and enhances few-shot KGR performance in scenarios with complex relations. In pursuit of improved inference efficiency, SMetaR~\cite{320} streamlines the model GANA by applying a linear mapping to head-tail entity pairs in the support set to learn relation meta-representations. Meta-iKG~\cite{319} reformulates the few-shot KGR task as a subgraph modeling task. The parameters of GNN encoding the long-tailed relation subgraphs are initially learned via MAML from triples of different long-tail relations, yielding an optimal parameter initialization that is subsequently fine-tuned with the support set, thereby improving representations of long-tailed relations. 

HiRe~\cite{321} employs a three-level framework to learn relation meta representations: first, a neighbor aggregator captures entity-level representations. Second, a contrastive learning strategy derives relation-context-level representations. Finally, a Transformer encodes the support set at the triple level, offering a comprehensive representation of long-tail relations. MTRN~\cite{322} introduces a self-attention-based entity pair encoder to capture interactions between head and tail entities associated with long-tailed relations. These refined embeddings are then aggregated via convolution operations to form a relation meta-representation, with a mechanism to weight the contributions of various entity pair embeddings differently. Thus, meta learning-based few-shot KGR approaches primarily leverage the MAML mechanism to learn relation representations from a limited number of triples. However, these models involve a parameter update process that can negatively impact inference efficiency and are inherently constrained by their reliance on only a few triple facts, which limits their overall performance in representing and reasoning about long-tailed relations.

\subsection{Auxiliary Information-Enhanced Model}

To address the limitations imposed by the scarcity of triples for long-tailed relations in few-shot scenarios, several approaches incorporate auxiliary information to enrich the representation of these relations. The primary auxiliary sources include textual descriptions, ontological information, and path information.

\subsubsection{Textual Description-Enhanced Model}

TCVAE~\cite{323} enriches the KG by leveraging textual descriptions of entities and relations. It employs a text encoder to extract salient features from the textual descriptions and utilizes a generator to produce additional triples, thereby alleviating data sparsity in few-shot settings. For the more challenging zero-shot scenario, ZSGAN~\cite{324} enables effective generation of relation embeddings solely from textual descriptions via a Generative Adversarial Network (GAN), even in the absence of direct training data for certain relations. The generated relation embeddings are then compared with entity pair embeddings using cosine similarity to evaluate triple plausibility. In addition, HAPZSL~\cite{325} introduces a hybrid attention mechanism that incorporates both relation and entity attention during the encoding of textual descriptions. This mechanism allows the model to better capture the semantic dependencies between entities and relations, thereby enhancing its generalization capability. Meanwhile, a prototypical network is employed to learn a relation prototype that represents the typical features of each relation, so that entity pair embeddings can be aligned with their corresponding prototypes.

\subsubsection{Ontology-Enhanced Model}

Given that KGs typically include an ontological layer comprising concepts and their relations, OntoZSL~\cite{326} leverages the inherent concept hierarchy to construct a prior concept graph. Then, a GNN is used to encode this graph, enhancing the embeddings of long-tail relations. Similarly, DOZSL~\cite{327} proposes an attribute-guided decoupled ontology embedding method to extract fine-grained inter-class associations between seen and unseen relations. This decoupled embedding is subsequently fused using a GAN-based generative model and a GNN-based propagation model to generate embeddings for unseen relations. Furthermore, DMoG~\cite{328} employs Graph Convolutional Networks (GCNs) to jointly learn textual and ontological features from a corresponding ontology and text graph, thereby improving the embeddings of these unseen relations.

\subsubsection{Path Information-Enhanced Model}

The aforementioned few-shot single-step KGR models primarily rely on direct associations between entities for matching and encoding, often neglecting the indirect semantic connections embedded in multi-hop paths. P-INT~\cite{329} exploits a metric learning approach that, unlike earlier methods, represents entity pairs by leveraging the multi-hop paths connecting the head and tail entities. It learns embeddings for all paths between each entity pair and introduces a path-interaction mechanism with attention over the paths to compute the similarity between entity pairs in the support set and those in the query set, thereby facilitating the matching process and yielding improved few-shot KGR results. Along the same lines, EPIRL~\cite{330} utilizes reinforcement learning to construct reasoning subgraphs from the KG, forming inference path rules between reference triples and query triples. It further designs a path-based matching mechanism to capture intrinsic associations among these reasoning paths and employs a relation attention mechanism to highlight the most influential paths.

\subsection{Multi-Step Few-Shot KGR Model}

\begin{figure*}
    \centering
    \includegraphics[width=1\linewidth]{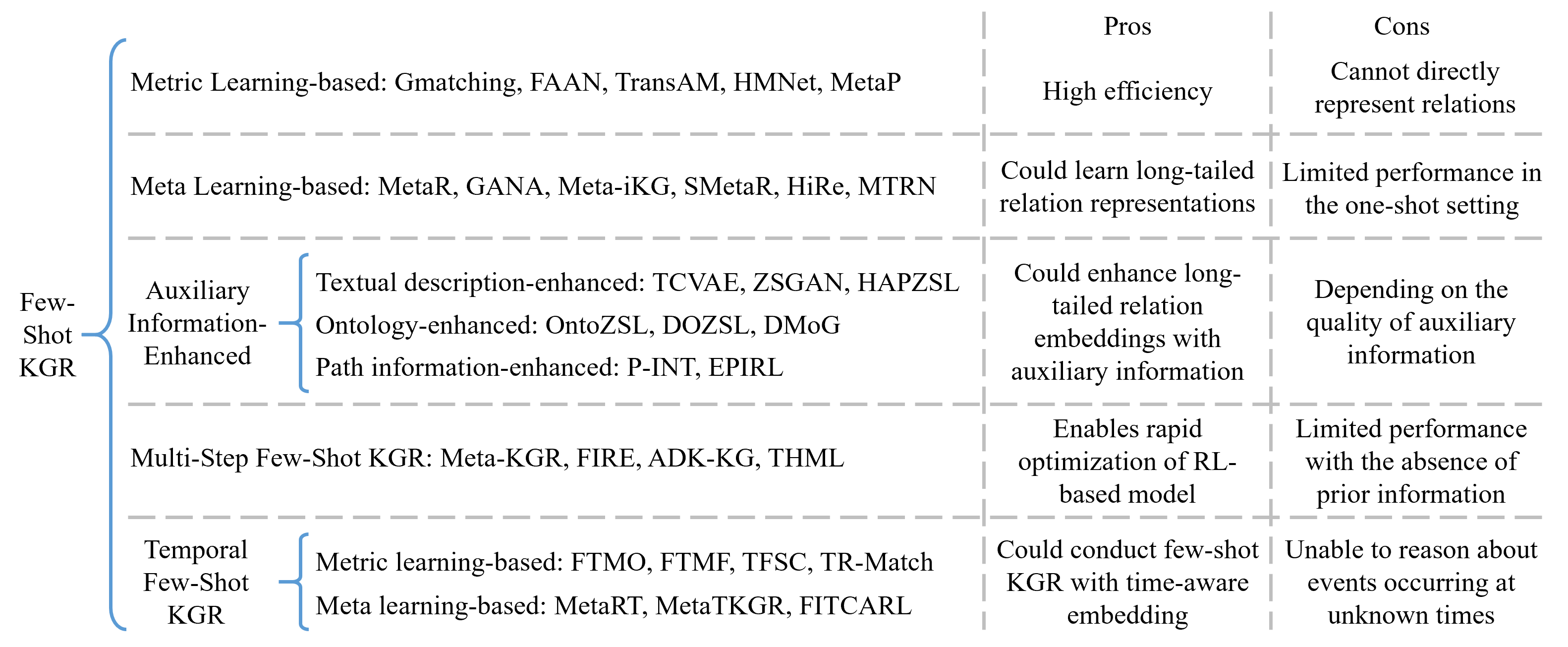}
    \caption{Taxonomy of few-shot KGR approaches with the comparison of their advantages and disadvantages.}
    \label{fig:tax_fskgr}
\end{figure*}

To enable multi-step KGR in few-shot scenarios, Meta-KGR~\cite{331} incorporates MAML mechanism into a reinforcement learning framework for multi-step reasoning. This model trains the reinforcement learning agent on high-frequency relation samples and treats the learned parameters as meta-parameters. For long-tailed relations, these meta-parameters are rapidly adapted using the MAML mechanism, thereby tailoring the multi-step reasoning process to the new relation. Based on Meta-KGR, FIRE~\cite{332} introduces a neighbor aggregator to encode entity representations and prunes the path search space using KG embeddings, thereby enhancing path search efficiency and reasoning performance. Subsequently, ADK-KG\cite{333} refines FIRE by designing a text-enhanced GNN to improve entity encoding. In this model, textual information is encoded with a pre-trained BERT model, and a self-attention mechanism computes weights specific to different few-shot KGR tasks (namely relations), thereby reinforcing the MAML-based meta-learning process. Recognizing that reinforcement learning-based models often struggle with hard relations exhibiting more loss during training, THML~\cite{334} designs a hardness-aware meta-reinforcement learning module with the MAML mechanism to predict the missing element by training hardness-aware batches. Thus, THML could identify hard relations in each training batch and generates corresponding samples for training, resulting in performance improvements.

In summary, existing few-shot multi-step KGR models primarily focus on rapidly updating reinforcement learning model parameters via meta learning mechanisms to adapt to long-tailed relations. However, these approaches still do not incorporate certain types of prior information such as logical rules, which limits their overall performance.

\subsection{Temporal Few-Shot KGR Model}

Temporal few-shot KGR can be viewed as an intersection between few-shot KGR and temporal KGR. For example, metric learning-based models for temporal KGR not only consider matching between entity pairs in the query and support triples, but also learn time-aware representations for these pairs. Overall, current few-shot temporal KGR models are largely extensions of existing few-shot reasoning approaches and can similarly be divided into metric learning-based and meta learning-based approaches.

\subsubsection{Metric Learning-based Model}

FTMO~\cite{335} extends FSRL~\cite{310} from a temporal information modeling perspective by proposing a temporal relation-aware neighbor encoder. This encoder aggregates neighbors while incorporating temporal information, and a recurrent auto-aggregation network fuses semantic information from all reference triples in the support set. A matching mechanism then assesses the similarity between reference and query triples. TFSC~\cite{336} improves upon FAAN~\cite{311} by replacing its original approach with TTransE~\cite{258} to encode entities enriched with temporal information. After encoding the entity's neighbors with an attention mechanism, a Transformer encoder learns time-aware representations for entity pairs, enhancing their quality under temporal KGR settings and improving matching effectiveness in both temporal and few-shot scenarios. TR-Match~\cite{337} designs a multi-scale temporal relation attention encoder that adaptively captures both local and global information. By fusing temporal information with relational data, it generates entity embeddings at different scales, thus enhancing the modeling of temporal characteristics. FTMF~\cite{338} employs a self-attention mechanism to encode entities and a recurrent recursive aggregation network to combine neighboring information from the support set. A fault-tolerant mechanism is also incorporated to mitigate the impact of erroneous information, and a matching network computes similarity scores.

\subsubsection{Meta Learning-based Model}

MetaRT~\cite{339} adapts the MetaR~\cite{317} framework to temporal KGR by replacing the traditional TransE scoring function with that of TTransE~\cite{258}. This modification allows MetaRT to better handle both the few-shot reasoning and the temporal reasoning tasks. MetaTKGR~\cite{340} treats time information as a supervisory signal by dynamically sampling and aggregating neighboring information from recent events, thereby enhancing the representations and reasoning for temporal-aware entities. It also employs a nested dual-loop meta learning optimization mechanism, where an inner loop encodes recent events and an outer loop facilitates rapid knowledge transfer for adaptation to future events. FITCARL~\cite{341} introduces a reinforcement learning strategy within a meta learning framework. It proposes a time-aware Transformer that utilizes time-aware positional encoding to capture enriched entity meta-representations from few-shot data, and a confidence learner is integrated to mitigate the challenges of data scarcity. By leveraging reinforcement learning to generate inference results and incorporating concept information from the KG as prior constraints, FITCARL demonstrates notable performance improvements in few-shot settings.

Fig.~\ref{fig:tax_fskgr} summarizes representative models for each category in few-shot KGR tasks, including models for single-step FSKGR, multi-step FSKGR and temporal FSKGR tasks.

\section{Inductive KGR}
\label{sec:inductive}

\begin{figure}
    \centering
    \includegraphics[width=1\linewidth]{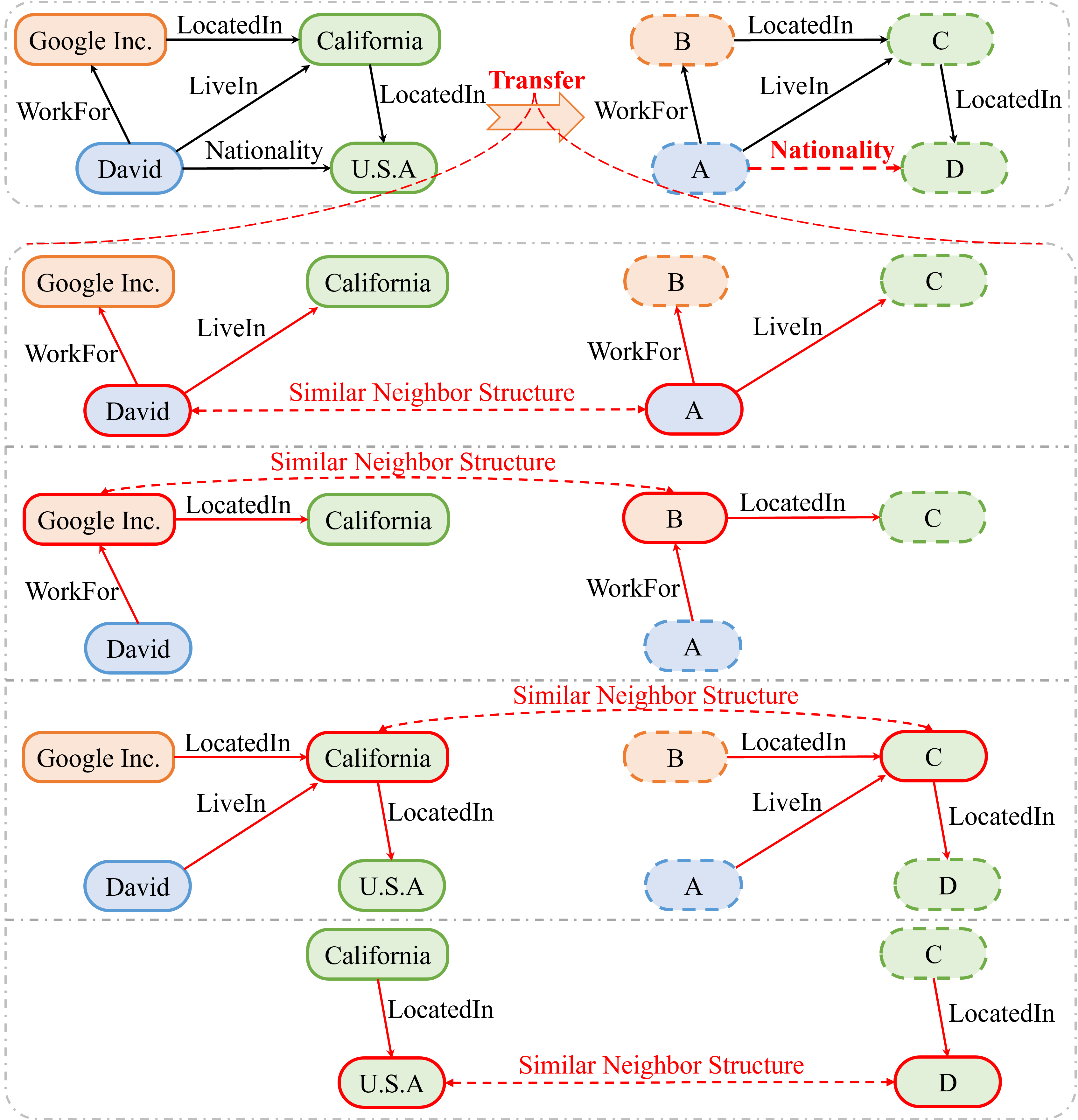}
    \caption{The illustration of inductive KGR.}
    \label{fig:ikgr}
\end{figure}

Currently, one of the most challenging tasks is inductive KGR, which primarily evaluates a model's ability to reason over unseen entities and relations~\cite{342}. Unlike few-shot KGR, which focuses on how models leverage a limited number of training samples during training, inductive KGR emphasizes the model’s generalization ability during testing. As illustrated in Fig.~\ref{fig:ikgr}, inductive KGR (IKGR) centers on generalizing transferable structural patterns from a source KG to a target KG. This process involves identifying entities with similar neighborhood structures across both KGs, thereby enabling the migration of structural patterns from the source to the target domain. By leveraging these transferred patterns, the reasoning framework could infer potential associations among previously unseen entities, effectively extending the KG's inferential capabilities beyond its original scope. In terms of strategies for endowing models with inductive capabilities, approaches can be categorized into rule-based models, GNN-based models, and multi-modal-enhanced models.

\subsection{Rule-based Model}

Since rules encapsulate high-level semantic knowledge and typically involve only variables, thus remaining independent of specific entities, they can be directly applied to reason over unseen entities. In particular, rule-based KGR models, as discussed in Section~\ref{sec:ruleKGR}, are also applicable to inductive settings. Here, we review rule learning models that have been specifically proposed for inductive KGR tasks.

\subsubsection{Graph Search-based Model}

Unlike the algorithm AMIE~\cite{202} presented in Section~\ref{sec:rule}, which computes rule confidence scores over the entire KG, RuleNet~\cite{343} proposes a dual graph traversal approach. This method leverages relation parameter information to exhaustively traverse all possible candidate rules. A key challenge in rule learning is that the candidate search space grows exponentially. To address this, CBGNN~\cite{344} exploits the linear structure of cyclic spaces to efficiently compute cycle bases, which better express the underlying rules. Within these cyclic spaces, a GNN is utilized to learn rule representations through message passing. Moreover, RED-GNN~\cite{345} constructs rule patterns by modeling relation-directed graphs composed of overlapping segments of relation paths, thereby enabling the learning of more complex logical rules.

\subsubsection{Rule and Embedding-based Model}

VN~\cite{346} introduces rule-based virtual neighbor prediction to alleviate the issue of sparse neighbor information. It identifies symmetric path rules and establishes an iterative optimization mechanism by integrating soft labels with a KGE model, thereby enriching the neighbor context for unseen entities. To further improve the generalization of rule-based reasoning approaches, ARGCN~\cite{347} and ELPE~\cite{348} both employ GCN and Graph Transformers, respectively, to encode embeddings for unseen entities. Besides, these models incorporate reinforcement learning techniques to search for reasoning paths between head entities and target tail entities as rule patterns, which enhances explainability when reasoning over unseen entities.

\subsection{GNN-based Model}

As discussed in Section~\ref{sec:KGEwtriple}, GNN-based KGE models encode an entity’s neighbor information to update its representation. This idea could be extended to encode the neighbors of unseen entities, thereby deriving their embeddings for inductive KGR. For example, MEAN~\cite{349} uses a GNN to encode an entity’s neighbors and decodes the resulting representation using the model TransE to produce inference results. However, MEAN aggregates neighbor information via a simple pooling function, which falls short in fully exploring and exploiting the available neighbor contexts. Consequently, many researches have been devoted to effectively extracting and leveraging neighbor context information for the inductive KGR task.

\subsubsection{Subgraph and Path-based Model}

NBFNet~\cite{350} is a representative approach that endows the model with inductive learning capabilities by combining path-based together with GNN-based KGE models and defining three operators to enhance the inductive capability of the GNN. Similarly, GraIL~\cite{91} predicts the relation between two entities based on the subgraph structures, and then learns relation semantics independent of specific entities. In particular, many subsequent methods extend NBFNet and GraIL. PathCon~\cite{165}, SNRI~\cite{351}, REPORT~\cite{352}, LogCo~\cite{353}, and RPC-IR~\cite{354} all improve upon NBFNet by extracting the relation context subgraphs and relation paths of unseen entity pairs, then aggregating their representations to perform relation reasoning. Each of these five models employs different strategies to learn subgraph and path embeddings. Specifically, PathCon exploits a relation-specific message passing mechanism to aggregate neighborhood information for unseen entities, while SNRI utilizes a GNN and a gated recurrent unit (GRU) to encode context subgraphs and relation paths separately. REPORT encodes both the context subgraph and relation paths between entity pairs via a Transformer. LogCo and RPC-IR address the issue of insufficient supervisory signals by constructing positive and negative relation path embeddings to derive richer self-supervised signals, and they incorporate contrastive learning mechanisms during model training.

Moreover, models such as TACT~\cite{355}, NRTG~\cite{356}, and CoMPILE~\cite{357} further refine GraIL. TACT models the semantic correlations between relations according to seven pre-defined topological patterns, namely ``head-to-tail'', ``tail-to-tail'', ``head-to-head'', ``tail-to-head'', ``parallel'', ``loop'' and ``not connected''. Then, it designs a scoring network to combine the outputs of a relational correlation module and a graph structure module to score a given triple. NRTG partitions the inter-relation structure into six topological patterns similar as TACT but without the pattern ``not connected''. Thus, TACT and NRTG could effectively utilize relation topology in an entity-independent manner for inductive reasoning. CoMPILE introduces a novel node-edge communication message passing mechanism to model directed subgraphs, which facilitates differentiating the relative importance of various relations and efficiently handles symmetric and antisymmetric relations. Given that the inductive learning ability of GNN-based models primarily originates from subgraph information, extracting the most informative subgraphs is crucial for inductive reasoning. To this end, LCILP~\cite{358} and ReCoLe~\cite{359} both employ local clustering methods to extract subgraphs that are semantically relevant to the current relation, which are then encoded via a GNN. Furthermore, DEKG-ILP~\cite{360} designs a GNN-based subgraph modeling module to leverage the global relation-based semantic features shared between source and target KGs and mine local subgraph information around each relation.

\subsubsection{Subgraph Attention-based Model}

To further enhance the utilization of subgraph information, some methods focus on assigning different importance to various pieces of information within the subgraph. For instance, CG-AGG~\cite{361} generates entity representations with multiple semantic perspectives using a global aggregator based on a hyper-graph neural network (HGNN) in conjunction with a local aggregator based on a GNN. In particular, several models consider the varying importance of different neighbors within a subgraph. FCLEntity-Att~\cite{362} combines convolutional neural networks and graph attention networks to encode the contextual representations of unseen entities and relations. SAGNN~\cite{363} assigns different weights to various neighbor nodes of an unseen entity based on topological features such as in-degree, out-degree, and co-occurrence frequency, thereby learning robust embeddings for unseen entities and relations. LAN~\cite{364} integrates logical rules and attention mechanisms to assign weights to an entity’s neighbors during subgraph encoding, while SLAN~\cite{365} designs attention weights specific to each query triple to evaluate the importance of similar entities and their neighbors within the subgraph. Additionally, ARP~\cite{366} employs an attention network to extract the most relevant subgraph and contextual features for the current query triple, and TransNS~\cite{367} selects pertinent neighbors as attributes of the entity while leveraging the semantic affinity between entities to choose related negative samples, thereby enhancing the inductive learning capability.

\subsection{Multi-Modal-Enhanced IKGR Model}

To avoid relying solely on the internal subgraph structure of a KG to obtain embeddings for unseen entities, methods that incorporate multi-modal information can fully exploit text, image, and other modalities related to entities and relations to directly learn their embeddings. The series of multi-modal KGR models introduced in Section~\ref{sec:multi-modal} are directly applicable here, so this section focuses on approaches specifically designed for inductive KGR tasks that incorporate multi-modal information.

\subsubsection{Concept-Enhanced IKGR Model}

GNN-based approaches encode neighbor information to represent unseen entities, offering strong scalability and good performance on large-scale KGs. However, these models lack the explainability and accuracy as logic rule-based models, and they merely handle the unseen entities or relations that are linked to the original KG. This makes them unsuitable for fully inductive reasoning scenarios where the target KG shares no common entities with the original KG. Besides, GNN-based methods rely on rich neighbor subgraph information, which is often sparse for unseen entities, thereby limiting their performance. To address these challenges, CatE~\cite{368} employs a Transformer to encode the contextual information of concepts from the ontological graph. CatE enhances the embeddings of unseen entities by integrating their concepts into the neighbor information. However, compared to the knowledge available at the entity level, the concept information within the ontology is relatively limited, so relying solely on structured internal knowledge may not sufficiently represent entities and relations for the IKGR task.

\begin{figure*}
    \centering
    \includegraphics[width=1\linewidth]{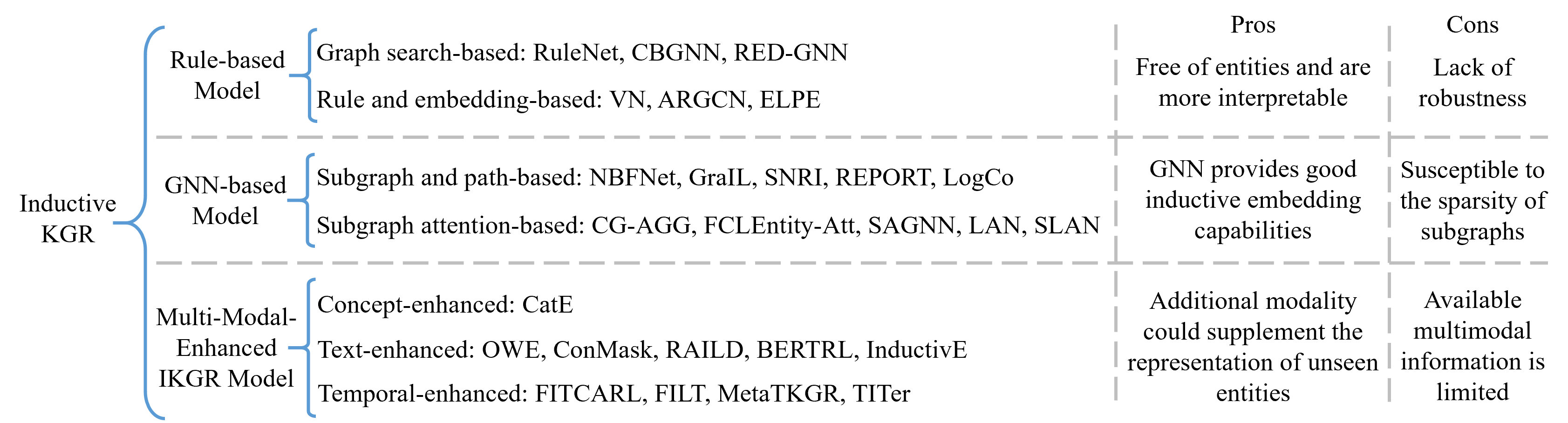}
    \caption{Taxonomy of inductive KGR approaches with the comparison of their advantages and disadvantages.}
    \label{fig:tax_ikgr}
\end{figure*}

\subsubsection{Text-Enhanced IKGR Model}

Motivated by the idea of DKRL~\cite{287}, OWE~\cite{369} uses the entity’s name and textual description to learn text embeddings for unseen entities, which are then projected into the KGE space to enhance the reasoning capability for unseen entities. Several subsequent models have refined both the text embedding techniques and the projecting mechanisms between embedding spaces. For instance, WOWE~\cite{370} and Caps-OWKG~\cite{371} employ attention networks and relation-aware similarity measures to compute weights for each word in an entity’s textual description. OWE-MRC~\cite{372} leverages machine reading comprehension techniques to extract meaningful short text snippets from lengthy descriptions for improving accurate text embedding. OWE-RST~\cite{373} designs a transfer function to map text embeddings into the KGR space. EmReCo~\cite{374} employs a relation-aware attention aggregator and gating mechanism to learn text embeddings for unseen entities under specific relations. ConMask~\cite{71} further utilizes relation masks, a fully convolutional neural network, and semantic mean pooling to extract relation-aware embeddings from the textual descriptions of entities and relations. MIA~\cite{375} modifies ConMask by modeling the interactions among the head entity’s description and name, the relation's name, and the candidate tail entity’s description, learning richer representations for unseen entities. Besides, SDT~\cite{376} integrates an entity’s structural information, description, and hierarchical type information into a unified framework for learning embeddings of unseen entities, while DMoG~\cite{328} fuses ontological information and textual descriptions to represent unseen relations.

In recent years, the emergence of LLMs has naturally benefited text processing. For example, Bi-Link~\cite{377} leverages the pre-trained language model BERT to learn text embeddings from the descriptions of entities and relations, enriching the relation embeddings with probabilistic rule-based prompt expressions. RAILD~\cite{378} fine-tunes a PLM to encode textual descriptions of entities and relations, learning embeddings for unseen entities and relations. BERTRL~\cite{379} linearizes the neighbor subgraph of an entity into serialized paths, which are then input into BERT for fine-tuning to encode neighbor information. InductivE~\cite{380} combines PLM and word embeddings to learn text embeddings for entities and further leverages a GNN to encode neighbor subgraphs, enhancing the representations of unseen entities by integrating both text embeddings and subgraph features. Besides, multi-modal KGR approaches that incorporate both text and visual information are equally applicable to inductive KGR tasks, which have been comprehensively reviewed in Section~\ref{sec:multi-modal} and will not be repeated here.

\subsubsection{Temporal-Enhanced IKGR Model}

Beyond text and images, some studies attempt to incorporate temporal information to enhance the representations of unseen entities. Such models can be considered as a comprehensive task addressing both temporal and inductive KGR. For instance, FITCARL~\cite{341} uses a time-aware Transformer to encode embeddings for unseen entities in a temporal KG. TITer~\cite{278} designs a relative time encoding function and a time reward strategy within a reinforcement learning-based multi-step reasoning framework, enabling multi-step reasoning for unseen entities on a temporal KG. MetaTKGR~\cite{340} samples and aggregates neighbor information from the temporal KG to learn time-aware representations for unseen entities, employing a meta learning mechanism to learn optimal sampling and aggregation parameters, thereby improving adaptability and robustness. Furthermore, FILT~\cite{381} mines concept-aware information from the visible entities in a temporal KG and transfers it to the representations of unseen entities, while using a time difference-based GNN to learn the contextual representations of unseen entities.

Although numerous inductive KGR models have been proposed, their performance remains limited due to the finite number of rules mined from KGs and external multi-modal sources, as well as the sparsity of neighbor subgraphs of unseen entities.

Fig.~\ref{fig:tax_ikgr} summarizes the representative models for each category in inductive KGR tasks and provides a summary of the advantages and limitations associated with each approach.

\section{Application of KGR}
\label{sec:app}

In this section, we introduce several benchmark datasets corresponding to diverse KGR tasks, which are commonly used for illustrating the effectiveness of KGR techniques in real‐world scenarios. Furthermore, KGR techniques not only enhance downstream applications such as question answering systems, recommendation systems, and visual reasoning in horizontal domains but also exhibit broad applicability across vertical domains including healthcare, commerce, cybersecurity, education, social governance, transportation, and environmental monitoring, ultimately facilitating the development of knowledge‐driven intelligent systems.

\subsection{Benchmarks}
\label{sec:benchmark}

For evaluating various KGR tasks, it is imperative to employ task-specific benchmark datasets. Both static single-step and multi-step KGR tasks are performed on static KGs, which allows these tasks to share common datasets as summarized in TABLE~\ref{tab:dataset}. In this table, \#Ent, \#Rel, \#Train, \#Valid and \#Test denote the amount of entities, relations as well as the samples in training set, validation set and testing set, respectively. In specific, while these datasets are inherently suitable for static single-step KGR, static multi-step KGR necessitates KGs that are sufficiently dense and enriched with numerous relations to enable the construction of multi-hop paths between entities within a limited number of reasoning steps. Consequently, datasets characterized by a higher number of relations such as FB15K, FB15K-237, and NELL-995 are commonly utilized for this task. 

\begin{table}[htbp]
\centering
\caption{Commonly Used Representative Datasets for Two Types of Static KGR Tasks}
\label{tab:dataset}
\setlength{\tabcolsep}{4pt} 
\renewcommand\arraystretch{1.5}
\begin{tabular}{lrrrrrr}
\toprule
Dataset & \#Ent & \#Rel & \#Train & \#Valid & \#Test \\
\midrule
Countries\cite{65}  & 271    & 2     & 1,110    & 24     & 24     \\
Kinship\cite{66}    & 104    & 25    & 8,544    & 1,068  & 1,074  \\
FB13\cite{67}       & 75,043  & 13    & 316,232  & 11,816  & 47,464  \\
FB122\cite{68}      & 9,738   & 122   & 91,638   & 9,595   & 11,243  \\
FB15K\cite{69}      & 14,951  & 1,345 & 483,142  & 50,000  & 59,071  \\
FB15K237\cite{70}   & 14,505  & 237   & 272,115  & 17,535  & 20,466  \\
FB20K\cite{71}      & 19,923  & 1,452 & 378,072  & 89,040  & 90,143  \\
FB5M\cite{72}       & 5,385,322 & 1,192 & 19,193,556 & 50,000  & 50,000  \\
WN11\cite{67}       & 38,588  & 11    & 110,361  & 5,212   & 21,035  \\
WN18\cite{70}       & 40,943  & 18    & 141,442  & 5,000   & 5,000   \\
WN18RR\cite{70}     & 40,559  & 11    & 86,835   & 2,924   & 2,924   \\
YAGO3-10\cite{73}   & 123,143 & 37    & 1,079,040 & 4,978   & 4,982   \\
YAGO37\cite{74}     & 123,189 & 37    & 420,623  & 50,000  & 50,000  \\
NELL-995\cite{75}   & 75,492  & 200   & 126,176  & 5,000   & 5,000   \\
\bottomrule
\end{tabular}
\end{table}

The most frequently employed datasets for dynamic KGR tasks are derived from two open-source databases: the Global Database of Events, Language, and Tone (GDELT)~\cite{77} and the Integrated Crisis Early Warning System (ICEWS)~\cite{78}. GDELT has been recording publicly available event information from global broadcasts, publications, and news media since 1979, offering a temporal resolution of 15 minutes. In contrast, ICEWS focuses on military and political events, providing event data with a daily temporal granularity~\cite{43}. Besides, the well-known KG Wikidata~\cite{79} can be leveraged for constructing dynamic KGR datasets when temporal information is incorporated. Representative datasets for dynamic KGR tasks are summarized in TABLE~\ref{tab:dynamic_datasets}, where ``Temp'' indicates temporal information, ``TS'' denotes timestamp, and ``TI'' implies time interval.

\begin{table}[htbp]
\centering
\caption{Representative Datasets for Dynamic KGR Tasks}
\label{tab:dynamic_datasets}
\setlength{\tabcolsep}{3pt} 
\renewcommand\arraystretch{1.5}
\begin{tabular}{lrrcrrr}
\toprule
Dataset & \#Ent & \#Rel & Temp & \#Train & \#Valid & \#Test \\
\midrule
GDELT\cite{76}      & 7,691   & 240   & TS   & 1,033,270  & 238,765  & 305,241  \\
ICEWS14\cite{79}    & 6,738   & 235   & TS   & 118,766   & 14,859   & 14,756   \\
ICEWS05-15\cite{80} & 10,488  & 251   & TS   & 386,962   & 46,092   & 46,275   \\
Wikidata12k\cite{81} & 12,554  & 24    & TI  & 2,735,685  & 341,961  & 341,961  \\
YAGO11k\cite{82}    & 10,623  & 10    & TI  & 161,540   & 19,523   & 20,026   \\
YAGO15k\cite{80}    & 15,403  & 34    & TI  & 110,441   & 13,815   & 13,800   \\
\bottomrule
\end{tabular}
\end{table}

In comparison to single-modal KGR datasets, those designed for MMKGR tasks integrate unstructured data from multiple modalities such as text, images, audio, and video. The statistical details for several commonly used open-source MMKGR datasets are presented in TABLE~\ref{tab:mm_datasets}, in which ``I'', ``T'' and ``V'' indicate the modalities of image, text, and video, respectively. The images in these three datasets are collected by image search engines or extracted from ImageNet. Besides, the textual descriptions are obtained from DBpedia or Wikidata.

\begin{table}[ht]
\centering
\caption{Representative Datasets for Multi-Modal KGR Tasks}
\label{tab:mm_datasets}
\setlength{\tabcolsep}{2pt} 
\renewcommand\arraystretch{1.5}
\begin{tabular}{lrrcrrr}
\toprule
Dataset & \#Ent & \#Rel & Modality & \#Train & \#Valid & \#Test \\
\midrule
FB-IMG-TXT\cite{83} & 11,757 & 1,231 & I+T & 285,850 & 34,863 & 29,580 \\
FB15K237-IMG\cite{84} & 14,541 & 237 & I & 272,115 & 17,535 & 20,466 \\
WN9-IMG-TXT\cite{83} & 6,555 & 9 & I+T & 11,741 & 1,319 & 1,337 \\
WN18-IMG\cite{84} & 40,943 & 18 & I & 141,442 & 5,000 & 5,000 \\
MKG-Wikipedia\cite{85} & 15,000 & 169 & I & 34,196 & 4,274 & 4,276 \\
MKG-YAGO\cite{86} & 15,000 & 28 & I & 21,310 & 2,663 & 2,665 \\
TIVA\cite{87} & 11,858 & 16 & V & 20,071 & 2,000 & 2,000 \\
\bottomrule
\end{tabular}
\end{table}

Existing studies on few-shot KGR primarily address scenarios involving long-tailed relations, where only a limited number of reference triples in the support set are available for such relations. The datasets for this task are typically constructed based on NELL, Wikidata, and Freebase. Within these datasets, any relation associated with 50-500 triples is classified as a few-shot relation. Commonly used public datasets for few-shot KGR are listed in TABLE~\ref{tab:fewshot_datasets}. Each dataset provides the total number of triples and the ratio of the splits into training, validation, and test sets.

\begin{table}[ht]
\centering
\caption{Representative Datasets for Few-Shot KGR Tasks}
\label{tab:fewshot_datasets}
\setlength{\tabcolsep}{3pt} 
\renewcommand\arraystretch{1.5}
\begin{tabular}{lrrrc}
\toprule
Dataset & \#Ent & \#Rel & \#Triple & \#Train/Valid/Test Splits\\
\midrule
NELL-One\cite{89}    & 68,545    & 358   & 181,109   & 51/5/1  \\
Wiki-One\cite{89}    & 4,868,244 & 822   & 5,859,240 & 133/16/34      \\
FB15K-One\cite{90}   & 14,541    & 231   & 281,624   & 75/11/33      \\
\bottomrule
\end{tabular}
\end{table}

Inductive KGR datasets are mainly derived from static KGR datasets such as FB15K-237, WN18RR, and NELL-995 as well as from large-scale KGs like DBPedia and Wikidata. Under the inductive reasoning paradigm, all entities in the test set are ensured to be unseen during training. Notably, datasets based on FB15K-237, WN18RR, and NELL-995 are provided in four versions, namely FB15K-237 v1/v2/v3/v4~\cite{91}, WN18RR v1/v2/v3/v4~\cite{91}, and NELL-995 v1/v2/v3/v4~\cite{91}. Besides, both DBPedia50k~\cite{71} and Wikidata5M~\cite{92} offer inductive as well as transductive versions, though only the inductive versions are considered here. TABLE~\ref{tab:inductive_datasets} summarizes these widely used public inductive KGR datasets.

\begin{table}[ht]
\centering
\caption{Representative Inductive KGR Datasets}
\label{tab:inductive_datasets}
\resizebox{\columnwidth}{!}{%
  \begin{tabular}{@{}c@{}}
    {\setlength{\tabcolsep}{3pt}%
    \renewcommand\arraystretch{1.5}
    \begin{tabular}{cc|rrr|rrr|rrr}
      \toprule
      \multirow{2}{*}{Version} & \multirow{2}{*}{Set} & \multicolumn{3}{c|}{FB15K237\cite{91}} & \multicolumn{3}{c|}{WN18RR\cite{91}} & \multicolumn{3}{c}{NELL-995\cite{91}} \\
      \cmidrule(lr){3-11}
       &  & \#Ent & \#Rel & \#Triple & \#Ent & \#Rel & \#Triple & \#Ent & \#Rel & \#Triple \\
      \midrule
      \multirow{2}{*}{v1} & Train & 2,000  & 183  & 5,226  & 2,746  & 9   & 6,678  & 10,915  & 14  & 5,540  \\
                         & Test  & 1,500  & 146  & 2,404  & 922    & 9   & 1,991  & 225     & 14  & 1,034  \\
      \midrule
      \multirow{2}{*}{v2} & Train & 3,000  & 203  & 12,085 & 6,954  & 10  & 18,968 & 2,564   & 88  & 10,109 \\
                         & Test  & 2,000  & 176  & 5,092  & 2,923  & 10  & 4,863  & 4,937   & 79  & 5,521  \\
      \midrule
      \multirow{2}{*}{v3} & Train & 4,000  & 218  & 22,394 & 12,078 & 11  & 32,150 & 4,647   & 142 & 20,117 \\
                         & Test  & 3,000  & 187  & 9,137  & 5,084  & 11  & 7,470  & 4,921   & 122 & 9,668  \\
      \midrule
      \multirow{2}{*}{v4} & Train & 5,000  & 222  & 33,916 & 3,861  & 9   & 9,842  & 2,092   & 77  & 9,289  \\
                         & Test  & 3,500  & 204  & 14,554 & 7,208  & 9   & 15,157 & 3,294   & 61  & 8,520  \\
      \bottomrule
    \end{tabular}%
    }\\[1em]\\
    \setlength{\tabcolsep}{7.2pt}%
    \renewcommand\arraystretch{1.5}
    \begin{tabular}{l|ccc|ccc}
      \toprule
      Dataset & \multicolumn{3}{c|}{DBPedia50k\cite{71}} & \multicolumn{3}{c}{Wikidata5M\cite{92}} \\
      \cmidrule(lr){2-7}
       & \#Ent & \#Rel & \#Triple & \#Ent & \#Rel & \#Triple \\
      \midrule
      Training Set & 24,624  & 351  & 32,388  & 4,579,609  & 822  & 20,496,514 \\
      Test Set     & 3,636   & -    & 6,459   & 7,475      & 201  & 6,894 \\
      \bottomrule
    \end{tabular}
  \end{tabular}%
}
\end{table}

\subsection{Downstream Tasks in Horizontal Domains}
\label{sec:hor-downstream}

The existing KGR techniques have been extensively applied in a variety of horizontal domains, including question answering systems, recommendation systems, and visual reasoning. These applications fundamentally depend on the prior knowledge encapsulated within KGs and employ KGR techniques to enhance both accuracy and explainability.

\subsubsection{Question Answering Systems}

\begin{figure*}
    \centering
    \includegraphics[width=1\linewidth]{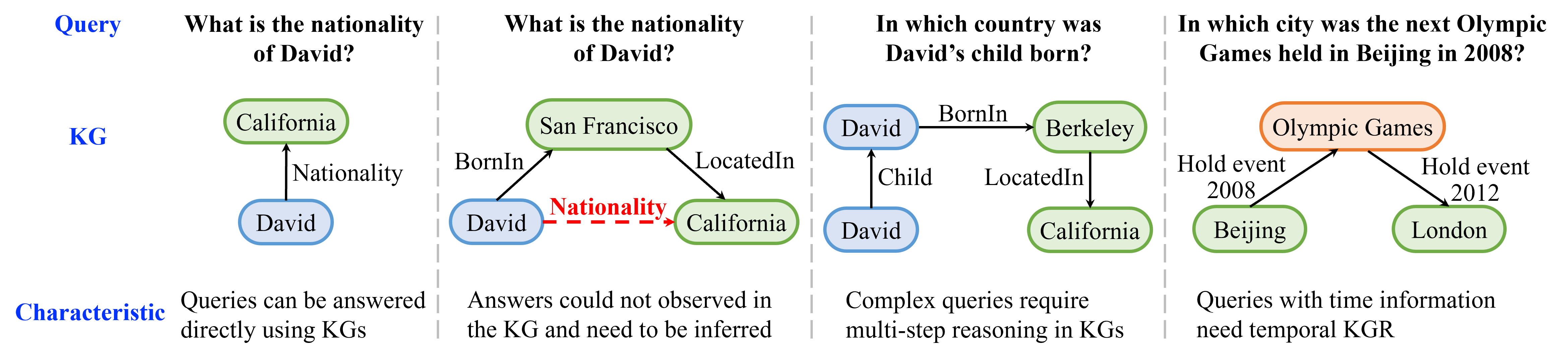}
    \caption{Illustrative examples of the KGR technique applied to QA systems.}
    \label{fig:qa}
\end{figure*}

KG-based question answering (KGQA) systems have been integrated into numerous practical applications, such as intelligent customer service robot. Their significant advantage is the ability to generate precise responses by leveraging structured knowledge stored in KGs~\cite{382}. Nevertheless, KGQA systems encounter three primary challenges that require advanced KGR techniques: (a) For queries whose answers are not explicitly stored in the KG, KGR techniques have to infer the correct answer. (b) For complex queries, multi-hop reasoning across the KG is essential to derive accurate responses. (c) For queries incorporating temporal constraints, temporal KGR is required to retrieve time-sensitive answers. Various query types alongside their corresponding KGs and question answering (QA) characteristics are illustrated in Fig.~\ref{fig:qa}.

(a) \textbf{Question Answering Systems for Simple Queries}:
To infer new knowledge and enhance the robustness of QA systems, KGE techniques have been employed to derive embeddings for entities and relations within KGs, thereby facilitating KGR. KEQA~\cite{383} projects KGs into a continuous vector space and separately embeds the head entity and the relation derived from the query into their respective vector spaces, effectively reformulating the QA task as a static single-step KGR task for predicting the tail entity. Similarly, models such as TRL-KEQA~\cite{384}, TransE-QA~\cite{385}, and CAPKGQA~\cite{386} integrate KGE into KGQA systems to improve reasoning over previously unknown answers. Although these approaches efficiently utilize implicit features within the KG for inference, they remain confined to queries involving only first-order relations.

(b) \textbf{Question Answering Systems for Unknown Answers}:
For more intricate QA tasks that require multi-step reasoning, EmbedKGQA~\cite{387} extends KEQA by incorporating a BERT model to encode complex queries. It computes the similarity between query embeddings and candidate entity embeddings to retrieve the correct answer. However, EmbedKGQA does not incorporate the path information inherent to multi-step reasoning, which limits its effectiveness for complex queries. To overcome this limitation, models such as PKEEQA~\cite{388}, PA-KGQA~\cite{389}, and HamQA~\cite{390} explicitly integrate path information extracted from KGs into the KGE process, thereby enhancing multi-step reasoning and enhancing the prediction of unknown answers. Besides, BRGNN~\cite{391} utilizes bidirectional reasoning to generate multi-step query paths, capturing richer semantic correlations than the previous methods and consequently improving QA accuracy. GRRN~\cite{392} further exploits deep path extraction techniques to infer potential relations between entities, thereby reinforcing commonsense QA. Recognizing that relations convey substantial semantic information in multi-step paths, Li et al.~\cite{393} leverage relation embeddings derived from KGE models to refine both query and path representations, which results in more efficient multi-step QA.

(c) \textbf{Question Answering Systems for Complex Queries}:
DSSAGN~\cite{394} integrates both syntactic and semantic information by employing an enhanced graph neural network to model dependencies between various segments of the KG and the query, thereby facilitating accurate multi-hop reasoning. The explicit representation of dependency relationships also enhances the explainability of the QA process. In a similar idea, Jiao et al.~\cite{395} incorporate relation embeddings to capture nuanced interactions among entities, supporting multi-hop reasoning and enabling seamless traversal across relations. Zhou et al.~\cite{396} improve entity mention linking robustness in the chemical domain by employing a BERT-based model and embedding the KG in multiple representation spaces. Their methodology combines several knowledge representation models to generate diverse candidate answers, which are then re-ranked using a score alignment model while executing multi-step reasoning to address complex multi-hop QA. Moreover, CF-KGQA~\cite{397} integrates causal relationships into the multi-hop QA process, thereby achieving more precise answers by effectively utilizing inherent causal structures within the KG.

(d) \textbf{Temporal Question Answering Systems with Explicit Time Representations}:
Temporal KG question answering (TKGQA) has recently attracted significant attention as an application of temporal KGR in QA systems. TwiRGCN~\cite{398} incorporates a time-weighted mechanism within GCN to encode entities and employs a gating mechanism to determine whether the answer corresponds to an entity or a timestamp, with the prediction result informing the answer scoring. However, this approach encounters efficiency issues when scaled to large KGs. In contrast, CRONKGQA~\cite{399} extends the static KGQA model EmbedKGQA~\cite{387} to accommodate temporal queries. It constructs a large-scale temporal QA dataset, namely CRONQUESTIONS, and employs a temporal KGE model to produce time-aware embeddings for entities and relations. By scoring candidate answers from both entity and temporal perspectives, CRONKGQA significantly enhances TKGQA performance. Similarly, TempoQR~\cite{400} transforms temporally annotated questions into entity-aware and time-aware query representations using dedicated embeddings, which are subsequently processed through a temporal KGE model to score candidate answers. Notably, current temporal KBQA models primarily address explicit temporal information and tend to struggle with implicit temporal constraints, such as ``before/after'' or ``first/next/last''.

(e) \textbf{Temporal Question Answering Systems with Implicit Temporal Constraints}:
To conduct QA tasks involving implicit temporal constraints, CTRN~\cite{401} extends the model TempoQR by developing a complex temporal reasoning network capable of capturing implicit temporal features and relational representations within queries. This integration facilitates the generation of implicit temporal relation representations, which are subsequently utilized in temporal KGE for answer inference to effectively handle implicit temporal constraints. For addressing complex temporal QA, EXAQT~\cite{402} initially extracts dense KG subgraphs relevant to the query and enhances event data via a fine-tuned BERT model. Then, it encodes time-aware entity representations using the KGE model R-GCN~\cite{100}, recasting the QA process as a node classification task. GATQR~\cite{403} further employs a graph attention network to capture implicit temporal information in complex queries and uses a BERT model to learn temporal relation representations, which are integrated with pretrained temporal KG embeddings to identify the highest-scoring entity or timestamp as the answer. Besides, recent researches focus on exploring the integration of LLMs with temporal KGQA techniques to improve implicit temporal reasoning. Prog-TQA~\cite{404} introduces fundamental temporal operators and a novel self-evolving programming approach that leverages the contextual learning capabilities of LLMs to interpret combinatorial temporal constraints. Meanwhile, it generates preliminary programs via few-shot learning and aligns them with the temporal KG through a linking module for guiding LLMs toward accurate inference. Similarly, GenTKGQA~\cite{405} develops a two-stage approach, which proposes a subgraph retrieval enabling LLMs to extract subgraphs under temporal constraints and fuses subgraph encodings with textual representations generated by LLMs. Furthermore, it fine-tunes open-source LLMs via instruction tuning to enhance their comprehension of temporal and structural semantics across the retrieved events.

\subsubsection{Recommendation Systems}

While QA systems focus on delivering precise answers to user queries, recommendation systems are designed to generate personalized suggestions that align with user preferences for enhancing user experience. Traditional collaborative filtering-based recommendation systems rely solely on users’ historical records to evaluate correlations between users or items. However, this technique often struggle with deeply capturing individual preferences.

To improve user experience, it is essential to not only provide recommendations but also conduct an in-depth analysis of user preferences while ensuring explainability and diversity of the recommendation results. These requirements illustrate the critical role of KGR techniques in recommendation systems. Therefore, we discuss representative approaches from two key perspectives: (a) leveraging KGR to uncover deeper insights into user preferences and (b) enhancing the explainability of recommendation results via KGR. Based on these two perspectives, we categorize the existing KGR-enhanced recommendation systems into the following three types.

(a) \textbf{KGR for Mining User Preferences}:
KGCN~\cite{406} is the first to employ GCN to aggregate neighborhood information for encoding entities in a KG. This approach captures both local and global structural features. By computing the similarity between item features and user features, enhanced with contextual semantic information from the KG, the model generates personalized recommendations. The inherent implicit reasoning capability of entity embeddings improves the understanding of item characteristics and user preferences. Based on this idea, KGNCF-RRN~\cite{407} encodes relational paths between entities using a modified residual recurrent neural network (RRN) to enable the multi-hop reasoning. Besides, KGECF~\cite{408} treats user–item interactions as a KG with a single relation and leverages the KGE model RotatE~\cite{117} to transform collaborative filtering into a single-step KGR task.

(b) \textbf{KGR for Enhancing Explainability of Recommendations}:
To address the challenge of explainability in recommendation systems, the existing techniques typically employ explicit reasoning over the KG by using path information to assess the similarity between items or users~\cite{409}. PGPR~\cite{410} integrates a reinforcement learning-based multi-hop reasoning approach for recommendation. It employs several strategies, including soft reward, user-conditioned action pruning, and multi-hop reasoning scoring function to train an agent that could search reasoning paths in the KG. These paths are leveraged not only for generating recommendation results but also for providing rational explanations. Similarly, KPRN~\cite{12} learns path embeddings via fusing entity and relation embeddings, and applies a weighted pooling operation to differentiate the importance of various paths, leading to better explainability. CogER~\cite{411} is the first effort to emulate human cognitive processes in recommendation tasks by using a quick, intuitive estimation (System 1) and then refining it through a reinforcement learning-based multi-hop reasoning framework (System 2). This iterative collaboration yields high-quality recommendation results along with transparent reasoning paths. Furthermore, Hsu et al.\cite{412} designs a recommendation algorithm that identifies fund entities likely to have purchasing correlations with users by searching for intermediate paths between user and fund entities. Lee et al.\cite{413} combines KGs with review texts to improve explainability via executing a GCN-based collaborative filtering algorithm concurrently on both the user–item interaction graph and the KG, generating a path from the target user to the recommended item as a rationale. Finally, Markchom et al.~\cite{414} propose a meta-path grammar coupled with a machine translation model to convert complex meta-paths into natural language-like explanations, enhancing the explainability of the recommendation results.

(c) \textbf{Balancing Personalization and Explainability with KGR}:
Since many explainable recommendation algorithms based on KGs ignore the heterogeneity of user preferences, resulting in biased or unfair recommendations, Fu et al.\cite{415} introduce a fairness-constrained approach that incorporates heuristic re-ranking based on multi-hop reasoning over KGs to ensure equitable recommendation results. KRRL\cite{416} employs self-supervised reinforcement learning to enhance explainability in massive open online courses (MOOCs) by reasoning over both course content and user preferences via a KG. Besides, several studies such as RippleNet\cite{418}, AKUPM~\cite{419}, RCoLM~\cite{420}, KGCN~\cite{407}, KGAT~\cite{421}, IntentGC~\cite{422}, and AKGE~\cite{423} integrate KG semantic information with multi-hop path reasoning to improve the representation of users and items. Similarly, Ryotaro et al.\cite{417} present a modified KGAT model that compresses auxiliary information to reduce computational costs while maintaining accuracy and explainability. These hybrid approaches combine the inherent transparency of path-based KGR models with a deeper exploration of user preference patterns, striking a balance between personalization and explainability.

\subsubsection{Visual Reasoning}

Recent advances in computer vision are increasingly moving beyond pure perception toward higher-level cognition, necessitating the development of robust visual reasoning capabilities. In this section, we review visual reasoning tasks enhanced by KGR, including visual question answering, cross-modal retrieval, and scene understanding.

(a) \textbf{Visual Question Answering}: Traditional visual question answering (VQA) systems analyze the question and image content to generate an answer, but they often struggle with questions that require commonsense or external knowledge. To address this limitation, Wang et al.~\cite{424} introduce the dataset named FVQA, which represents each instance as an tuple ``image–question–answer–supporting fact subgraph'', requiring the integration of background knowledge. Based on this idea, Wang et al.~\cite{425} propose a model for FVQA that links visual concepts detected in an image to corresponding entities in a KG and incorporates the path information between the question concepts and those in the KG as a reasoning process to derive the answer. This approach significantly outperforms conventional LSTM-based VQA approaches. Graphhopper~\cite{426} takes a different approach by first extracting a scene graph from the image to represent objects, their attributes and relations, and then applying a reinforcement learning-based multi-hop reasoning technique to automatically search for reasoning paths in the extracted scene graph to obtain the answer. Besides, models such as the Hypergraph Transformer~\cite{427}, CMRL~\cite{428}, and KRISP~\cite{429} build query-relevant KGs by integrating information from the question, image targets, and external KGs, and design tailored reasoning methods to derive effective answers. Furthermore, recent research~\cite{430} has demonstrated that leveraging multi-modal LLMs could outperform fine-tuned models on VQA tasks that require external knowledge. However, the explicit integration of external KGs and reasoning mechanisms remains essential for further enhancing VQA performance.

(b) \textbf{Cross-Modal Retrieval}: Cross-modal retrieval aims to identify the most semantically similar image or video given a textual query. KCR~\cite{431} leverages a ResNet to extract image features and an enhanced BERT model to capture text features enriched by relevant external KGs. It then employs a triple loss strategy to jointly optimize the alignment of image and text features. In contrast, MMRG~\cite{432} constructs both text-based and visual KGs, and enhances visual relational reasoning by employing pre-training tasks namely attribute masking and contextual prediction similar to those in BERT, improving the performance of graph matching-based cross-modal retrieval. IRGR~\cite{433} further refines this approach by introducing three distinct reasoning mechanisms namely intra-modal, inter-modal, and instance-based KGR that collectively model the relations among the various KGs and their neighboring instances to achieve a more effective similarity matrix.

(c) \textbf{Scene Graph Generation}: Scene graph generation involves detecting object entities within an image and inferring the semantic relations among them, representing the image as a directed graph with entities as nodes and relations as directed edges similar to a KG. GB-Net~\cite{434} uses Faster R-CNN to generate an initial scene graph, which is then enhanced by bridging it with external KGs such as ConceptNet~\cite{435} and WordNet. A message propagation algorithm updates the entity embeddings in the scene graph, and subsequent KGR enriches both the scene and external graphs. To address the degradation of detection accuracy in noisy images, HiKER-SGG~\cite{436} improves upon GB-Net by constructing a hierarchical KG from external sources and bridging it with the initial scene graph. Reasoning on the hierarchical graph guides the prediction of entities and their relations in a manner consistent with the semantic hierarchy of concepts, which could enhance the robustness. CGR~\cite{437} further bridges the semantic gap between visual scenes and external KGs by selectively composing knowledge routing paths through matching and retrieval of diverse paths, which improves generalization. Finally, COACHER~\cite{438} incorporates a novel graph mining module that leverages the neighbor and path information from commonsense KGs. By integrating this information into the scene graph generation framework, COACHER performs relation reasoning under low-resource scenarios, leading to the generation of more comprehensive scene graphs.

\subsection{Downstream Tasks in Vertical Domains}
\label{sec:ver-downstream}

In recent years, KGR technology empowers many downstream applications with knowledge-based inference capabilities. This approach has attracted significant attention and is now deployed in various vertical domains, including healthcare, business, information security, education, social sciences, transportation, and environmental management.

\subsubsection{Healthcare Domain}

Zhu et al.~\cite{439} construct a disease-specific KG, SDKG-11, and employ multi-modal KGR by projecting text embeddings into a hyperplane to uncover novel insights about specific diseases. Chai et al.~\cite{440} apply the KGE model TransE~\cite{69} to derive embeddings for a medical KG and then utilize a BiLSTM network to predict associations between pathologies and diseases. Besides, both Nyamabo et al.~\cite{441} and Lin et al.~\cite{442} leverage KGs to capture higher-order structural and semantic relationships among drugs and relevant entities, enhancing the accuracy of drug–drug interaction predictions. Gong et al.~\cite{443} integrate electronic medical records with a medical KG, using KGE techniques to recommend safe medications. Zheng et al.~\cite{444} combine a heterogeneous graph attention network with global graph structures and heterogeneous features to capture complex biomedical information via learned embeddings. Moreover, Gao et al.~\cite{445} leverage KGE to infer novel drug–disease interactions, demonstrating its effectiveness and practical value in drug repurposing for Alzheimer’s disease.

\subsubsection{Business Domain}

Deng et al.~\cite{446} develop OpenBG, a multi-modal commercial KG at a billion-scale, which is pre-trained by multi-modal KGE models and vision–language foundational models. This framework has been effectively applied to downstream e-commerce tasks such as product classification and prediction. Zhang et al.~\cite{447} propose a graph attention network-based KGR approach that generates transferable logical rules, which is validated on real-world e-commerce KG datasets and demonstrate practicality in brand completion, lifestyle recommendation, and automatic categorization. Yang et al.~\cite{448} improve reasoning accuracy through a multi-feature reasoning module that employs a dynamic strategy network to perform multi-hop reasoning over the KG, thereby predicting product substitutability and complementarity. Mitropoulou et al.~\cite{449} utilize various KGE techniques to uncover latent links among dispersed customer needs and to identify implicit relations between customer requirements and product features. Kosasih et al.~\cite{450} leverage GNNs along with neuro-symbolic machine learning techniques to infer multiple implicit relations in supply chain risk management. Yang et al.~\cite{451} introduce temporal information to construct a dynamic KG for enterprise risk by aggregating neighborhood information from KGs and employing LSTM networks to capture temporal patterns, achieving more accurate predictions of risk-related entities and relations. Besides, Zhang et al.~\cite{452} design a pre-trained KGE model for e-commerce product graphs, which is subsequently applied to tasks such as product recommendation and classification.

\subsubsection{Cybersecurity Domain}

Sikos~\cite{453} constructs a domain-specific KG for information security and employs rule-learning-based reasoning models to automatically infer relations among vulnerabilities, weaknesses, platforms, and attack patterns with targeted defense strategies. Qi et al.~\cite{454} utilize the rule-learning algorithm AMIE~\cite{202} to mine logic rules and discover new network insights by learning rule patterns for each relation. Ezekia Gilliard et al.~\cite{455} apply a KGR technique based on the KGE model TransH to enhance the accuracy of automatic attack pattern recognition. Furthermore, to address the challenging issues of contract version diversity and limited detection time, Hu et al.~\cite{456} develop a KG based on the programming language solidity and apply rule learning algorithms for reasoning, achieving efficient detection and localization of multiple defects in smart contracts.

\subsubsection{Other Domains}

In the field of education, Liang et al.~\cite{457} combine GCN with reinforcement learning to learn multi-path embeddings between entities, enabling efficient KGR and interpretable recommendations for various learning resources. In the field of sociology, Zhou et al.~\cite{458} learn hierarchical KG embeddings that integrate global and domain-specific information, optimizing positional similarity losses in reasoning models to generate location embeddings for predicting socio-economic indicators. Gao et al.~\cite{459} construct a tourism KG encompassing 340 cities and employ a GNN-based KGE model to formulate the competitive analysis task as a KGR task, which provides in-depth analyses of tourist preferences and attraction competition. Within the transport sector, Zeng et al.~\cite{460} pioneer the development of a metro KG using historical origin–destination matrices and complex network construction techniques, then assign semantic types to each station to build a subway KG. A graph attention network-based model is subsequently employed to predict passenger flow, which is validated on the subway systems of two Chinese cities Shen zhen and Hang zhou. In environment science applications, Liu et al.~\cite{461} tackle the challenging task of oil spill detection through an innovative integration of KG techniques. They initially identify non-oil spill regions using predefined feature thresholds, and then combine rule learning with GNN-based KGR techinque to enhance the accuracy and generalization of oil spill detection, particularly in complex marine environments where oil spills may exhibit variable spectral signatures and be obscured by other phenomena.

\section{Challenge and Opportunity}
\label{sec:challenge}

Although KGR techniques tailored to various tasks have achieved significant research progress, several promising avenues remain unexplored. These approaches focus on two key perspectives: (1) the models specific to special characteristics of the KG such as sparse KGR, uncertain KGR, and KG error detection as well as (2) models aimed at reasoning process, including trustworthy KGR and the integration of KGR with LLMs.

\subsection{Sparse KGR}

Due to the inherent sparsity of real-world KGs, traditional KGE models often struggle to capture effective topological features, leading to suboptimal reasoning results. To address this issue, researchers have explored sparse KGR techniques. IterE~\cite{227} and HoGRN~\cite{462} leverage logical rules to capture inter-relational correlations for enriching semantics and enhancing the embeddings of entities and relations. Jia et al.~\cite{463} augment sparse KGs by incorporating external knowledge sources and fusing them using GNN technique, achieving performance improvements. KRACL~\cite{464} combines contrastive loss with cross-entropy loss to provide stronger supervisory signals for sparse entities by introducing a larger number of negative triples. Moreover, BERT-ConvE~\cite{465} integrates the transfer learning capabilities of BERT with the KGE model ConvE~\cite{126} by using textual representations to compensate for structural sparsity within the graph.

Furthermore, the sparsity of KGs often results in equally sparse paths between entities, limiting the effectiveness of multi-hop reasoning. To alleviate this issue, both DacKGR~\cite{466} and RuMER-RL~\cite{467} dynamically add edges as auxiliary actions during reinforcement learning-based path searching processes. Similarly, WAR~\cite{468} employs random walks to insert additional information into paths, improving multi-hop reasoning performance on sparse KGs. However, most current approaches supplement existing knowledge by introducing a degree of uncertainty, and they do not fully resolve the sparsity issue. Consequently, sparse KGR remains a significant but challenging research direction.

\subsection{Uncertain KGR}

Uncertainty reasoning is a classical and critical area in artificial intelligence, reflecting the fact that real-world knowledge often carries inherent uncertainty. Recent researches have increasingly focused on uncertain KGs by incorporating confidence measures into their reasoning tasks. The primary challenge is to represent uncertainty levels and to infer uncertain results accurately. In terms of conventional vector and tensor representations are insufficient for capturing uncertainty, exploiting the specific embedding spaces is a significant idea for this task. For instance, KG2E~\cite{109} represents entities and relations as Gaussian distributions, where the covariance encodes the inherent uncertainty. BEUrRE~\cite{469} models each entity as a rectangle, using the rectangular region to express uncertainty, and represents relations as affine transformations between these rectangular embeddings. SUKE~\cite{470} leverages the confidence scores of positive triples during training to build a confidence generator that predicts the confidence of inferred triples. Besides, MUKGE~\cite{471} introduces an uncertainty resource ranking reasoning algorithm for more precise confidence inference, while UKRM~\cite{472} employs the rule learning algorithm for treating rule mining as a sequence-to-sequence task and utilizes a differentiable reasoning mechanism based on TensorLog~\cite{473} to derive new factual triples, which are then assigned confidence scores by a PLM. Despite these advances, existing models mainly focus on representing uncertainty, and lacks an effective uncertainty measurement mechanism in the reasoning process.

\subsection{KG Error Detection}

The construction of large-scale KGs using automated extraction techniques inevitably introduces noise and errors. For instance, the automatically constructed KG NELL contains approximately 26\% erroneous knowledge, corresponding to roughly 600,000 incorrect triples~\cite{474}. Consequently, reasoning results derived from such noisy KGs may be unreliable. Since manual validation of large-scale KGs is impractical, automatic error detection techniques are imperative. The key challenge lies in the scarcity or absence of reliable ground-truth labels for erroneous knowledge, making the discovery and representation of errors critical. Current approaches to error detection can be broadly classified into rule-based methods and KGE-based models. The former detects anomalies by mining violations of predefined inductive rules. For example, Belthd et al.~\cite{475} propose a set of soft rules based on common graphical patterns to specify valid inter-entity relations and flag anomalies. CAGED~\cite{474} treats each triple as a node, expanding the KG into a hyper-graph, and jointly trains a KGE framework with contrastive learning to assess the credibility of each triple. Similarly, HEAR~\cite{476} leverages path information between entity pairs to evaluate the correctness of a triple. Given the critical importance of precise error detection, integrating human oversight, external knowledge, and multi-modal data represents a promising direction for future research.

\subsection{Trustworthy KGR}

At present, the explainability of KGR mainly relies on rule learning-based models and multi-hop reasoning techniques, which provide explainability through symbolic rules~\cite{477} and explicit path information~\cite{478}. However, these approaches often struggle with limited generalization compared to KGE models. A promising research direction is to integrate symbolic logic with KGE through a neuro-symbolic reasoning framework, balancing explainability with generalization. Furthermore, existing temporal rule learning models such as StreamLearner~\cite{270}, TLogic~\cite{271}, LCGE~\cite{265}, and TLIP~\cite{272} have only addressed a few simple temporal patterns. For instance, StreamLearner could only generate temporal rules where all atoms in the rule body occur simultaneously. TLogic, LCGE and TLIP extend these temporal rules by considering non-decreasing temporal orders among atoms. Nevertheless, these models still fall short of traditional rule learning approaches in both the diversity of discovered temporal rules and rule search efficiency. Thus, there is an urgent need for temporal rule learning algorithms that can capture more complex temporal patterns among events to enhance the accuracy and explainability of temporal KGR. Besides, evaluating the explainability of KGR remains underexplored. Xu et al.~\cite{479} propose a human-centric evaluation platform for assessing the credibility of explanations in KGR. However, standardized metrics and methods for explainability assessment are still lacking. While most KGR researches have focused on deductive and inductive paradigms, abductive reasoning on KGs remains under-investigated. To fill this gap, Zhang et al~\cite{487} introduce complex logical hypothesis generation to implement abductive KGR, proposing RLF-KG with reinforcement learning to enhance hypothesis generation by minimizing discrepancies between observations and conclusions.

\subsection{LLM-enhanced KGR}

In recent years, LLM techniques have demonstrated outstanding reasoning capabilities in many fields, including chain-of-thought (CoT) methodologies. However, such approaches often lack transparency both in the reasoning process. Consequently, integrating KGR with LLMs has emerged as a research hotspot in knowledge-aware applications. The existing research in this area can be categorized into following several approaches.

(a) \textbf{Utilizing Single-Step Reasoning of LLMs}: KG-GPT~\cite{480} employs LLMs to retrieve relevant KG subgraphs and generate reasoning results. Similarly, MPIKGC~\cite{481} leverages the reasoning, explanation, and summarization capabilities of LLMs to respectively expand entity descriptions, interpret relations, and extract subgraph structures to enhance the performance of KGR.

(b) \textbf{Utilizing Complex Logical Reasoning of LLMs}: LARK~\cite{482} combines graph extraction algorithms with LLMs, reformulating complex KGR as a synthesis of context search within the KG and logical query reasoning.

(c) \textbf{Utilizing Rule Learning Capabilities of LLMs}: ChatRule~\cite{483} designs an LLM-based rule generator that produces logical rules from the semantic and structural information encoded in the KG. GENTKG~\cite{gentkg} develops a retrieval augmented generation (RAG) framework which combines a temporal rule learning and an efficient instruction tuning for the temporal KGR task.

(d) \textbf{Utilizing Multi-Hop Reasoning of LLMs}: Nguyen et al.~\cite{242} explore the chain-of-thought (CoT) reasoning ability of LLMs in multi-hop reasoning tasks. By proposing both discriminative and generative CoT evaluation paradigms, they assess the accuracy of LLMs in KGR and generating coherent CoT outputs.

(e) \textbf{Utilizing Temporal Reasoning of LLMs}: LLM-DA~\cite{484} leverages LLMs to analyze historical data and extract temporal logical rules, which are then applied to temporal KGR. Besides, both Xia et al.~\cite{485} and Luo et al.~\cite{486} exploit the CoT reasoning capabilities of LLMs by iteratively reasoning based on higher-order historical information, effectively treating temporal KGR as a dual process of fine-tuning LLMs along historical chains and generating reasoning results.

By harnessing the sophisticated reasoning abilities and the extensive implicit knowledge contained within LLMs, there is substantial potential to enhance the performance of various tasks, including few-shot KGR, inductive KGR, multi-modal KGR, sparse KGR, and KGR error detection.

\section{Conclusion}
\label{sec:conclusion}

KGR is one of the key technologies in cognitive intelligence. Our survey categorizes the KGR tasks into six primary tasks: static single-step KGR, static multi-step KGR, dynamic KGR, multi-modal KGR, few-shot KGR, and inductive KGR. For each category, we carefully review and analyze relevant models including some advanced LLM-based models while comparing the strengths and weaknesses of different approaches. Besides, we introduce some commonly used benchmark datasets specific to each task and summarize typical applications of KGR in both horizontal and vertical domains over recent years. Furthermore, we discuss the remaining challenges of KGR from the perspectives of reasoning with existing knowledge and reasoning process, highlighting key research trends and outlining promising future directions in the field of KGR.

\bibliographystyle{IEEEtran}
\bibliography{IEEEabrv, kgrsurvey}

\end{document}